\documentclass{article}

\IfFileExists{neurips_2026.sty}{%
  \usepackage[preprint]{neurips_2026}
}{%
  \usepackage[a4paper, total={6in, 8in}]{geometry}
  \usepackage[numbers]{natbib}
}

\usepackage{xcolor}
\usepackage{algorithm}
\usepackage{algorithmic}
\usepackage{amssymb}
\usepackage{mathtools}
\usepackage{physics}
\usepackage{enumitem}
\usepackage{tabularx}
\usepackage{hyperref}
\usepackage{amsthm}
\usepackage{booktabs,makecell,threeparttable,array}

\DeclareMathOperator{\KL}{KL}

\DeclareMathOperator{\E}{\mathbb{E}}
\DeclareMathOperator{\Var}{Var}
\newcommand{\C}{\mathbb{C}}
\newcommand{\R}{\mathbb{R}}
\newcommand{\N}{\mathbb{N}}

\newcommand{\ii}{\mathrm{i}}
\newcommand{\D}{\mathcal{D}}
\newcommand{\M}{\mathcal{M}}
\newcommand{\F}{\mathcal{F}}
\DeclareMathOperator{\diam}{diam}

\newcolumntype{Y}{>{\raggedright\arraybackslash}X}

\newtheorem{Theorem}{Theorem}
\newtheorem{Lemma}{Lemma}
\newtheorem{Remark}{Remark}
\newtheorem{Definition}{Definition}
\newtheorem{Corollary}{Corollary}
\newtheorem{Proposition}{Proposition}

\title{Learning Latent Graph Geometry via Fixed-Point Schr\"odinger-Type Activation: A Theoretical Study}

\author{%
  Dmitry Pasechnyuk-Vilensky\\
  MBZUAI, UAE \\
  \texttt{dmivilensky1@gmail.com} \\
  \And
  Martin Taká\v{c} \\
  MBZUAI, UAE \\
  \texttt{takac.mt@gmail.com}
}

\begin{document}

\maketitle

\begin{abstract}
We study neural architectures in which each hidden layer is defined by the stationary state of a dissipative Schr\"odinger-type dynamics on a learned latent graph. On stable branches, the local stationary problem defines a differentiable implicit graph layer. To learn the graph itself, we optimize over the stratified moduli space of weighted graphs and equip each stratum with a non-degenerate K\"ahler--Hessian metric that keeps natural-gradient descent and face crossing well posed. We then show that a multilayer stationary network is equivalent to an exact global stationary problem on a supra-graph, and that it admits a penalized global relaxation whose stationary states converge to the exact one as the penalty parameter tends to infinity. Reverse-mode differentiation is recovered as the adjoint of the exact global system, and the penalized adjoint converges to it in the same limit. Finally, under finite-dimensional strong-monotonicity and admissible-lift assumptions, the corresponding represented hypothesis classes coincide among resolvent feed-forward networks, graph-stationary networks, supra-graph stationary systems, and sheaf-based architectures with unitary connection. The resulting structural identifications yield complexity bounds controlled by sparse graph or supra-graph geometry rather than dense ambient connectivity.
\end{abstract}

\section{Introduction}
Graph neural networks expose relational structure, while implicit and equilibrium models replace explicit layer updates by stationary equations or fixed points. This paper studies the regime in which both principles are imposed simultaneously: each hidden layer is defined by a stationary Schr\"odinger-type graph dynamics, and the graph itself is learned. The resulting theory must answer four questions. Does the local stationary dynamics define a legitimate implicit layer? Can graph learning remain well posed when edge creation and deletion move the iterate across a stratified parameter space? Can the multilayer architecture be represented globally rather than only as a sequential procedure? And, once directed or vector-valued extensions are admitted, which represented hypothesis classes coincide on the admissible regime?

The paper answers these questions by constructing one theorem-driven chain. First, the local Schr\"odinger-type dynamics yields a differentiable stationary graph layer on any stable branch. Second, graph learning is posed on the stratified moduli space of weighted graphs and regularized by a K\"ahler--Hessian metric that remains non-degenerate at faces, making natural-gradient descent and support changes geometrically well posed. Third, a multilayer stationary network is identified with an exact constrained global stationary problem on a supra-graph and with a penalized global relaxation whose stationary states converge to the exact one as the penalty parameter tends to infinity. Reverse-mode differentiation is recovered as the adjoint of the exact global system, while the penalized adjoint converges to it in the same limit. Fourth, under finite-dimensional strong-monotonicity and admissible-lift assumptions, the corresponding represented hypothesis classes coincide among resolvent-activation feed-forward networks, graph-stationary networks, supra-graph stationary systems, and sheaf-based architectures with unitary connection. The same sparse graph or supra-graph structure then controls the statistical complexity of the learned model. 

\paragraph{Contributions.}
\begin{enumerate}[leftmargin=1.2em]
\item We define a stationary Schr\"odinger-type graph layer and prove local well posedness, smooth dependence on graph parameters, and exponential convergence on a stable branch.
\item We formulate graph learning on a stratified moduli space of weighted graphs and prove fixed-stratum natural-gradient descent and finite face crossing. Under additional geometric assumptions we obtain support identification and metric recovery. A separate interventional consequence is recorded only as an optional regime in the appendix.
\item We prove that a multilayer stationary graph network is equivalent to an exact constrained global stationary system on a supra-graph, and that it admits a penalized global relaxation whose stationary states converge to the exact one as the penalty parameter tends to infinity. Reverse-mode differentiation coincides with the adjoint of the exact global linearization, and the penalized adjoint converges to it in the same limit.
\item We characterize the represented hypothesis class of the architecture across resolvent-activation feed-forward networks, graph-stationary networks, supra-graph stationary systems, and sheaf-based architectures under the admissible-lift assumptions, and we record the resulting structure-aware statistical consequences.
\end{enumerate}

\paragraph{Relation to prior work.}
The local stationary block extends physics-inspired and equilibrium models by making the latent graph itself part of the learned object \cite{haber2017stable,ruthotto2019pde,amos2017optnet,bai2019deq,winston2020monotone,elghaoui2021implicit}. The graph-learning component is related to latent-graph and structure-learning viewpoints in geometric deep learning \cite{bronstein2017geomdl,hamilton2017graphsage,zhou2020gnnreview}, but the optimization variable here is a stratified graph moduli space rather than a smooth Euclidean parametrization. The operator-theoretic and sheaf-theoretic conclusions connect the framework to resolvent-based implicit networks and recent sheaf neural architectures \cite{curry2014sheaves}. What is new here is the theorem chain linking stable stationary graph layers, stratified support dynamics, an exact supra-graph formulation together with a penalized global relaxation, represented equivalence, and structure-aware complexity control under one compatible regime of assumptions. Bloch geometry, geometric recovery, and causal recovery are then developed as consequences or extensions in the appendices. The detailed sheaf-lift constructions are also deferred to the appendices, while only their represented-equivalence consequence is used in the main text.

\section{Stationary Graph Layers and Standing Assumptions}
\label{sec:stationary-layer}
Let $G=(V,E,w)$ be a weighted graph, let $\psi\in\C^V\setminus\{0\}$, and define
\[
P_\psi^\perp = I - \frac{\psi\psi^\dagger}{\|\psi\|^2}.
\]
For an input state $\psi^0\in\C^V\setminus\{0\}$ and $\gamma>0$, consider the dissipative Schr\"odinger-type flow
\begin{equation}
\label{eq:main-sch}
\dot\psi = -\ii\big(\Delta(w)+\mathrm{diag}(|\psi^0|^2)\big)\psi-\gamma P_\psi^\perp\Big(\Delta(w)\psi+\mathrm{diag}(|\psi|^2-|\psi^0|^2)\psi\Big).
\end{equation}
The projector removes the component parallel to the current state, so the dissipative correction is tangent to the norm sphere.

\begin{Definition}[Stationary graph layer]
\label{def:main-layer}
Fix a parameter region on which \eqref{eq:main-sch} admits a unique exponentially stable equilibrium branch $w\mapsto\psi_{\mathrm s}(w)$. The associated {stationary graph layer} is the implicit map
\[
\mathcal S_G(w,\psi^0):=\psi_{\mathrm s}(w).
\]
\end{Definition}

\begin{Theorem}[Stable stationary graph layer]
\label{thm:main-layer}
Assume that for some $w_0\in\R_{>0}^E$ the flow \eqref{eq:main-sch} admits an isolated exponentially stable equilibrium $\psi_{\mathrm s}^0\neq 0$, and that the Jacobian with respect to $\psi$ at $(\psi_{\mathrm s}^0,w_0)$ is invertible. Then there exist neighborhoods $U\ni\psi_{\mathrm s}^0$ and $W\ni w_0$ such that:
\begin{enumerate}[leftmargin=1.2em]
\item for every $w\in W$ there is a unique equilibrium $\psi_{\mathrm s}(w)\in U$ and the map $w\mapsto\psi_{\mathrm s}(w)$ is $C^\infty$;
\item trajectories starting in $U$ converge exponentially to $\psi_{\mathrm s}(w)$ uniformly for $w\in W$; and
\item the stationary map $\mathcal S_G$ is a differentiable implicit layer on the stable branch.
\end{enumerate}
\end{Theorem}

\paragraph{Standing assumptions.}
Theorems in Sections~\ref{sec:stratified-learning}--\ref{sec:statistical} use the following assumptions.
\begin{enumerate}[label=\textbf{A\arabic*},leftmargin=1.2cm]
\item \textbf{Stable stationary branches.} On each active graph stratum under consideration, the stationary equation induced by \eqref{eq:main-sch} admits a unique exponentially stable branch with smooth dependence on the active graph parameters.
\item \textbf{Stratum-wise regularity and convexity.} On each active stratum $\M(E)$, the regularized objective is $C^{1,1}$ in the active edge weights and strongly convex on bounded active-weight boxes.
\item \textbf{Admissible stochastic support dynamics.} Mini-batch gradients are unbiased with bounded variance, the step-size and batch-growth schedules satisfy the Robbins--Monro and concentration conditions used in Appendix~\ref{app:kahler}, and activation/pruning decisions are made by measurable threshold rules depending only on the current iterate and sampled mini-batch. On terminal regimes where the sign-gap and concentration conditions of Appendix~\ref{app:recovery} hold, these rules eventually stabilize.
\item \textbf{Affine global couplings.} Inter-layer maps are affine and lie in compact parameter boxes on which all stationary branches remain in the regime of Assumption~\textbf{A1}.
\item \textbf{Finite-dimensional monotone realizations and admissible lifts.} The operator-theoretic and sheaf-lift constructions are finite-dimensional, generated by strongly monotone quadratic-plus-convex stationary operators or by unitary sheaf lifts satisfying the compatibility conditions stated in Appendix~\ref{app:equiv}, and remain on compact parameter boxes.
\item \textbf{Bounded loss and regular readout.} Readout maps are $C^{1,1}$ on the relevant compact sets and losses are bounded and Lipschitz.
\end{enumerate}
Additional assumptions are used only for optional regimes. Any appendix-local assumptions beyond the main-text standing assumptions \textbf{A1}--\textbf{A6} are introduced separately and are labeled by \textbf{R}-symbols.
\begin{enumerate}[label=\textbf{G\arabic*},leftmargin=1.2cm]
\item \textbf{Geometric recovery regime.} The sampled vertices form a $\delta$-net on a compact Riemannian manifold, the target support is the corresponding geometric neighborhood graph, and the realizability, margin, gradient-separation, and scale conditions of Appendix~\ref{app:recovery} hold uniformly on the terminal regime.
\item \textbf{Interventional causal regime.} The data are generated by a causally sufficient acyclic SCM satisfying Markov and faithfulness conditions, and intervention coverage together with the interventional gradient-separation conditions of Appendix~\ref{app:recovery} hold.
\end{enumerate}

\begin{table}[t]
\centering
\small
\caption{Assumption-to-result map for the main text.}
\label{tab:assumption-map}
\begin{tabularx}{\linewidth}{lYY}
\toprule
Result & Assumptions & Proof location \\
\midrule
Theorem~\ref{thm:main-layer} & local stability at one branch point & Appendix~\ref{app:local} \\
Theorem~\ref{thm:main-fixed} & A1--A3 & Appendix~\ref{app:kahler} \\
Theorem~\ref{thm:main-crossing} & A1--A3, together with the recovery-side threshold-stability conditions under the relevant R-assumptions & Appendix~\ref{app:kahler} + Appendix~\ref{app:recovery} \\
Theorem~\ref{thm:main-support}, Corollaries~\ref{cor:main-metric}--\ref{cor:main-homology} & A1--A3, G1 & Appendix~\ref{app:recovery} \\
Corollary~\ref{cor:main-causal} & A1--A3, G2 & Appendix~\ref{app:recovery} \\
Propositions~\ref{prop:main-global-regularity}--\ref{prop:main-global-unique}, Theorems~\ref{thm:main-global-exact}--\ref{thm:main-super}, including \ref{thm:main-supra-penalized} and \ref{thm:main-supra-adjoint} & A1, A4, A5 & Appendix~\ref{app:equiv} \\
Definition~\ref{def:main-complexity-profile}, Proposition~\ref{prop:main-complexity}, Theorem~\ref{thm:main-stat} & A1--A6 & Appendix~\ref{app:gen} \\
Corollary~\ref{cor:main-sparse-regimes} & A1--A6 together with G1/G2 and sparse-regime assumptions & Appendix~\ref{app:gen} \\
\bottomrule
\end{tabularx}
\end{table}

\section{Learning on a Stratified K\"ahler--Hessian Moduli Space}
\label{sec:stratified-learning}
For a fixed vertex set $V$, the weighted graph variable ranges over the stratified moduli space
\[
\M=\bigsqcup_{E\subseteq V\times V}\M(E),\qquad \M(E)\cong\R_{>0}^{|E|},
\]
where each stratum corresponds to a fixed active edge set and face crossing corresponds to edge activation or deletion. Let
\begin{equation}
\label{eq:main-loss}
\mathcal L(E,w)=\E_{(X,y)\sim\D}\!\left[\mathcal L_{\mathrm{sample}}(X,y;(E,w))\right]+\frac{\mu_2}{2}\|w(E)\|_2^2+\mu_1\|w(E)\|_1,
\end{equation}
where the sample loss is induced by the stationary graph layer. To regularize optimization at the faces we choose a smooth positive profile $m_\delta$ with $m_\delta(t)=1/t^2$ for $t\ge 2\delta$ and bounded positive continuation near $t=0$, and equip each stratum with the Hessian metric
\begin{equation}
\label{eq:main-metric}
g_\delta(w)=\mathrm{diag}\big(m_\delta(w_e)\big)_{e\in E}.
\end{equation}
The associated natural gradient is $\grad_{g_\delta}f=g_\delta^{-1}\nabla f$.

\begin{Definition}[Stratified natural-gradient update]
\label{def:main-ng}
On a current stratum $\M(E_t)$, the natural-gradient update of a differentiable objective $f$ is
\[
\widetilde w^{(t+1)}=w^{(t)}-\alpha_t g_\delta(w^{(t)})^{-1}\nabla f(w^{(t)}),
\]
followed by projection onto the nonnegative orthant and measurable activation/pruning decisions along faces according to Assumption~\textbf{A3}.
\end{Definition}

\subsection{Optimization on the graph moduli}

\begin{Theorem}[Fixed-stratum natural-gradient descent]
\label{thm:main-fixed}
Assume \textbf{A1}--\textbf{A3}. On any fixed active stratum $\M(E)$, the K\"ahler--Hessian metric \eqref{eq:main-metric} is uniformly positive definite up to the faces. If the restriction of the objective to $\M(E)$ is geodesically $L$-regular and $\alpha_t\in(0,2/L)$, then the natural-gradient iterates satisfy
\[
\textstyle \mathcal L(w^{(t+1)})\le \mathcal L(w^{(t)})-\frac{\alpha_t}{2}\|\grad_{g_\delta}\mathcal L(w^{(t)})\|_{g_\delta}^2,
\qquad
\sum_t \alpha_t\,\|\grad_{g_\delta}\mathcal L(w^{(t)})\|_{g_\delta}^2<\infty.
\]
If, in addition, the restriction is geodesically PL on a terminal neighborhood, then convergence on a terminal stratum is linear; with stochastic gradients of variance $O(B_t^{-1})$ and $\alpha_t=\alpha_0 t^{-1/2}$, the corresponding gradient norm admits the standard $O(t^{-1/2})$ rate stated in Appendix~\ref{app:kahler}.
\end{Theorem}

\begin{Theorem}[Finite face crossing and terminal convergence]
\label{thm:main-crossing}
Assume \textbf{A1}--\textbf{A3}. The metric $g_\delta$ assigns finite distance to every face of the stratified moduli space, and the stratified natural-gradient dynamics are therefore well posed across face crossings. If, in addition, the threshold-stability conditions of Appendix~\ref{app:recovery} hold on the visited support regimes, then the support process performs only finitely many face crossings almost surely and enters a terminal stratum after finitely many support changes. On that terminal stratum the iterates satisfy the fixed-stratum convergence guarantees of Theorem~\ref{thm:main-fixed}.
\end{Theorem}

\subsection{Recovery consequences under \textbf{G1}}

These results specialize the terminal-stratum dynamics to the additional geometric realizability and separation assumptions of \textbf{G1}.

\begin{Theorem}[Support identification under geometric realizability]
\label{thm:main-support}
Assume \textbf{A1}--\textbf{A3} and \textbf{G1}. Then for every $\varepsilon\in(0,1)$ there exist thresholds, step-size schedules, batch-size growth, and a finite time $T_0$ such that
\[
\mathbb P\!\left(E_t = E_{\mathrm{true}}\ \ \forall t\ge T_0\right)\ \ge\ 1-\varepsilon.
\]
On the terminal stratum the active weights converge to the unique geometric minimizer.
\end{Theorem}

\begin{Corollary}[Metric recovery on the terminal support]
\label{cor:main-metric}
Under Theorem~\ref{thm:main-support}, let $d_{G_t}$ denote the shortest-path metric induced by the learned edge lengths $\ell_t(e)=1/w_t(e)$. Then, on the event of support identification,
\[
d_{\mathrm{GH}}\big((V,d_{G_t}),\mathcal G\big)
\le C_1\delta + C_2\|w_t-w^\star\|_\infty,
\]
and therefore the learned graph metric converges to the latent manifold metric up to the discretization error $O(\delta)$ and the terminal optimization error.
\end{Corollary}

\begin{Corollary}[Homology consistency of the recovered clique complex]
\label{cor:main-homology}
Under Assumptions~\textbf{A1}--\textbf{A3} and \textbf{G1}, assume moreover that the geometric scale parameters lie in the regime of Appendix~\ref{app:recovery}. Then, for all sufficiently large $t$, the clique complex of the learned terminal graph coincides with the corresponding recovered Vietoris--Rips complex and hence recovers $H_0$ and $H_1$ of the latent manifold. In particular, the connected components and first Betti number of the recovered clique complex agree with those of the underlying geometric support.
\end{Corollary}

\subsection{Optional interventional extension under \textbf{G2}}

This subsection is not used elsewhere in the main text. It records the separate interventional consequence obtained when the observational data are generated by a causally sufficient SCM with intervention coverage.

\begin{Corollary}[Interventional causal consequence]
\label{cor:main-causal}
Assume \textbf{A1}--\textbf{A3} and \textbf{G2}. Then the same graph-learning mechanism identifies the CPDAG skeleton and the compelled orientations under the interventional gradient-separation conditions stated in Appendix~\ref{app:recovery}.
\end{Corollary}

\section{Global Stationary Formulations on the Supra-Graph}
\label{sec:supra}
Let the layer states be $\psi^\ell\in\C^{n_\ell}$ and let the inter-layer signal variables be $q_\ell\in\C^{m_\ell}$ for $\ell\ge 2$, with external input $q_1$ fixed by the data. The affine signal flow is
\[
q_{\ell+1}=A_\ell\psi^\ell+c_\ell.
\]
Each layer is governed by a stationary operator $F_\ell(\psi^\ell;w_\ell,q_\ell)=0$ induced by the local stationary graph layer.

\begin{Definition}[Global stationary operators and formulations]
\label{def:main-global-operators}
Define the augmented global state space
\[
\textstyle \mathcal X:=\Big(\prod_{\ell=1}^L \C^{n_\ell}\Big)\times \Big(\prod_{\ell=2}^L \C^{m_\ell}\Big),
\qquad
\Xi:=(\psi^1,\dots,\psi^L,q_2,\dots,q_L)\in\mathcal X.
\]
The {exact global operator} is the map $\mathfrak F_{\mathrm{ex}}:\mathcal X\to\mathcal X$ defined by
\[
\textstyle \mathfrak F_{\mathrm{ex}}(\Xi):=
\Big(F_1(\psi^1;w_1,q_1),\dots,F_L(\psi^L;w_L,q_L),
q_2-A_1\psi^1-c_1,\dots,q_L-A_{L-1}\psi^{L-1}-c_{L-1}\Big).
\]
Let
\[
\mathfrak F_0^\sharp(\Xi):=
\big(F_1(\psi^1;w_1,q_1),\dots,F_L(\psi^L;w_L,q_L),0,\dots,0\big)\in\mathcal X
\]
be the uncoupled block operator, and let
\[
\textstyle C:\mathcal X\to \prod_{\ell=2}^L \C^{m_\ell},
\qquad
C\Xi:=\big(q_2-A_1\psi^1-c_1,\dots,q_L-A_{L-1}\psi^{L-1}-c_{L-1}\big)
\]
be the linear coupling residual operator. The {penalized global operator} is
\[
\mathfrak F_\tau(\Xi):=\mathfrak F_0^\sharp(\Xi)+\tau\,C^\ast C\Xi,
\qquad \tau>0.
\]
For a scalar loss $\mathcal J(\Xi)$ depending on the terminal state, the {exact global adjoint} and the {penalized global adjoint} are the linearized adjoint systems
\[
D\mathfrak F_{\mathrm{ex}}(\Xi)^\ast \Lambda_{\mathrm{ex}}=\nabla_\Xi \mathcal J(\Xi),
\qquad
D\mathfrak F_\tau(\Xi)^\ast \Lambda_\tau=\nabla_\Xi \mathcal J(\Xi).
\]
Accordingly, the {exact supra-graph formulation} is the root problem $\mathfrak F_{\mathrm{ex}}(\Xi)=0$, the {penalized supra-graph formulation} is the root problem $\mathfrak F_\tau(\Xi)=0$, and the corresponding global adjoint formulations are the dual linear systems associated with their linearizations.
\end{Definition}

\begin{Proposition}[Regularity of the global stationary operators]
\label{prop:main-global-regularity}
Under Assumptions~\textbf{A1}, \textbf{A4}, and \textbf{A5}, the exact operator $\mathfrak F_{\mathrm{ex}}$ and the penalized operators $\mathfrak F_\tau$ are well defined on the common augmented state space $\mathcal X$. Their local linearizations are bounded on compact parameter boxes. On every stable branch, $D\mathfrak F_{\mathrm{ex}}(\Xi)$ is invertible; for each fixed $\tau>0$, the penalized linearization $D\mathfrak F_\tau(\Xi)$ is invertible on the same compact regime whenever the layerwise stationary Jacobians remain uniformly invertible there. Consequently the exact stationary map and the penalized stationary maps, together with their adjoint systems, are differentiable on the stable regime.
\end{Proposition}

\begin{Proposition}[Strong monotonicity implies uniqueness of the global stationary state]
\label{prop:main-global-unique}
Assume \textbf{A1}, \textbf{A4}, and \textbf{A5}. If the layerwise stationary operators are strongly monotone and the affine couplings remain inside the compact regime of Assumption~\textbf{A4}, then the exact global operator admits a unique stationary state on $\mathcal X$, and for every $\tau>0$ the penalized global operator admits a unique stationary state on $\mathcal X$.
\end{Proposition}

\begin{Theorem}[Exact global stationary equivalence]
\label{thm:main-global-exact}
Assume \textbf{A1}, \textbf{A4}, and \textbf{A5}. The sequential composition of stationary graph layers with affine inter-layer couplings and the exact supra-graph formulation of Definition~\ref{def:main-global-operators} represent the same global stationary state. Equivalently, the feed-forward stationary network is identical to the unique solution of the exact global operator equation $\mathfrak F_{\mathrm{ex}}(\Xi)=0$.
\end{Theorem}

\begin{Theorem}[Penalized supra-graph consistency]
\label{thm:main-supra-penalized}
Assume \textbf{A1}, \textbf{A4}, and \textbf{A5}. Let $\Xi_{\mathrm{ex}}$ be the unique exact stationary state solving $\mathfrak F_{\mathrm{ex}}(\Xi)=0$. For each $\tau>0$, let $\Xi_\tau$ be the unique penalized stationary state solving $\mathfrak F_\tau(\Xi)=0$. Then
\[
\|C\Xi_\tau\|\to 0
\qquad\text{and}\qquad
\Xi_\tau\to \Xi_{\mathrm{ex}}
\quad\text{as }\tau\to\infty.
\]
In particular, the penalized supra-graph formulation recovers the same global stationary state only in the limit $\tau\to\infty$.
\end{Theorem}

\begin{Theorem}[Global adjoint and reverse-mode differentiation]
\label{thm:main-supra-adjoint}
Assume \textbf{A1}, \textbf{A4}, and \textbf{A5}. For any smooth scalar loss depending on the terminal state, the gradients produced by reverse-mode differentiation through the sequential stationary network coincide with those produced by the exact global adjoint system associated with $\mathfrak F_{\mathrm{ex}}$. If $\Lambda_\tau$ denotes the penalized global adjoint at the penalized stationary state $\Xi_\tau$, then
\[
\Lambda_\tau \to \Lambda_{\mathrm{ex}}
\quad\text{as }\tau\to\infty,
\]
where $\Lambda_{\mathrm{ex}}$ is the exact global adjoint at $\Xi_{\mathrm{ex}}$.
\end{Theorem}

\begin{Corollary}[Depth--width duality through the supra-graph]
\label{cor:main-depthwidth}
Under Theorems~\ref{thm:main-global-exact}, \ref{thm:main-supra-penalized}, and \ref{thm:main-supra-adjoint}, a multilayer stationary graph architecture can be viewed exactly either as a depth-wise composition of local stationary layers or as a single exact stationary problem on a wider supra-graph carrying all intra-layer and inter-layer couplings. The penalized supra-graph formulation provides a global relaxation whose stationary states and adjoints converge to the exact ones as $\tau\to\infty$.
\end{Corollary}

\section{Represented Architectural Equivalence}
\label{sec:equivalence}

\begin{Definition}[Architectural classes]
\label{def:main-classes}
We use the following finite-dimensional classes.
\begin{enumerate}[leftmargin=1.2em]
\item $\mathsf{FFNN}_{\mathrm{prox}}$: feed-forward networks whose activations are proximal maps.
\item $\mathsf{FFNN}_{\mathrm{res}}$: feed-forward networks whose activations are resolvents of strongly monotone operators.
\item $\mathsf{FFGN}$: layered stationary graph networks built from local graph energies.
\item $\mathsf{SGN}$: global stationary systems on a supra-graph.
\item $\mathsf{SFFN}$: sheaf-based stationary architectures with unitary connection.
\end{enumerate}
\end{Definition}

\begin{Definition}[Represented hypothesis class and represented equivalence]
\label{def:main-represented}
For an architectural class $\mathcal C$, write $\mathcal H(\mathcal C)$ for the set of input--output maps represented by members of $\mathcal C$, modulo admissible linear state changes, finite-dimensional lifts, and gauge identifications that preserve the represented map. Two classes are {represented-equivalent} if they have the same represented hypothesis class.
\end{Definition}

\begin{Proposition}[Graph-stationary layers are resolvent layers]
\label{prop:main-resolvent}
Under Assumption~\textbf{A5}, every graph-stationary layer is a resolvent layer of the form
\[
\psi=(I+L+\partial\Phi)^{-1}(u).
\]
Conversely, every finite-dimensional resolvent activation generated by a strongly monotone operator of admissible sparse quadratic-plus-convex form can be represented as a graph-stationary layer after choosing the corresponding quadratic form and convex potential on the same finite-dimensional state space.
\end{Proposition}

\begin{Theorem}[Represented equivalence of layered graph-stationary and supra-graph systems]
\label{thm:main-ffgn-sgn}
Under Assumptions~\textbf{A1}, \textbf{A4}, and \textbf{A5}, the represented hypothesis classes of layered graph-stationary networks and exact supra-graph stationary systems coincide:
\[
\mathcal H(\mathsf{FFGN})=\mathcal H(\mathsf{SGN}).
\]
Moreover, the exact constrained formulation has the same represented hypothesis class as the layered stationary architecture. The penalized supra-graph family approaches the same represented map in the limit $\tau\to\infty$ on the stable regime of Theorems~\ref{thm:main-supra-penalized} and \ref{thm:main-supra-adjoint}. The gradients obtained from the exact global adjoint agree with those obtained by reverse-mode differentiation through the layered stationary architecture.
\end{Theorem}

\begin{Theorem}[Directed-to-undirected reduction through diagonal and sheaf lifts]
\label{thm:main-directed-reduction}
Under Assumption~\textbf{A5}, every orientable directed stationary layer is diagonally conjugate to an undirected stationary layer. More generally, every directed or vector-valued stationary layer produced by the admissible sheaf-lift constructions of Appendix~\ref{app:equiv} admits an equivalent realization as an undirected sheaf layer with unitary connection on an enlarged state space, and therefore as a graph-stationary layer after an isometric lift.
\end{Theorem}

\begin{Theorem}[Equivalence of represented hypothesis classes under admissible lifts]
\label{thm:main-super}
Under Assumptions~\textbf{A1}, \textbf{A4}, and \textbf{A5}, together with the finite-dimensional and admissible-lift conditions of Appendix~\ref{app:equiv}, the represented hypothesis classes satisfy
\[
\mathcal H(\mathsf{FFNN}_{\mathrm{prox}})\subseteq \mathcal H(\mathsf{FFNN}_{\mathrm{res}}),
\qquad
\mathcal H(\mathsf{FFNN}_{\mathrm{res}})=\mathcal H(\mathsf{FFGN})=\mathcal H(\mathsf{SGN})=\mathcal H(\mathsf{SFFN}).
\]
Here the equality $\mathcal H(\mathsf{FFGN})=\mathcal H(\mathsf{SGN})$ is the represented equivalence of Theorem~\ref{thm:main-ffgn-sgn}, the identification $\mathcal H(\mathsf{FFNN}_{\mathrm{res}})=\mathcal H(\mathsf{FFGN})$ follows from Proposition~\ref{prop:main-resolvent}, and the passage to directed or vector-valued architectures is governed by Theorem~\ref{thm:main-directed-reduction}. Thus these four architectural classes have the same represented hypothesis class on the admissible strongly monotone regime.
\end{Theorem}

\begin{Remark}[Scope of represented equivalence]
\label{rem:main-scope}
Theorem~\ref{thm:main-super} is a representational statement. It identifies the same represented hypothesis class under finite-dimensional, strongly monotone, and admissible-lift assumptions. It does not claim identical optimization landscapes for all native parametrizations, nor does it cover non-orientable directed operators or branches on which uniqueness or regularity fails.
\end{Remark}

\begin{Definition}[Sparse-support quadratic interaction family]
\label{def:main-sparse-quadratic}
Fix layer widths and a support family $(E_\ell)_\ell$. The associated {sparse-support quadratic interaction family} consists of those layerwise quadratic forms whose cross-coordinate interactions vanish off the prescribed supports $E_\ell$ and whose remaining affine couplings are held fixed.
\end{Definition}

\begin{Corollary}[Parameter compactness under sparse graph geometry]
\label{cor:main-compactness}
Fix the sparse-support quadratic interaction family of Definition~\ref{def:main-sparse-quadratic}. If the stationary graph or sheaf realizations are restricted to separable convex potentials with $O(n_\ell)$ parameters, then they require
\[
\textstyle O\!\left(\sum_\ell |E_\ell| + \dim(A_\ell,B_\ell) + n_\ell\right)
\]
trainable degrees of freedom. Any dense feed-forward realization of the same sparse-support quadratic interaction family requires $\Omega(n_\ell^2)$ parameters at layer $\ell$ unless the same support restriction is imposed explicitly in the dense parametrization.
\end{Corollary}

\section{Structure-Aware Statistical Consequences}
\label{sec:statistical}
The same sparse graph or supra-graph structure that drives the representation also controls statistical complexity. Appendix~\ref{app:gen} develops the full PAC--Bayes, stability, and Rademacher bounds. The main text records the structural drivers that survive after these bounds are specialized to the learned graph or supra-graph regime.

\begin{Definition}[Structural complexity profile]
\label{def:main-complexity-profile}
For a learned graph or supra-graph realization, define its {structural complexity profile} by the triple
\[
\mathfrak c=(s,\deg_{\max},p),
\]
where $s$ is the number of active graph or supra-graph interactions, $\deg_{\max}$ is the maximal active degree, and $p$ is the number of remaining non-structural parameters. The statistical bounds of Appendix~\ref{app:gen} are expressed in terms of this profile rather than the dense ambient interaction count.
\end{Definition}

\begin{Proposition}[Degree and support control of effective complexity]
\label{prop:main-complexity}
Let $G_T=(V,E_T,w_T)$ be a learned graph on a terminal stratum. Then the weighted Laplacian satisfies
\[
\|\Delta(w_T)\|\le (1+\|w_T\|_\infty)\deg_{\max}(G_T).
\]
Under the stationary stability assumptions of Appendix~\ref{app:gen}, the effective Lipschitz constant of the associated stationary layer is therefore controlled by the maximal degree and the stability gap. In the global supra-graph formulation, the same conclusion holds with $G_T$ replaced by the active supra-graph $\mathbb G_T$.
\end{Proposition}

\begin{Theorem}[Structure-aware generalization bound]
\label{thm:main-stat}
Under Assumptions~\textbf{A1}--\textbf{A6}, the results of Appendix~\ref{app:gen} imply that the learned graph or supra-graph model returned by training satisfies
\[
\textstyle R(f_T)\le \widehat R_S(f_T)
 + C_{\mathrm{PB}}\sqrt{\frac{s\log(eP/s)+\log(1/\delta)}{M}}
 + C_{\mathrm{stab}}\frac{\deg_{\max}}{M}
 + C_{\mathrm{Rad}}\sqrt{\frac{p+s}{M}},
\]
where $(s,\deg_{\max},p)$ is the structural complexity profile of Definition~\ref{def:main-complexity-profile}, $P$ is the ambient number of potential interactions, and $p$ counts the remaining non-structural parameters.
\end{Theorem}

\begin{Corollary}[Sparse-geometric and sparse-causal improvement]
\label{cor:main-sparse-regimes}
Under Theorem~\ref{thm:main-stat}, if the geometric regime of Corollary~\ref{cor:main-metric} implies $|E_T|=O(N)$ and $\deg_{\max}(G_T)=O(1)$, then the leading structural terms scale as
\[
\textstyle\sqrt{\frac{N\log N}{M}}+\sqrt{\frac{p+N}{M}}+O(M^{-1}),
\]
whereas dense graph or dense attention baselines under the same ambient interaction model incur $\Theta(N^2)$ support terms. Under the sparse-causal regime of Corollary~\ref{cor:main-causal}, together with an explicit sparse-skeleton assumption $|E_T|=O(d)$, the same replacement holds with $N$ replaced by $d$ and active support size $O(d)$ rather than $\Theta(d^2)$.
\end{Corollary}

\begin{table}[t]
\centering
\small
\caption{Leading complexity drivers under fixed parameter boxes. Precise constants and full bounds are given in Appendix~\ref{app:gen}.}
\label{tab:main-generalization}
\begin{tabularx}{\linewidth}{lYYY}
\toprule
Model class & PAC--Bayes code & Stability & Rademacher \\
\midrule
Dense graph / dense attention & ambient support size $\Theta(N^2)$ & width or dense degree & all pairwise interactions \\
Learned graph (ours) & active edges $|E_T|$ & $\deg_{\max}(G_T)$ & active interactions $|E_T|$ \\
Supra-graph (ours) & active supra-edges $|E_{\mathrm{sup}}|$ & $\deg_{\max}(\mathbb G_T)$ & active supra-interactions \\
\bottomrule
\end{tabularx}
\end{table}

\section{Experiments}

The code of the experiments is available at: \url{https://anonymous.4open.science/r/Learning-Latent-Graph-Geometry-2FDA/}.

\paragraph{Numerical consistency of the exact supra-graph formulation.}
We compare two evaluations of the same two-layer strongly convex stationary model: sequential layerwise stationary solves and one exact global supra-graph solve. Both layers use quadratic strongly convex local potentials, so the stationary states are uniquely defined and the comparison is performed in a clean monotone regime. We evaluate both realizations on $B=256$ synthetic inputs in double precision, with input dimension $d=6$, hidden dimensions $(n_1,n_2)=(14,11)$, and $8$ random seeds. For each seed we measure the mean and maximum relative output discrepancy, the relative loss discrepancy, and the relative discrepancy between parameter gradients computed from the sequential and global formulations (Table~\ref{tab:supra_equiv_sanity}).

\begin{table}[ht!]
\centering
\small
\begin{tabular}{rccccc}
\toprule
Seed & Mean rel. output & Max rel. output & Rel. loss & Total rel. grad. & Max param. rel. grad. \\
\midrule
0 & 4.142e-16 & 2.355e-14 & 9.600e-16 & 1.987e-14 & 2.181e-14 \\
1 & 1.350e-15 & 6.032e-14 & 4.255e-16 & 8.037e-15 & 1.246e-14 \\
2 & 1.564e-15 & 2.258e-14 & 0.000e+00 & 2.296e-14 & 4.366e-14 \\
3 & 6.735e-16 & 1.585e-14 & 2.020e-15 & 5.037e-14 & 9.127e-14 \\
4 & 8.402e-16 & 6.905e-14 & 4.165e-16 & 2.595e-14 & 1.268e-13 \\
5 & 1.844e-15 & 4.277e-14 & 6.521e-16 & 6.996e-14 & 9.876e-14 \\
6 & 4.571e-16 & 1.462e-14 & 1.355e-16 & 1.034e-14 & 1.437e-14 \\
7 & 5.860e-15 & 1.257e-12 & 1.288e-16 & 1.994e-14 & 4.805e-13 \\
\midrule
Mean over seeds & 1.625e-15 & 1.883e-13 & 5.923e-16 & 2.843e-14 & 1.112e-13 \\
Worst seed & 5.860e-15 & 1.257e-12 & 2.020e-15 & 6.996e-14 & 4.805e-13 \\
\bottomrule
\end{tabular}
\caption{Numerical consistency for the exact supra-graph formulation in a strongly convex two-layer stationary regime. Sequential layerwise evaluation and a single global block solve agree up to machine precision in outputs, losses, and parameter gradients. Computations use double precision with batch size $B=256$, input dimension $d=6$, and hidden dimensions $(n_1,n_2)=(14,11)$.}
\label{tab:supra_equiv_sanity}
\end{table}

The two realizations agree up to floating-point precision. Averaged over the $8$ seeds, the mean relative output error is $1.625\times 10^{-15}$, the mean maximum relative output error is $1.883\times 10^{-13}$, the mean relative loss error is $5.923\times 10^{-16}$, and the mean total relative gradient error is $2.843\times 10^{-14}$. This provides a compact numerical consistency check for the exact supra-graph formulation in the strongly convex regime.

\paragraph{Geometric graph adaptation on a ring reconstruction task.}
We include a small synthetic illustration of graph adaptation in a controlled geometric regime. The data are arranged on a ring, and the target signal is generated from a stationary reconstruction problem on a graph whose support contains both local ring edges and a small number of additional second-neighbor interactions. At training time, the model is initialized from the local ring graph and is allowed to update edge weights together with support through thresholded activation and pruning over a candidate pool consisting of first- and second-neighbor edges. We compare this learned sparse graph against a fixed local graph and a dense candidate graph, using masked-node reconstruction error as the predictive metric and edge-level overlap with the planted graph as a structural metric.

\begin{table}[ht!]
\centering
\small
\begin{tabular}{lcccc}
\toprule
Model & Test MSE & Active edges & Two-hop F1 & Wins \\
\midrule
Fixed local graph & 0.0842 & 36.0 & 0.0000 & -- \\
Dense candidate graph & 0.0772 & 72.0 & 1.0000 & -- \\
Learned sparse graph & 0.0742 & 41.0000 & 0.2398 & 8/8 vs fixed, 4/8 vs dense \\
\bottomrule
\end{tabular}
\caption{The learned sparse graph starts from the 1-hop ring
and may activate/prune 2-hop edges through thresholded gradient dynamics. ``Two-hop F1'' denotes
edge-level F1 computed only on the planted second-neighbor interactions.}
\label{tab:ring-partial-completion}
\end{table}

Averaged over 8 random seeds, the learned sparse graph improves the mean test reconstruction error
over the fixed local graph (0.0742 vs.\ 0.0842) while remaining competitive with the dense candidate
graph (0.0772). At the structural level, the learned support partially recovers the planted second-neighbor
interactions: the mean two-hop F1 reported in Table~\ref{tab:ring-partial-completion} is 0.2398, while the model uses on average
41 active edges and stabilizes after a modest number of support updates. We view this experiment as
a controlled geometric illustration that the proposed support dynamics can adapt a local initial graph
toward a more informative sparse latent structure.

\section{Discussion}
\label{sec:discussion}
Under Assumptions~\textbf{A1}--\textbf{A5}, the core contribution of the paper is a single theorem chain linking four statements: stable local stationary graph layers, stratified graph learning with support changes, an exact supra-graph formulation together with a penalized global relaxation and a global adjoint interpretation, and represented equivalence across several architectural realizations. The geometric and causal recovery results require the stronger auxiliary regimes \textbf{G1} and \textbf{G2} and should be read as consequences rather than as part of the minimal core theory. The main limitation is therefore regime dependence: once stable branches, admissible lifts, orientability, or the additional recovery assumptions fail, the corresponding consequence theorems need not persist. In particular, the paper proves represented equivalence of realized hypothesis classes in the admissible strongly monotone regime, not identity of native parametrizations or optimization landscapes.

\bibliography{main}

\clearpage
\appendix

\section*{Appendix overview}
Each appendix begins with explicit proof anchors for the main-text results proved there. The appendices contain both core proofs and optional consequences. The central theorem chain of the main text relies on Appendix~\ref{app:local}, Appendix~\ref{app:kahler}, the supra-graph/adjoint part of Appendix~\ref{app:equiv}, and the structural bound in Appendix~\ref{app:gen}. Bloch geometry, geometric recovery, causal recovery, and sheaf-lift extensions are auxiliary consequences recorded separately. Appendix~\ref{app:local} proves Theorem~\ref{thm:main-layer}. Appendix~\ref{app:kahler} proves the fixed-stratum descent statement of Theorem~\ref{thm:main-fixed} together with the geometric face-crossing part of Theorem~\ref{thm:main-crossing}; the threshold-stability conditions used in the terminal-stabilization clause of Theorem~\ref{thm:main-crossing} are supplied in Appendix~\ref{app:recovery}. Appendix~\ref{app:recovery} proves Theorem~\ref{thm:main-support} together with Corollaries~\ref{cor:main-metric}, \ref{cor:main-homology}, and \ref{cor:main-causal}. Appendix~\ref{app:equiv} proves Propositions~\ref{prop:main-global-regularity} and \ref{prop:main-global-unique}, Theorems~\ref{thm:main-global-exact}, \ref{thm:main-supra-penalized}, \ref{thm:main-supra-adjoint}, \ref{thm:main-ffgn-sgn}, \ref{thm:main-directed-reduction}, and \ref{thm:main-super}, as well as Corollaries~\ref{cor:main-depthwidth} and \ref{cor:main-compactness}. Appendix~\ref{app:gen} proves Definition~\ref{def:main-complexity-profile} in context, Proposition~\ref{prop:main-complexity}, Theorem~\ref{thm:main-stat}, and Corollary~\ref{cor:main-sparse-regimes}.

\section{Proofs for Section~\ref{sec:stationary-layer}: local dynamics and Bloch geometry}\label{app:local}
\paragraph{Proof traceability.} This appendix proves Theorem~\ref{thm:main-layer}.

\subsection*{Proof of Theorem~\ref{thm:main-layer}}
The proof uses only the local stationary analysis stated below. The Bloch/Landau--Lifshitz reduction is recorded as a geometric interpretation of the stationary dynamics and is not used in the later optimization, equivalence, or statistical arguments.

\subsection{Schr\"odinger-type Activation}
\paragraph{Model and notation.}
Let $G=(V,E)$ be a finite graph with $|V|=N$ and nonnegative edge weights $w:E\to\R_{>0}$. We identify state vectors with $\C^{V}$, and write $\|\cdot\|$ for any fixed norm on $\C^{V}$ (all norms are equivalent in finite dimension). For $\psi\in\C^{V}\setminus\{0\}$ set the orthogonal projector
\[
P_\psi^\perp \;=\; I - \frac{\psi\psi^\dagger}{\|\psi\|^2}.
\]
The weighted graph Laplacian $\Delta(w):\C^{V}\to \C^{V}$ is linear in $w$ and given componentwise by
\[
(\Delta(w)\psi)_i \;=\; \sum_{(i,j)\in E} w(i,j)\,(\psi_i-\psi_j).
\]
Fix an initial vector $\psi^0\in\C^{V}\setminus\{0\}$ and a dissipation parameter $\gamma>0$. We consider the ODE on $\C^{V}\setminus\{0\}$
\begin{gather}\label{eq:sch}
\nonumber\frac{d\psi}{dt} \;=\; F(\psi,w),\\
F(\psi,w) \;=\; -\mathrm{i}\,\big(\Delta(w)+\mathrm{diag}(|\psi^0|^2)\big)\psi \;\; -\; \gamma\,P_\psi^\perp\Big(\Delta(w)\psi + \mathrm{diag}(|\psi|^2-|\psi^0|^2)\psi\Big),
\end{gather}
which combines a linear Hamiltonian part with a norm-preserving nonlinear dissipative correction. The presence of $P_\psi^\perp$ removes the component of the dissipative force parallel to $\psi$, hence $\frac{d}{dt}\|\psi\|^2=0$ along solutions.

\medskip
\noindent
\textbf{Equilibria and stability.}
An equilibrium $\psi_{\mathrm s}\neq 0$ for a fixed $w$ solves $F(\psi_{\mathrm s},w)=0$. We assume that for some $w_0\in\R_{>0}^{E}$ there exists an isolated exponentially stable equilibrium $\psi_{\mathrm s}^0\neq 0$; i.e., the Jacobian
\[
A_0 \;:=\; D_\psi F(\psi_{\mathrm s}^0,w_0)
\]
is Hurwitz: $\max\{\Re\lambda:\lambda\in\sigma(A_0)\}\le -\alpha<0$.

\begin{Theorem}[Existence of the limit and $C^\infty$ dependence on $w$]\label{th:sm_sch}
There exist neighborhoods $W\ni w_0$ and $U\ni\psi_{\mathrm s}^0$ with the following properties:
\begin{enumerate}
\item For every $w\in W$ there exists a unique equilibrium $\psi_{\mathrm s}(w)\in U$ with $F(\psi_{\mathrm s}(w),w)=0$, and the map $w\mapsto \psi_{\mathrm s}(w)$ is $C^\infty$.
\item There exist constants $C,\beta>0$ such that for all $w\in W$ and all solutions of \eqref{eq:sch} with $\psi(0)\in U$ one has
\[
\|\psi(t)-\psi_{\mathrm s}(w)\|\le C e^{-\beta t}\,\|\psi(0)-\psi_{\mathrm s}(w)\|\qquad\forall t\ge 0.
\]
In particular, $\lim_{t\to\infty}\psi(t;\psi^0,w)=\psi_{\mathrm s}(w)$ for every $\psi^0\in U$ and every $w\in W$.
\end{enumerate}
\end{Theorem}

\begin{proof}
Since $\Delta(w)$ is linear in $w$, $P^\perp_\psi$ is analytic for $\psi\neq 0$, and $|\psi|^2\psi$ is polynomial, the map $(\psi,w)\mapsto F(\psi,w)$ is $C^\infty$ on $(\C^{V}\setminus\{0\})\times \R_{>0}^{E}$. The implicit function theorem (e.g., \cite{KrantzParks2002,Kato1976}) applied to $F(\cdot,\cdot)=0$ at $(\psi_{\mathrm s}^0,w_0)$ and the invertibility of $A_0$ yield Item~1. Continuity of the spectrum implies that $A(w):=D_\psi F(\psi_{\mathrm s}(w),w)$ remains uniformly Hurwitz for $w$ in a smaller neighborhood $W$, hence there exists a positive definite $P(w)$ solving the Lyapunov equation \cite{Khalil2002} $A(w)^\dagger P(w)+P(w)A(w)=-I$ with uniform bounds $mI\preceq P(w)\preceq MI$. Writing the dynamics in deviation $z=\psi-\psi_{\mathrm s}(w)$ gives $\dot z=A(w)z+R(z,w)$ with $R=\mathcal{O}(\|z\|^2)$; the standard Lyapunov estimate for $V(z)=z^\dagger P(w)z$ then yields the exponential bound in Item~2 for sufficiently small $\|z(0)\|$, possibly after shrinking $U$ and $W$.
\end{proof}

The next result records the (trivial) smooth dependence on the initial condition inside the basin of the stable equilibrium; we state it in a way convenient for later use.

\begin{Theorem}[$C^\infty$ dependence on the initial condition]\label{th:sm_in_sch}
Fix $w\in W$ from Theorem~\ref{th:sm_sch} and let $\psi_{\mathrm s}=\psi_{\mathrm s}(w)$. There exist $U\subset\C^{V}$ and $C,\beta>0$ such that for all $\psi^0\in U$ the solution of \eqref{eq:sch} exists globally and
\[
\|\psi(t;\psi^0,w)-\psi_{\mathrm s}\|\le C e^{-\beta t}\,\|\psi^0-\psi_{\mathrm s}\|\qquad\forall t\ge 0.
\]
In particular, the limit map $L:U\to\C^{V}$ given by $L(\psi^0):=\lim_{t\to\infty}\psi(t;\psi^0,w)$ is $C^\infty$ (indeed, $L\equiv \psi_{\mathrm s}$ on $U$).
\end{Theorem}

\begin{proof}
Identical to the nonlinear Lyapunov argument in the proof of Theorem~\ref{th:sm_sch}, now with fixed $w$ (see, e.g., \cite{Khalil2002}).
\end{proof}

\medskip
\noindent
\textbf{Sensitivity with respect to a single edge-weight.}
We next quantify the (local) response of the stationary state to perturbations of one edge. Let $A_e$ denote the elementary Laplacian contribution of an undirected edge $e=(i,j)$:
\[
(A_e)_{kk}=
\begin{cases}
1,&k=i\text{ or }k=j,\\
0,&\text{otherwise,}
\end{cases}
\qquad
(A_e)_{ij}=(A_e)_{ji}=-1,\quad(A_e)_{kl}=0\ \text{otherwise.}
\]

\begin{Lemma}[Edge-weight sensitivity and decay]\label{lem:edge-sensitivity}
Let $\psi_\infty(w)$ be the $C^\infty$ branch of equilibria from Theorem~\ref{th:sm_sch}, with Jacobian $J=D_\psi F(\psi_\infty(w_0),w_0)$ invertible and $\|J^{-1}\|\le \mu^{-1}$. Then, for any $e\in E$,
\[
\partial_{w(e)}\psi_\infty(w)\big|_{w=w_0}=\delta\psi_e
\]
exists and is the unique solution of
\begin{equation}\label{eq:lin-sens}
J\,\delta\psi_e \;=\; -\,\Big(\,-\mathrm{i}\,A_e \;\; -\; \gamma\,P_{\psi_\infty}^\perp A_e\,\Big)\,\psi_\infty.
\end{equation}
Moreover, $\|\delta\psi_e\|\le \mu^{-1}(1+\gamma)\|A_e\|\,\|\psi_\infty\|$. If, in addition, $J$ is Hermitian positive definite with eigenvalues in $[\mu,M]$ and shares the sparsity pattern of $G$, then there exist $C>0$ and $\rho\in(0,1)$ (depending only on $\mu,M$ and the maximal degree) such that
\[
\big|(\delta\psi_e)_u\big| \;\le\; C\,\rho^{\,\mathrm{dist}(u,\{i,j\})}\,\|\psi_\infty\|\qquad \forall u\in V.
\]
\end{Lemma}

\begin{proof}
Differentiate $G(\psi_\infty(w),w)\equiv 0$ with $G\equiv F$; since $G\in C^\infty$ and $J$ is invertible, the implicit function theorem gives differentiability \cite{KrantzParks2002} and \eqref{eq:lin-sens}. The uniform bound follows from $\|J^{-1}\|\le \mu^{-1}$ and $\|P_{\psi_\infty}^\perp\|=1$. For the spatial decay, use the Demko--Moss--Smith off-diagonal decay \cite{Demko1984} for $(J^{-1})_{uv}$ on sparse SPD matrices and the fact that the right-hand side is supported on $\{i,j\}$.
\end{proof}

\medskip
\noindent
\textbf{Passage to the Landau--Lifshitz form via the Bloch map.}
To connect \eqref{eq:sch} with a spin dynamics on $(\mathbb{S}^2)^N$, we use the stereographic (Bloch) map at each vertex $j\in V$:
\begin{equation}\label{eq:Bloch}
\vec{S}_j
=\mathcal{B}(\psi_j)
=\left(
\frac{\psi_j+\overline{\psi}_j}{1+|\psi_j|^2},\;
\frac{\mathrm{i}(\overline{\psi}_j-\psi_j)}{1+|\psi_j|^2},\;
\frac{1-|\psi_j|^2}{1+|\psi_j|^2}
\right)^{\!\top},
\qquad
\psi_j=\mathcal{B}^{-1}(\vec{S}_j)=\frac{S_j^x+\mathrm{i}S_j^y}{1+S_j^z},
\end{equation}
which is smooth and norm-preserving in the sense that $\|\vec{S}_j\|=1$ for all $\psi_j\in\C$, and smooth inverse exists away from the south pole $S_j^z=-1$.

\begin{Lemma}[Smoothness, tangency and conservation under the Bloch map]\label{lem:bloch-smooth}
If $\psi_j(t)$ is $C^1$ then $\vec{S}_j(t)=\mathcal{B}(\psi_j(t))$ is $C^1$, $\frac{d}{dt}\|\vec{S}_j\|^2=0$, and
\[
\frac{d\vec{S}_j}{dt}
=\frac{\partial \vec{S}_j}{\partial \psi_j}\,\dot\psi_j
+\frac{\partial \vec{S}_j}{\partial \overline{\psi}_j}\,\dot{\overline{\psi}}_j
\ \in\ T_{\vec{S}_j}\mathbb{S}^2,
\]
i.e., the induced velocity is tangent to $\mathbb{S}^2$.
\end{Lemma}

\begin{proof}
Differentiate \eqref{eq:Bloch} using the chain rule (Wirtinger calculus) and note that $\|\vec{S}_j\|^2\equiv 1$ algebraically.
\end{proof}

Represent each single-site density as
\[
Q_j=\frac{I+\vec{S}_j\cdot \sigma}{2},
\]
where $\sigma=(\sigma_x,\sigma_y,\sigma_z)$ are the Pauli matrices. The identity (see, e.g., \cite{Sakurai2017})
\begin{equation}\label{eq:comm}
[a\cdot \sigma,\ b\cdot \sigma]=2\mathrm{i}\,(a\times b)\cdot \sigma,\qquad a,b\in\R^3,
\end{equation}
and the relation $\dot Q_j=-\mathrm{i}[H_j,Q_j]$ with Hermitian $H_j=a_j\cdot \sigma$ imply
\begin{equation}\label{eq:precess}
\dot{\vec{S}}_j=\vec{S}_j\times (2a_j).
\end{equation}

\begin{Lemma}[Hamiltonian part $\Rightarrow$ precession]\label{lem:precession}
The Hamiltonian part of \eqref{eq:sch},
\[
\dot\psi=-\mathrm{i}\big(\Delta(w)\psi+\mathrm{diag}(|\psi^0|^2)\psi\big),
\]
induces, under $\mathcal{B}$, the precession
\[
\left.\frac{d\vec{S}_j}{dt}\right|_{\mathrm{Ham}}
=\vec{S}_j \times \Big(-2 \sum_{k} w_{jk}\,\vec{S}_k \;\; +\; 2\,|\psi_j^0|^2\,e_3\Big),
\qquad e_3=(0,0,1)^\top.
\]
\end{Lemma}

\begin{proof}
The linear nearest-neighbor coupling and on-site real potential can be encoded in $H_j=-\sum_k w_{jk}(\vec{S}_k\cdot \sigma)+|\psi_j^0|^2\,\sigma_z\equiv a_j\cdot\sigma$, whence \eqref{eq:precess} yields the claim with $2a_j=-2\sum_k w_{jk}\vec{S}_k+2|\psi_j^0|^2 e_3$.
\end{proof}

\begin{Lemma}[Dissipative projector $\Rightarrow$ Gilbert damping]\label{lem:damping}
The dissipative part in \eqref{eq:sch} contributes, under $\mathcal{B}$, the term
\[
\left.\frac{d\vec{S}_j}{dt}\right|_{\mathrm{diss}}
\;=\; -\gamma\; \vec{S}_j \times \big(\vec{S}_j \times \vec{\mathcal D}_j\big),
\]
where
\[
\vec{\mathcal D}_j
=\; -2\sum_k w_{jk}\,(\vec{S}_k-\vec{S}_j)
\;\; +\; 2\Big(|\psi_j^0|^2-\tfrac{1}{|V|}\sum_{i}|\psi_i^0|^2\Big)e_3.
\]
\end{Lemma}

\begin{proof}
The projector $P_\psi^\perp$ removes the parallel-to-$\psi$ component of the vector $D(\psi,w)=\Delta(w)\psi+\mathrm{diag}(|\psi|^2-|\psi^0|^2)\psi$. On the spin side, orthogonal projection onto $T_{\vec{S}_j}\S^2$ is $u-(u\!\cdot\!\vec{S}_j)\vec{S}_j=\vec{S}_j\times(\vec{S}_j\times u)$. The Laplacian term produces the exchange $-2\sum_k w_{jk}(\vec{S}_k-\vec{S}_j)$; the on-site real term contributes along $e_3$, and subtracting its spatial mean captures the effect of $P_\psi^\perp$ (the mean-parallel piece is annihilated). Hence the stated form (cf.~\cite{Gilbert2004}).
\end{proof}

\begin{Lemma}[Invariance and well-posedness on $(\mathbb{S}^2)^N$]\label{lem:invariance}
If $\psi(t)$ solves \eqref{eq:sch} with $\psi(0)\in \C^{V}\setminus\{0\}$, then the spin trajectory $\{\vec{S}_j(t)\}_{j\in V}$ produced by $\mathcal{B}$ lies in $(\mathbb{S}^2)^N$ and satisfies
\[
\frac{d\vec{S}_j}{dt}
= \vec{S}_j \times \Big(-2 \sum_k w_{jk}\vec{S}_k + 2|\psi_j^0|^2 e_3\Big)
\;\; -\; \gamma\; \vec{S}_j \times \big(\vec{S}_j \times \vec{\mathcal D}_j\big),
\qquad j\in V,
\]
with $\|\vec{S}_j(t)\|\equiv 1$. The right-hand side is locally Lipschitz on the open set $\{(\vec{S}_j)\in(\S^2)^N:\, S_j^z>-1\ \forall j\}$, hence the spin system is locally well-posed there.
\end{Lemma}

\begin{proof}
Combine Lemmas~\ref{lem:bloch-smooth}, \ref{lem:precession}, and \ref{lem:damping}.
\end{proof}

\begin{Theorem}[Legitimate passage to the Landau--Lifshitz--Gilbert form]\label{th:LL}
On the domain where all $\psi_j$ are finite (equivalently, $S_j^z>-1$ under $\mathcal{B}$), the Schr\"odinger-type system \eqref{eq:sch} is smoothly equivalent to the Landau--Lifshitz--Gilbert-type system \cite{Gilbert2004} of Lemma~\ref{lem:invariance}. The transformation \eqref{eq:Bloch} is $C^\infty$, preserves the product-of-spheres phase space, and produces tangent (norm-preserving) dynamics.
\end{Theorem}

\begin{proof}
Immediate from Lemmas~\ref{lem:bloch-smooth}--\ref{lem:invariance}.
\end{proof}

\medskip
\noindent
\textbf{Phase spaces.}
For the Schr\"odinger-type flow with norm preservation one may restrict to the unit sphere
\[
\mathcal{M}_{\mathrm{Sch}}=S^{2N-1}=\{\psi\in\C^{V}:\ \|\psi\|=1\}.
\]
Under the Bloch map (with the harmless gauge fixing $\sum_{j}(1-S_j^z)/(1+S_j^z)=1$), the corresponding spin phase space is the submanifold
\[
\mathcal{M}_{\mathrm{LL}}
=\Big\{(\vec{S}_1,\dots,\vec{S}_N)\in(S^2)^N:\ \sum_{j=1}^N \frac{1-S_j^z}{1+S_j^z}=1,\ S_j^z>-1\Big\},
\]
which is diffeomorphic to $S^{2N-1}$ (the diffeomorphism is induced by \eqref{eq:Bloch}).

\section{Proofs for Section~\ref{sec:stratified-learning}: K\"ahler--Hessian geometry and natural gradient}\label{app:kahler}
\paragraph{Proof traceability.} This appendix proves the fixed-stratum descent statement of Theorem~\ref{thm:main-fixed} together with the geometric face-crossing part of Theorem~\ref{thm:main-crossing}. The threshold-stability conditions used in the terminal-stabilization clause of Theorem~\ref{thm:main-crossing} are supplied in Appendix~\ref{app:recovery}.

\subsection*{Proof of Theorem~\ref{thm:main-fixed}}
The proof uses the metric and regularity lemmas stated below.

\subsection*{Proof of Theorem~\ref{thm:main-crossing}}
The proof combines the finite-face geometry with the support-update lemmas proved in Appendix~\ref{app:recovery}.

\paragraph{Full K\"ahler--Hessian and natural-gradient analysis on the graph moduli.}

\paragraph{K\"ahler--Hessian geometry on the moduli and a natural gradient method.}
The inside-the-stratum inverse-length metric $\sum u_ev_e/w_e^2$ degenerates near faces.  
We construct a {non-degenerate} separable Hessian metric compatible across strata and extendable to a toric K\"ahler structure~\cite{KobayashiNomizu1963}.

\paragraph{Radial Hessian metrics.} Equipped with a K\"ahler--Hessian metric that regularizes face crossings~\cite{KobayashiNomizu1963,Warner1983}.
Fix $\delta\in(0,1]$ and $0<c_0\le c_1$.  
Choose smooth $m_\delta:\R_{>0}\to\R_{>0}$ with
\begin{equation}\label{eq:mdelta}
m_\delta(t)=
\begin{cases}
c_0, & 0<t\le \delta,\\
\text{monotone $C^\infty$ transition}, & \delta<t<2\delta,\\
1/t^2, & t\ge 2\delta,
\end{cases}
\quad\text{and}\quad c_0\le m_\delta(t)\le c_1.
\end{equation}
Let $\psi_\delta$ satisfy $\psi_\delta''=m_\delta$ and set
\begin{equation}\label{eq:phi-metric}
\Phi_\delta(w)=\sum_{e\in E}\psi_\delta(w_e), 
\qquad 
g_\delta(w)=\nabla^2 \Phi_\delta(w)=\mathrm{diag}\big(m_\delta(w_e)\big)_{e\in E}.
\end{equation}
Then $g_\delta$ matches the inverse-tensor metric when $w_e\ge 2\delta$ and is bounded, positive definite up to faces.

\paragraph{Complexification (toric K\"ahler).}
In log-coordinates $s_e=\log w_e$ and angles $\theta_e$, on $(\C^\ast)^{|E|}$ with $z_e=e^{s_e+i\theta_e}$ define
\begin{equation}\label{eq:kahler-potential}
\mathcal{K}_\delta(s)=\sum_{e}\kappa_\delta(s_e),\qquad \kappa_\delta''(s)=e^{2s}\,m_\delta(e^s).
\end{equation}
Then $\omega_\delta=i\partial\bar\partial \mathcal{K}_\delta$ and $g_\delta^\C(\cdot,\cdot)=\omega_\delta(\cdot,J\cdot)$; restricting to $\theta=0$ recovers $g_\delta$.

\begin{Lemma}[Non-degeneracy and asymptotics]\label{lem:nd-asymp}
On $\overline{\mathcal{M}(E)}$, $g_\delta(w)\succeq c_0 I$ and $g_\delta(w)=\mathrm{diag}(1/w_e^2)$ if $w_e\ge 2\delta$ for all $e$.  
In log-coordinates, $\kappa_\delta''(s)\in[c_0 e^{2s},c_1 e^{2s}]$ for $s\le \log\delta$ and $\kappa_\delta''(s)=1$ for $s\ge \log(2\delta)$.
\end{Lemma}

\begin{proof}
Directly from \eqref{eq:mdelta}--\eqref{eq:kahler-potential}.
\end{proof}

\begin{Lemma}[Finite distance to faces]\label{lem:finite-face}
The $g_\delta$-distance to $\{w_e=0\}$ equals $\int_0^{w_e}\sqrt{m_\delta(t)}\,dt\le \sqrt{c_1}\,w_e$; hence face crossing occurs in finite time/steps under $g_\delta$-natural updates.
\end{Lemma}

\begin{proof}
Integrate along the coordinate ray using boundedness of $m_\delta$.
\end{proof}

\begin{Lemma}[Geodesic $L$-regularity]\label{lem:Lreg}
If $f:\overline{\mathcal{M}(E)}\to\R$ is $C^2$ with bounded Euclidean gradient/Hessian, then $f$ is $L$-regular w.r.t.~$g_\delta$, with $L$ determined by bounds on $\nabla f,\nabla^2 f$ and on $m_\delta,m_\delta',m_\delta''$.
\end{Lemma}

\begin{proof}
On a Hessian manifold, bounded third derivatives of $\Phi_\delta$ and bounded $\nabla f,\nabla^2 f$ yield bounded Christoffel symbols and Lipschitz control of $t\mapsto \langle \grad f(\gamma(t)),\dot\gamma(t)\rangle$ along unit-speed geodesics.
\end{proof}

\begin{algorithm}[t]
\caption{Natural Gradient on Stratified Moduli (K\"ahler--Hessian Preconditioning)}\label{alg:natgrad}
\begin{algorithmic}[1]
\REQUIRE $f:\mathcal{M}\to\R$, initial $w^{(0)}\in\mathcal{M}$, steps $\{\alpha_t\}$, smoothing $\delta>0$
\FOR{$t=0,1,2,\dots$}
\STATE Current stratum $E_t=\{e:\ w^{(t)}_e>0\}$
\STATE Compute Euclidean gradient $\nabla f(w^{(t)})$ (subgradient if $w^{(t)}_e=0$)
\STATE $G_\delta^{(t)}=\mathrm{diag}\big(m_\delta(w^{(t)}_e)\big)_{e\in E_t}$
\STATE Natural step and orthant projection:
\[
\tilde w^{(t+1)}\gets w^{(t)}-\alpha_t\ G_\delta^{(t)^{-1}}\ \nabla f(w^{(t)}),\qquad
w^{(t+1)}\gets \Pi_{\R_{\ge 0}^{E_t^\uparrow}}\!\big(\tilde w^{(t+1)}\big),
\]
where $E_t^\uparrow$ augments $E_t$ by coordinates with $\tilde w^{(t+1)}_e>0$.
\ENDFOR
\end{algorithmic}
\end{algorithm}

\begin{Theorem}[Descent on a fixed stratum]\label{thm:natgrad-fixed}
If $f$ is $C^1$ and geodesically $L$-regular on $\overline{\mathcal{M}(E)}$, then for $\alpha_t\in(0,2/L)$ Algorithm~\ref{alg:natgrad} restricted to $E$ satisfies
\[
f(w^{(t+1)})\le f(w^{(t)})-\tfrac{\alpha_t}{2}\ \|\grad_{g_\delta} f(w^{(t)})\|_{g_\delta}^2,
\quad
\sum_t \alpha_t\,\|\grad_{g_\delta} f(w^{(t)})\|_{g_\delta}^2<\infty.
\]
If $f$ is $g_\delta$-PL in a terminal neighborhood, the descent estimate upgrades to linear convergence there; with stochastic gradients of variance $O(B_t^{-1})$ and $\alpha_t=\alpha_0 t^{-1/2}$, $\min_{s<t}\E\|\grad_{g_\delta} f(w^{(s)})\|_{g_\delta}^2=O(t^{-1/2})$.
\end{Theorem}

\begin{proof}
Standard natural-gradient descent on $L$-regular manifolds; PL and stochastic rates follow from classical analyses with preconditioning.
\end{proof}

\begin{Theorem}[Crossing faces and global convergence]\label{thm:crossing}
Assume $f$ is $C^1$ on each stratum, continuous across faces, and satisfies the stratum-wise convexity/regularity conditions of Assumption~\textbf{A2} on the terminal stratum.  
With $\alpha_t\in(0,2/L)$, Algorithm~\ref{alg:natgrad} is well posed across face crossings. If, in addition, the recovery-side threshold-stability conditions of Appendix~\ref{app:recovery} hold under the relevant R-assumptions, then the iterates reach a terminal stratum after finitely many face crossings almost surely, and terminal-stratum convergence reduces to Theorem~\ref{thm:natgrad-fixed}.
\end{Theorem}

\begin{proof}
Non-degeneracy and finite face distances (Lemmas~\ref{lem:nd-asymp}--\ref{lem:finite-face}) yield well posedness across face crossings.  
Finite stabilization of the support process is not a consequence of the metric argument alone: it uses the recovery-side threshold-stability and sign-gap conditions proved in Appendix~\ref{app:recovery} under the relevant R-assumptions, together with concentration. Once the terminal stratum is reached, apply Theorem~\ref{thm:natgrad-fixed}.
\end{proof}

\begin{Theorem}[Selection and geometry under natural gradient]\label{thm:gh-ng}
Under the assumptions of Theorem~\ref{thm:ident}, with stepsizes $\alpha_t=\alpha_0 t^{-1/2}$ and $B_t\asymp t$, Algorithm~\ref{alg:natgrad} adds all and only necessary edges with probability at least $1-\epsilon$ after some finite $T_0$. In the manifold setting of R1, under Assumptions~\textbf{A1}--\textbf{A3}, the learned graphs satisfy
\[
d_{\mathrm{GH}}\!\big((V,d_{G_t}),\mathcal{G}\big)\ \le\ C_1\delta + C_2 t^{-1/2},
\]
and the clique complexes at sub-injectivity thresholds recover $\beta_0,\beta_1$ with high probability for large $t$.

Under the assumptions of Theorem~\ref{thm:cpdag}, the same natural-gradient scheme recovers the CPDAG under intervention coverage.
\end{Theorem}

\begin{proof}
Natural preconditioning scales gradients by $1/m_\delta(w_e)\in[1/c_1,1/c_0]$, preserving signs and margins up to constants; adjust thresholds accordingly.  
Then reuse the proofs of Theorems~\ref{thm:ident}, \ref{thm:cpdag}, and \ref{th:gh}, together with Theorems~\ref{thm:natgrad-fixed}--\ref{thm:crossing}.
\end{proof}

\paragraph{Choice of $m_\delta$.}
A canonical family is $m_\delta(t)=1/(t^2+\delta^2)$, satisfying \eqref{eq:mdelta} with $c_0=c_1=\delta^{-2}$.  
To match $1/t^2$ beyond $2\delta$, splice with a $C^\infty$ partition of unity~\cite{Warner1983} on $[\delta,2\delta]$.  
Then $\kappa_\delta''(s)=\frac{e^{2s}}{e^{2s}+\delta^2}\in(0,1)$, yielding a uniformly elliptic toric K\"ahler metric whose real slice equals $g_\delta$.

\section{Proofs for Section~\ref{sec:stratified-learning}: support recovery, metric recovery, and causality}\label{app:recovery}
\paragraph{Proof traceability.} This appendix proves Theorem~\ref{thm:main-support} together with Corollaries~\ref{cor:main-metric} and \ref{cor:main-homology}. It also records the optional interventional consequence under regime \textbf{G2}.
\subsection*{Proof of Theorem~\ref{thm:main-support}}
The support-identification theorem is obtained from the margin, concentration, and persistence lemmas stated below.
\subsection*{Proof of Corollary~\ref{cor:main-metric}}
The metric corollary follows from the support theorem together with the shortest-path and discretization bounds proved below.
\subsection*{Optional interventional consequence under \textbf{G2}}

\subsection{Latent Graph Learning}
\paragraph{Problem formulation.}
We consider a latent graph model $G=(V,E,w)$, which relates to geometric deep learning and structure/causal graph learning~\cite{bronstein2017geomdl,hamilton2017graphsage,zhou2020gnnreview,zheng2018notears} with a fixed vertex set $V=\{1,\dots,N\}$, weighted edges $E\subseteq V\times V$, and edge weights $w:E\to\R_{>0}$.  
Each graph defines a {hidden-space dynamics} through the stationary solution $\psi_\infty(E,w;\psi^0)$ of the nonlinear Schr\"odinger-type system \eqref{eq:sch}.  
The global learning objective is to optimize $(E,w)$ so as to minimize the expected loss on data samples $(X,y)\sim\mathcal D$:
\begin{equation}\label{eq:Ldef}
    \mathcal L(E,w)
    = \mathbb E_{(X,y)\sim\mathcal D}
        \!\left[
          \mathcal L_{\mathrm{sample}}\!\big(X,y;(E,w)\big)
        \right]
    + \frac{\mu_2}{2}\|w(E)\|_2^2
    + \mu_1\|w(E)\|_1,
\end{equation}
where the sample-level loss is
\[
\mathcal L_{\mathrm{sample}}(X,y;(E,w))
  = \Big(k\big(\psi_\infty(E,w;\psi^0(X))\big) - y\Big)^2,
\]
with $k:\C^{V}\to\R$ a fixed $C^{1,1}$ readout map and $\psi^0(X)$ the encoded input state.  
The regularization parameters $\mu_2>0$ and $\mu_1>0$ enforce, respectively, strong convexity in $w$ on active edges and sparsity of the learned graph.

\paragraph{Moduli space of graphs.}
All graphs on $V$ with positive weights form a stratified smooth space~\cite{Absil2008,Bonnabel2013,Boumal2023}
\[
\mathcal M=\bigsqcup_{E\subseteq V\times V}\mathcal M(E),\qquad
\mathcal M(E)\cong\R^{|E|}_{>0}.
\]
Each stratum $\mathcal M(E)$ corresponds to a fixed edge set and continuous edge weights; transitioning between strata corresponds to adding or removing edges.

\paragraph{Assumptions.}
Throughout the optimization analysis we fix:

\paragraph{Assumptions local to Appendix~\ref{app:recovery}.}
Throughout this appendix we retain the main-text Assumptions~\textbf{A1}--\textbf{A3}.  
For the support, metric, homology, and causal recovery statements proved here, we additionally impose the following appendix-local hypotheses.

\begin{enumerate}[label=\textbf{R\arabic*},leftmargin=1.1cm]
\item \textbf{Geometric observation regime.} The data are sampled from the geometric/manifold regime used in the recovery arguments below, with the corresponding sampling scale and perturbation control.
\item \textbf{Recovery separation regime.} The activation/pruning statistics satisfy the sign-gap and concentration conditions required for support identification under the schedules from Assumption~\textbf{A3}.
\item \textbf{Identifiability regime.} The target support, metric, homology, or causal object is identifiable in the sense required by the corresponding theorem below; any further regime-specific hypotheses are stated locally in the relevant subsection.
\end{enumerate}

Under Assumptions~\textbf{A1}--\textbf{A3} together with the relevant R-assumptions, the recovery statements in this appendix are well posed.

\paragraph{Optimization algorithm on the moduli space.}
We combine continuous SGD steps in $w$ on the current stratum and discrete edge updates, following optimization on stratified/Riemannian manifolds~\cite{Absil2008,Bonnabel2013,Boumal2023}.

\begin{algorithm}[tb]
\caption{Stochastic Gradient Descent on the Moduli Space of Graphs}\label{alg:moduli_sgd}
\begin{algorithmic}[1]
\REQUIRE
iterations $T$; initial edge set $E_0$; initial weights $w_0(e)=1$ for $e\in E_0$; batch-size schedule $B_t$; step-size schedule $\eta_t$; detection threshold $\Theta>0$; activation threshold $\theta>0$; maximum weight $R_w>0$.
\ENSURE $(E_T,w_T)$.
\FOR{$t=0,\dots,T-1$}
    \STATE Sample mini-batch $D_t$, $|D_t|=B_t$.
    \STATE Compute $g_e(E_t,w_t;D_t)=\frac{1}{|D_t|}\sum_{(X,y)\in D_t}\frac{\partial \mathcal L_{\mathrm{sample}}}{\partial w(e)}(X,y,(E_t,w_t))$ for all $e\in E_t$.
    \STATE \textbf{Weight update} for $e\in E_t$:
        \[
        w_{t+1}(e)\leftarrow
        \min\!\Big\{R_w,\;
        \max\!\big\{0,\;w_t(e)-\eta_t\,g_e(E_t,w_t;D_t)\big\}\Big\}.
        \]
    \STATE $E'\gets E_t$.
    \STATE \textbf{Edge activation test} (two-phase detection): for $e\notin E_t$ compute
        \[
        g^{\mathrm{test}}_e=\tfrac{1}{|D_t|}\!\sum_{(X,y)\in D_t}
        \frac{\partial \mathcal L_{\mathrm{sample}}}{\partial w(e)}\!\big(X,y,(E_t\!\cup\!\{e\},w_t+(e,\theta))\big).
        \]
        If $g^{\mathrm{test}}_e<-\Theta$, set $E'\gets E'\cup\{e\}$ and $w_{t+1}(e)=\theta$.
    \STATE \textbf{Edge deactivation} (KKT pruning): if $e\in E_t$ and $|g_e(E_t,w_t;D_t)|\le \mu_1$ and $w_{t+1}(e)\le\theta$, set $E'\gets E'\setminus\{e\}$ and $w_{t+1}(e)=0$.
    \STATE $E_{t+1}\gets E'$.
\ENDFOR
\STATE \textbf{return} $(E_T,w_T)$.
\end{algorithmic}
\end{algorithm}

\paragraph{Parameter schedules.}
We use
\[
\eta_t=\frac{\eta_0}{1+t/t_\eta},\qquad
B_t=B_0(1+t/t_B),
\]
with $\eta_0,t_\eta,B_0,t_B>0$. Then $\sum_t\eta_t=\infty$, $\sum_t\eta_t^2<\infty$ and $\Var[g_e]\le\sigma^2/B_t\to 0$ (Robbins--Monro~\cite{RobbinsMonro1951}; see also \cite{Bottou2018}).  
Thresholds $(\theta,\Theta)$ are chosen so that $\mu_1+\mu_2\theta-\Theta>0$, separating add/remove decisions.

\paragraph{Data model (realizability) and margins.}
Assume there exists $(E_{\mathrm{true}},w^\star)$ with support $E_{\mathrm{true}}$ s.t.
\begin{equation}\label{eq:realizability}
y \;=\; k\!\big(\psi_\infty(E_{\mathrm{true}},w^\star;\psi^0(X))\big) + \xi,
\qquad \mathbb E[\xi\,|\,X]=0,\quad \mathbb E[\xi^2]\le \sigma_y^2,
\end{equation}
and
\begin{equation}\label{eq:true-margin}
w^\star(e)\;\ge\; w_{\min} \;>\; 2\,\theta \qquad \text{for all } e\in E_{\mathrm{true}}.
\end{equation}
Also keep
\begin{equation}\label{eq:threshold-separation}
\mu_1 + \mu_2\theta - \Theta > 0.
\end{equation}

\paragraph{Population test gradient and bounds.}
Let $f_{E,w}(X):=k(\psi_\infty(E,w;\psi^0(X)))$ and $r_{E,w}(X,y):=f_{E,w}(X)-y$.  
For $e=(i,j)$ define the {population test gradient} at $(E^+,w^+)=(E\cup\{e\},\,w+(e,\theta))$ by
\begin{equation}\label{eq:pop-test-grad}
G_e(E,w) \;:=\; \frac{\partial}{\partial w(e)}\,\mathbb E\!\left[r_{E',w'}(X,y)^2\right]
\Big|_{(E',w')=(E^+,\, w^+)}.
\end{equation}

\begin{Lemma}[Population gradient: representation and bounds]\label{lem:pop-grad}
On a fixed stratum $\mathcal M(E)$ and $w\in[\theta,R_w]^{|E|}$,
\begin{equation}\label{eq:pop-grad-repr}
G_e(E,w)
= 2\,\mathbb E\!\left[r_{E^+,w^+}(X,y)\;
\left\langle \nabla k\big(\psi_\infty(E^+,w^+;\psi^0(X))\big),\;
\frac{\partial \psi_\infty}{\partial w(e)}(E^+,w^+;\psi^0(X))\right\rangle\right].
\end{equation}
Moreover, there exist constants $L_k,C_{\psi},C_{\partial}$ (depending only on the model and the stability gap) such that
\[
\|\nabla k(\cdot)\|\le L_k,\quad
\|\psi_\infty(E^+,w^+;\psi^0(X))\|\le C_{\psi},\quad
\left\|\frac{\partial \psi_\infty}{\partial w(e)}(E^+,w^+;\psi^0(X))\right\|\le C_{\partial}.
\]
If in addition the Jacobian $J$ enjoys the sparsity and SPD bounds of Lemma~\ref{lem:edge-sensitivity}\,(ii), the same spatial decay holds for the sensitivity vector.
\end{Lemma}

\begin{proof}
Differentiate under the expectation using the $C^\infty$ map $w\mapsto\psi_\infty$ (Theorems~\ref{th:sm_sch}--\ref{th:sm_in_sch}) and apply the chain rule.  
Uniform bounds follow from local exponential stability and the implicit function theorem; spatial decay from Lemma~\ref{lem:edge-sensitivity}\,(ii).
\end{proof}

\begin{Lemma}[Population gradient separation]\label{lem:sep}
Under Assumptions~\textbf{A1}--\textbf{A3} together with R1--R3 and \eqref{eq:realizability}--\eqref{eq:true-margin} there exist $\gamma_0>0$ and $C_{\mathrm{spur}}>0$ such that for any $E\subseteq V\times V$ and $w\in[\theta,R_w]^{|E|}$:
\begin{enumerate}[label=(\alph*)]
\item If $e\in E_{\mathrm{true}}\setminus E$, then $G_e(E,w)\le -\gamma_0$.
\item If $e\notin E_{\mathrm{true}}$, then $|G_e(E,w)|\le C_{\mathrm{spur}}\,\delta$.
\end{enumerate}
\end{Lemma}

\begin{proof}
Write $r_{E^+,w^+}=f_{E^+,w^+}-f^\star-\xi$ with $f^\star(X)=k(\psi_\infty(E_{\mathrm{true}},w^\star;\psi^0(X)))$ and $\E[\xi|X]=0$.  
(a) Along the ray $w+(e,\tau)$, strong convexity in $w(e)$ and the margin $w^\star(e)\ge w_{\min}>2\theta$ imply a uniform negative directional derivative for $\tau\in[0,\theta]$, hence $G_e(E,w)\le -\gamma_0$.  
(b) For $e\notin E_{\mathrm{true}}$, the effect at test weight $\theta$ is $O(\theta)\,O(e^{-c\rho_0})$ by Lemma~\ref{lem:edge-sensitivity}\,(ii) and geometric separation; the $\delta$-net perturbation makes the correlation $O(\delta)$, giving the bound with $C_{\mathrm{spur}}$.
\end{proof}

\begin{Lemma}[Mini-batch concentration]\label{lem:conc}
Let $g_e^{\mathrm{test}}(E,w;D)$ be the mini-batch test gradient at $(E\cup\{e\},\,w+(e,\theta))$. Under Assumption~\textbf{A3}, for any $\Delta>0$,
\[
\mathbb P\!\left(\,\big|g_e^{\mathrm{test}}(E,w;D) - G_e(E,w)\big| \ge \Delta\,\right)
\;\le\; \frac{\sigma^2}{B_t\,\Delta^2}.
\]
\end{Lemma}

\begin{proof}
Chebyshev with $\Var[g_e^{\mathrm{test}}]\le \sigma^2/B_t$.
\end{proof}

\begin{Lemma}[Necessary edges are added]\label{lem:add}
Assume Assumptions~\textbf{A1}--\textbf{A3} together with R1--R3 and \eqref{eq:realizability}--\eqref{eq:threshold-separation}.  
Choose
\begin{equation}\label{eq:theta-choice}
0<\theta<\frac{\gamma_0-\mu_1}{2\mu_2},\qquad
0<\Theta<\gamma_0-\mu_1-\mu_2\theta.
\end{equation}
For $e\in E_{\mathrm{true}}\setminus E_t$,
\[
\mathbb P\!\left(\,g_{e}^{\mathrm{test}}(E_t,w_t;D_t)\,<\,-\Theta\,\right)
\;\ge\; 1 - \frac{\sigma^2}{B_t\,\Delta_e^2},
\qquad \Delta_e:=\gamma_0-\mu_1-\mu_2\theta-\Theta>0.
\]
\end{Lemma}

\begin{proof}
By Lemma~\ref{lem:sep}(a), $G_e\le -\gamma_0$. The full population gradient at the test point is $G_e+\mu_2\theta+\mu_1\le -\Theta$ by \eqref{eq:theta-choice}. Apply Lemma~\ref{lem:conc}.
\end{proof}

\begin{Lemma}[Spurious edges are not added]\label{lem:noadd}
Assume Assumptions~\textbf{A1}--\textbf{A3} together with R1--R3 and \eqref{eq:threshold-separation}, and choose $\delta>0$ so that
\begin{equation}\label{eq:delta-choice}
C_{\mathrm{spur}}\delta + \mu_2\theta + \mu_1 \;\le\; \Theta.
\end{equation}
Then, for $e\notin E_{\mathrm{true}}$,
\[
\mathbb P\!\left(\,g_{e}^{\mathrm{test}}(E_t,w_t;D_t)\,<\,-\Theta\,\right)
\;\le\; \frac{\sigma^2}{B_t\,(\Theta - \mu_2\theta-\mu_1 - C_{\mathrm{spur}}\delta)^2}.
\]
\end{Lemma}

\begin{proof}
Use Lemma~\ref{lem:sep}(b) and Lemma~\ref{lem:conc} with the gap in \eqref{eq:delta-choice}.
\end{proof}

\begin{Lemma}[True-edge weights stay above the floor]\label{lem:stay}
Under \eqref{eq:theta-choice}, any $e\in E_{\mathrm{true}}$ that is activated at some iteration thereafter satisfies a.s.
\[
\liminf_{t\to\infty} w_t(e) \;\ge\; \frac{\gamma_0-\mu_1}{\mu_2} \;>\; 2\theta,
\]
so the removal rule never triggers for $e$ after a finite time.
\end{Lemma}

\begin{proof}
On a fixed stratum, projected SGD with $\eta_t=\eta_0/(1+t/t_\eta)$ converges a.s.\ to the unique minimizer $w^\star(E)$ by the stratum-wise convexity part of Assumption~\textbf{A2}.  
The KKT condition gives $w^\star(e)\ge (\gamma_0-\mu_1)/\mu_2>2\theta$, hence the claim.
\end{proof}

\begin{Theorem}[High-probability identification in finite time]\label{thm:ident}
Assume Assumptions~\textbf{A1}--\textbf{A3} together with R1--R3, \eqref{eq:realizability}--\eqref{eq:true-margin}, the schedules of Algorithm~\ref{alg:moduli_sgd}, and \eqref{eq:threshold-separation}, \eqref{eq:theta-choice}, \eqref{eq:delta-choice}.  
Then for any $\varepsilon\in(0,1)$ there exist $T_0<\infty$ and $B_0$ such that with $B_t=B_0(1+t/t_B)$,
\[
\mathbb P\!\left(E_t=E_{\mathrm{true}}\ \ \forall t\ge T_0\right) \ \ge\ 1-\varepsilon.
\]
Moreover, on $\mathcal M(E_{\mathrm{true}})$ the projected SGD converges a.s.\ to the unique minimizer $w^\star(E_{\mathrm{true}})$, and $\liminf_{t\to\infty} w_t(e)\ge (\gamma_0-\mu_1)/\mu_2>2\theta$ for each $e\in E_{\mathrm{true}}$.
\end{Theorem}

\begin{proof}
Use union bounds over the ``bad'' events from Lemmas~\ref{lem:add} and \ref{lem:noadd}, choose $B_0$ to make the cumulative probability $\le\varepsilon$, and invoke Lemma~\ref{lem:stay} for persistence. Convergence on the terminal stratum follows from Robbins--Monro with strong convexity.
\end{proof}

\begin{Corollary}[Support identification under margin and strong convexity]\label{cor:pure}
Fix a target support $E^\dagger$ and suppose realizability holds with $w^\star$ supported on $E^\dagger$ with $w^\star(e)\ge w_{\min}>2\theta$.  
Assume Assumptions~\textbf{A1}--\textbf{A3} together with the relevant R-assumptions stated in this appendix and the schedules of Algorithm~\ref{alg:moduli_sgd}.  
If the separation in Lemma~\ref{lem:sep} holds with $E_{\mathrm{true}}$ replaced by $E^\dagger$, then Theorem~\ref{thm:ident} holds verbatim with $E^\dagger$.
\end{Corollary}

\paragraph{Homology and metric consequences.}
Let $\check C_r(V)$ be the \v{C}ech complex at scale $r$ (in $(\mathcal G,d_\mathcal G)$), and $\mathrm{Rips}_r(V)$ the Vietoris--Rips complex at scale $r$.  
For a graph $G$ write $\mathrm{Cl}(G)$ for its clique complex and $\beta_k(\cdot)$ for $k$-th Betti numbers.

\begin{Theorem}[Homology consistency via \v{C}ech/Rips~\cite{EdelsbrunnerHarer2010}]\label{th:homology}
Assume R1 and that $V$ is a $\delta$-net with $\delta<\rho_0/4$. Fix $r\in[\delta,\rho_0/4]$. Then:
\begin{enumerate}[label=(\roman*)]
\item $\check C_r(V)\simeq \mathcal G$, hence $H_k(\check C_r(V))\cong H_k(\mathcal G)$ and $\beta_k(\check C_r(V))=\beta_k(\mathcal G)$ for $k=0,1$.  
Moreover, $\mathrm{Rips}_r(V)\subset \check C_{\sqrt{2}\,r}(V)\subset \mathrm{Rips}_{\sqrt{2}\,r}(V)$, so $H_0,H_1$ stabilize to $H_\bullet(\mathcal G)$ on an interval of scales.
\item On the identification event of Theorem~\ref{thm:ident}, for every $t\ge T_0$,
\[
\mathrm{Cl}(G_t)=\mathrm{Rips}_{\rho_0/2}(V)\ \text{ and }\ \ 
H_k(\mathrm{Cl}(G_t))\cong H_k(\mathcal G),\ \ \beta_k(\mathrm{Cl}(G_t))=\beta_k(\mathcal G)\ \ (k=0,1).
\]
\item For the $1$-skeleton $G_t$, $\beta_0(G_t)=\beta_0(\mathcal G)$, while only $\beta_1(G_t)\ge \beta_1(\mathcal G)$ holds in general; equality for $\beta_1$ is not guaranteed for the graph alone.
\end{enumerate}
\end{Theorem}

\begin{proof}
(i) Good-cover and nerve lemma for $r<\rho/4$ and $\delta\ll r$ give the homotopy equivalence; the Rips--\v{C}ech interleaving yields stabilization of $H_0,H_1$.  
(ii) By construction $E_{\mathrm{true}}=\{(u,v): d_\mathcal G(u,v)<\rho_0\}$, so $\mathrm{Cl}(G_t)$ equals $\mathrm{Rips}_{\rho_0/2}(V)$ and (i) applies.  
(iii) Connectivity is standard for dense neighborhood graphs; extra $1$-cycles disappear when passing to the clique complex, hence only the inequality for $\beta_1(G_t)$.
\end{proof}

For metric control, write edge lengths $\ell(e)=1/w(e)$; let $\ell^\star(e)=d_\mathcal G(u,v)$ for $e=(u,v)\in E_{\mathrm{true}}$.

\begin{Lemma}[Stability of shortest-path metrics]\label{lem:metric-stab}
Let $G=(V,E)$ with edge lengths $\ell,\tilde\ell\in[\ell_{\min},\ell_{\max}]^{E}$, $\ell_{\min}>0$, and induced shortest-path metrics $d_\ell,d_{\tilde\ell}$. Then for all $u,v\in V$,
\[
|d_\ell(u,v)-d_{\tilde\ell}(u,v)| \;\le\; L_{\mathrm{hop}}\ \|\ell-\tilde\ell\|_\infty,
\]
where $L_{\mathrm{hop}}\le \left\lceil \frac{\mathrm{diam}_\mathcal G}{\ell_{\min}}\right\rceil$ bounds the number of edges in any shortest $\tilde\ell$-path.
\end{Lemma}

\begin{proof}
Compare along a $\tilde\ell$-shortest path $p^\star$ with $\le L_{\mathrm{hop}}$ edges:
$d_\ell-d_{\tilde\ell}\le \sum_{e\in p^\star}(\ell(e)-\tilde\ell(e))\le L_{\mathrm{hop}}\|\ell-\tilde\ell\|_\infty$; symmetry gives the reverse bound.
\end{proof}

\begin{Lemma}[Target edge-length approximation]\label{lem:geo-approx}
Let $G^\star=(V,E_{\mathrm{true}},w^\star)$ with $\ell^\star(e)=d_\mathcal G(u,v)$. Then
\[
d_{\mathrm{GH}}\!\big((V,d_{\ell^\star}),\ \mathcal G\big)\ \le\ C_1\,\delta,
\]
for a constant $C_1$ depending only on the geometry of $\mathcal G$ and $\rho_0$.
\end{Lemma}

\begin{proof}
Broken geodesics along $E_{\mathrm{true}}$ approximate manifold geodesics on a $\delta$-net up to $O(\delta)$ distortion; each point of $\mathcal G$ is within $\delta$ of some vertex.
\end{proof}

\begin{Theorem}[Gromov--Hausdorff control~\cite{Gromov1999,Burago2001}]\label{th:gh}
Assume R1 and the identification event of Theorem~\ref{thm:ident}, and let $t\ge T_0$. Let $d_{G_t}$ be the shortest-path metric with $\ell_t(e)=1/w_t(e)$. Then
\[
d_{\mathrm{GH}}\!\big((V,d_{G_t}),\ \mathcal G\big)
\;\le\; C_1\,\delta \;\; +\; C_2\,\|\ell_t-\ell^\star\|_\infty
\;\le\; C_1\,\delta \;\; +\; C_2\,\theta^{-2}\,\|w_t-w^\star\|_\infty,
\]
and, in expectation under projected-SGD on the terminal stratum,
\[
\mathbb E\, d_{\mathrm{GH}}\!\big((V,d_{G_t}),\ \mathcal G\big)\ \le\ C_1\delta + C_2\theta^{-2}\,\mathbb E\|w_t-w^\star\|_2= C_1\delta + O(t^{-1/2}).
\]
\end{Theorem}

\begin{proof}
Triangle inequality with Lemmas~\ref{lem:metric-stab} and \ref{lem:geo-approx}; Lipschitz change of variables $x\mapsto 1/x$ on $[\theta,\infty)$; standard $O(t^{-1/2})$ rate for strongly convex SGD.
\end{proof}

\paragraph{Causal setting: CPDAG recovery from interventional data.}
Let $X=(X_1,\dots,X_d)$ be generated by a causally sufficient, acyclic SCM with DAG $G^\star$, Markov/faithful to the observational distribution with strictly positive noises.  
We observe i.i.d.\ samples from environments $\mathcal{E}=\{e_0,\dots,e_L\}$, where $e_0$ is observational and $e_\ell$ applies perfect interventions on $I_\ell\subseteq[d]$, with coverage $\bigcup_{\ell=1}^L I_\ell=[d]$.  
We optimize the same loss \eqref{eq:Ldef} over $(E,w)$; mini-batches are drawn from the mixture (the environment label is used only for stratified expectations below).  
All schedules, stochastic-gradient conditions, and threshold rules are those from Assumptions~\textbf{A1}--\textbf{A3}; the recovery-specific separation, concentration, and identifiability conditions are supplied by the R-assumptions of this appendix.

\begin{Definition}[Environment-wise gradients and contrasts]\label{def:env-grad}
For $e=(i,j)$ and environment $e_\ell$, set
\[
\Gamma_e^{(\ell)} \;:=\; \E\!\left[\frac{\partial \mathcal{L}_{\text{sample}}}{\partial w(e)}\;\middle|\; e_\ell\right],
\qquad
\Delta^{(k)}_{i\!\to\! j} \;:=\; \Gamma_{(i,j)}^{(\ell)}-\Gamma_{(i,j)}^{(0)}\quad(k\in I_\ell).
\]
\end{Definition}

\begin{Lemma}[Markov-blanket locality and skeleton separation~\cite{Pearl2009}]\label{lem:MB-locality}
There exist constants $C>0$ and $\gamma_{\mathrm{sk}}>0$ such that for any $i\neq j$:
\begin{enumerate}
\item If $j\notin \mathrm{MB}(i)$ in $G^\star$ (Markov blanket), then $|\Gamma_{(i,j)}^{(0)}|\le C\,\delta$.
\item If $j\in \mathrm{MB}(i)$ (equivalently, $i,j$ adjacent in the moralized graph), then $\Gamma_{(i,j)}^{(0)}\le -\gamma_{\mathrm{sk}}$.
\end{enumerate}
Hence, with fixed $\Theta\in(0,\gamma_{\mathrm{sk}})$, observational gradients add exactly the moralized edges and suppress others, up to $O(\delta)$.
\end{Lemma}

\begin{proof}
Express $\partial_{w(e)}\mathcal{L}_{\text{sample}}=2r\cdot\langle \nabla k(\psi_\infty),\partial_{w(e)}\psi_\infty\rangle$ and average.  
By Lemma~\ref{lem:edge-sensitivity}, $\partial_{w(e)}\psi_\infty$ is localized; conditional independences imply $O(\delta)$ effect off the Markov blanket; faithfulness and positivity give a uniform negative drift on the blanket. Compactness and the stability gap yield margins.
\end{proof}

\begin{Lemma}[Orientation by single-node interventions]\label{lem:orient-single}
Fix $k$ and a neighbor $i$ with $(i,k)$ in the moralized skeleton. Under $do(k)$,
\[
\Delta^{(k)}_{i\!\to\! k} \approx 
\begin{cases}
0 & \text{if } i\to k \text{ in } G^\star,\\
-\gamma_{\mathrm{or}} & \text{if } k\to i \text{ in } G^\star,
\end{cases}
\]
for some $\gamma_{\mathrm{or}}>0$, up to $O(\delta)$.
\end{Lemma}

\begin{proof}
Cutting all incoming edges into $k$ cancels dependence on former parents (first case), while leaving outgoing effects intact (second case).  
Stability estimates translate this into a sign gap for expected gradients.
\end{proof}

\begin{Lemma}[V-structures and Meek closure~\cite{Meek1995}]\label{lem:meek}
Suppose $(i,k)$ and $(j,k)$ are in the skeleton and $i\not\sim j$.  
If $\Delta^{(k)}_{i\!\to\! k}\approx 0$ and $\Delta^{(k)}_{j\!\to\! k}\approx 0$, then $i\to k\leftarrow j$ is a compelled collider.  
If $\Delta^{(k)}_{i\!\to\! k}< -\Theta_{\mathrm{or}}$ and $\Delta^{(k)}_{j\!\to\! k}< -\Theta_{\mathrm{or}}$, then $k$ has outgoing orientation to both.  
Closing under Meek rules orients all compelled edges.
\end{Lemma}

\begin{proof}
Zero contrasts certify incoming directions to $k$; faithfulness compels the v-structure.  
Negative contrasts certify outgoing directions.  
Meek closure is standard and correct under acyclicity/faithfulness.
\end{proof}

\begin{Theorem}[Recovery of the CPDAG]\label{thm:cpdag}
Under the SCM assumptions and intervention coverage, with the schedules/thresholds of Algorithm~\ref{alg:moduli_sgd}, there exist $T_0<\infty$ and batch sizes $\{B_t\}$ such that, with probability at least $1-\epsilon$,
\[
\text{(i) the learned skeleton equals that of the CPDAG of } G^\star,\quad
\text{(ii) all compelled edges are oriented correctly.}
\]
\end{Theorem}

\begin{proof}
By Lemma~\ref{lem:MB-locality} and the concentration argument from Theorem~\ref{thm:ident}, observational gradients identify the moral skeleton.  
For each $k$, Lemma~\ref{lem:orient-single} yields a uniform sign margin for interventional contrasts; concentration and a union bound ensure correct empirical signs.  
Apply Lemma~\ref{lem:meek} and Meek closure.
\end{proof}

\begin{Theorem}[Gradient-based CPDAG identification]
\label{th:ours-causal}
Consider a causally sufficient structural causal model (SCM) on variables $X=(X_1,\dots,X_d)$ with a true DAG $G^\star$ (acyclic, Markov and faithful, strictly positive noises). 
Assume we observe i.i.d.\ samples from a mixture of environments $\{e_\ell\}_{\ell=1}^L$, where each environment $e_\ell$ applies a perfect intervention on a subset $I_\ell\subseteq[d]$ and $\bigcup_{\ell=1}^L I_\ell=[d]$ (coverage). 
Let Algorithm~\ref{alg:moduli_sgd} (or its natural-gradient analogue) update edge-weights $w_e$ by projected stochastic gradients of a bounded $L_\ell$--Lipschitz loss with $\ell_1$--regularization and a fixed activation threshold $\theta>0$, and let edges be {activated} when the (signed) gradient statistic exceeds $\theta$ in magnitude. 
Suppose there exist constants $\Delta_{\mathrm{grad}}>0$ and $M<\infty$ such that:

\begin{enumerate}[leftmargin=1.2em]
\item[(G1)] ({Population gradient separation at activation}) For every true skeleton edge $e\in E^\star$ and every iteration $t$ prior to activation, the population score satisfies 
\[
|\mu_e|\ :=\ \big|\ \E[g_e^{(t)}]\ \big|\ \ge \Delta_{\mathrm{grad}},
\]
where $g_e^{(t)}$ is the per-sample gradient contribution (under the current parameters) to the $e$-th weight update; for any spurious edge $e\notin E^\star$, $|\E[g_e^{(t)}]|\le \tfrac12 \Delta_{\mathrm{grad}}$.
\item[(G2)] ({Sub-Gaussian gradients}) For all $e,t$, the centered gradient $g_e^{(t)}-\E[g_e^{(t)}]$ is sub-Gaussian with proxy variance $\sigma^2\le M^2$ (uniformly in $e,t$ and environments).
\item[(G3)] ({Orientation contrast under interventions}) For every true directed edge $u\!\to\! v$ in $G^\star$, there exists an interventional contrast $\kappa_{u\to v}$, computable from the (population) gradients across environments, such that $\kappa_{u\to v}\ge \Delta_{\mathrm{grad}}$ and $\kappa_{v\to u}\le \tfrac12\Delta_{\mathrm{grad}}$. The induced set of compelled orientations is closed under Meek’s rules.
\item[(G4)] ({Optimization control}) The step sizes are chosen so that before activation the parameter drift keeps the population margins in (G1) and (G3) within a fixed fraction of $\Delta_{\mathrm{grad}}$, and projection keeps parameters in compact boxes.
\end{enumerate}

If the mini-batch sizes satisfy, for all $t\ge1$,
\begin{equation}\label{eq:batch-cond}
B_t\ \ge\ C\ \frac{\log\!\big( c\, d^2 t/\epsilon\big)}{\Delta_{\mathrm{grad}}^{\,2}},
\end{equation}
for universal constants $C,c>0$, then there exists a finite (data- and problem-dependent) time $T_0\le C'|E^\star|$ such that, with probability at least $1-\epsilon$ over the draws of mini-batches and environments up to time $T_0$,

\begin{enumerate}[leftmargin=1.2em]
\item[(i)] ({Skeleton recovery}) All and only the true edges are activated by time $T_0$, i.e., the learned skeleton equals that of $G^\star$.
\item[(ii)] ({Orientation}) The directed edges are oriented to the CPDAG of $G^\star$ by the interventional gradient contrasts together with Meek’s closure.
\end{enumerate}

Consequently, by time $T_0$ the algorithm recovers the true CPDAG with probability at least $1-\epsilon$.
\end{Theorem}

\begin{proof}
{Skeleton.}
Fix an iteration $t$ and edge $e$. Let $\widehat\mu_e^{(t)}$ be the mini-batch average of $g_e^{(t)}$ over $B_t$ i.i.d.\ samples (and environment draws). By (G2) and standard sub-Gaussian concentration (Hoeffding/Bernstein), for any $\eta>0$,
\[
\Pr\!\Big(\big|\widehat\mu_e^{(t)}-\E[g_e^{(t)}]\big|>\eta\Big)\ \le\ 2\exp\!\Big(-c_0 B_t \eta^2/M^2\Big).
\]
Choose $\eta=\Delta_{\mathrm{grad}}/4$. Then, using (G1),
\[
\Pr\!\Big(\big|\widehat\mu_e^{(t)}\big|\le \tfrac12\Delta_{\mathrm{grad}}\ \text{for } e\in E^\star\Big)\ \le\ 2\exp(-c_1 B_t \Delta_{\mathrm{grad}}^2),
\]
\[
\Pr\!\Big(\big|\widehat\mu_e^{(t)}\big|> \tfrac12\Delta_{\mathrm{grad}}\ \text{for } e\notin E^\star\Big)\ \le\ 2\exp(-c_1 B_t \Delta_{\mathrm{grad}}^2).
\]
By the activation rule (threshold $\theta$ chosen with $\tfrac12\Delta_{\mathrm{grad}}>\theta>0$), the first event is a {missed activation} for a true edge and the second is a {false activation} for a spurious edge. A union bound over all $e$ ($\le d(d-1)/2$ choices) and times $t\le T$ shows that the probability of any mis-activation up to time $T$ is at most
\[
\le c_2 d^2 T \, \exp(-c_1 B_{\min}\Delta_{\mathrm{grad}}^2),
\]
where $B_{\min}=\min_{t\le T} B_t$. The batch-size condition \eqref{eq:batch-cond} with $B_{\min}$ ensures this failure probability is $\le \epsilon/2$ for $T$ in the next paragraph.

Under (G4), once a true edge is activated its weight is driven away from zero and kept above the threshold by the $\ell_1$--regularizer and projected updates, while spurious edges (if ever activated under noise) are quickly damped below threshold; hence each true edge is activated after some finite number of iterations, and no spurious edge remains active. Each activation increases the number of active true edges by at least one; therefore after at most $|E^\star|$ successful activations the skeleton equals that of $G^\star$. Setting $T_0\le C'|E^\star|$ (to account for occasional non-activating steps due to stochasticity) completes part (i).

{Orientation.}
By (G3), for every true directed edge $u\!\to\! v$, the population interventional contrast satisfies $\kappa_{u\to v}-\kappa_{v\to u}\ge \tfrac12\Delta_{\mathrm{grad}}$. Let $\widehat\kappa$ be the corresponding mini-batch estimator; by sub-Gaussian concentration and the same choice $B_t$ as in \eqref{eq:batch-cond}, the sign of each contrast is correct with probability at least $1-\epsilon/(2 d^2 T_0)$. A union bound over all candidate adjacencies and all orientation steps up to $T_0$ yields total failure probability $\le \epsilon/2$. The compelled orientations are then closed under Meek’s rules, which are deterministic and sound, producing the CPDAG of $G^\star$. This proves (ii).

Combining the two parts and the probability budgets $\epsilon/2+\epsilon/2$ gives CPDAG recovery by time $T_0$ with probability at least $1-\epsilon$.
\end{proof}

\paragraph{Two trained layers and split geometry on a supra-graph.}
Consider two Schr\"odinger-type layers trained jointly with a learned linear map:
\[
\psi^{(1)}_\infty \;=\; L^{(2)}_{E_1,w_1}(\psi^0), \qquad
h \;=\; S\!\big(W\,\psi^{(1)}_\infty\big),\qquad
\psi^{(2)}_\infty \;=\; L^{(2)}_{E_2,w_2}(h),
\]
where $S:\R\to\R$ is $C^1$, bounded, strictly monotone on the range, and $W$ is learned with $\ell_2$-regularization ensuring $\sigma_{\min}(W)\ge \sigma_\bullet>0$, $\|W\|_2\le \Sigma^\bullet$ on terminal strata.

\begin{Definition}[Supra-graph and supra-metric]\label{def:supra}
Let $V^{(1)}=V^{(2)}=V$ be two copies.  
Define
\[
\mathbb{G}_t=\big(V^{(1)}\sqcup V^{(2)},\,E_1(t)\sqcup E_2(t)\sqcup E_{12}(t),\,\omega_t\big),
\]
with inter-layer $E_{12}(t)=V^{(1)}\times V^{(2)}$ and
\[
\omega_t\big((u^{(1)},v^{(2)})\big)=\big\| S'\!\big(W_t\,\psi^{(1)}_\infty(x_u)\big)\,W_t\big\|_{\mathrm{op}},
\]
and symmetric weights for the reverse direction.  
Let $d_{\mathbb{G}_t}$ be the shortest-path metric with edge lengths $1/\omega_t$.
\end{Definition}

\begin{Lemma}[Bi-layer Lipschitz/co-Lipschitz]\label{lem:bi-lip}
On terminal strata there exist $L_1,U_1,L_{12},U_{12},L_2,U_2>0$ such that for any $x,x'\in V$,
\[
\begin{aligned}
&L_1\,d_{\mathcal{G}}(x,x') \le \|\psi^{(1)}_\infty(x)-\psi^{(1)}_\infty(x')\|\le U_1\, d_{\mathcal{G}}(x,x'),\\
&L_{12}\,\|\psi^{(1)}_\infty(x)-\psi^{(1)}_\infty(x')\|\le \|h(x)-h(x')\|\le U_{12}\,\|\psi^{(1)}_\infty(x)-\psi^{(1)}_\infty(x')\|,\\
&L_2\,\|h(x)-h(x')\|\le \|\psi^{(2)}_\infty(x)-\psi^{(2)}_\infty(x')\|\le U_2\,\|h(x)-h(x')\|.
\end{aligned}
\]
Here $L_{12}\ge m_S\,\sigma_\bullet$ and $U_{12}\le M_S\,\Sigma^\bullet$; the other constants follow from stability/smoothness of the Schr\"odinger layer on compact strata.
\end{Lemma}

\begin{proof}
Combine standard bi-Lipschitz bounds for the Schr\"odinger layer with the mean-value bound for $S\circ W$ on a compact image~\cite{Federer1969,EvansGariepy1992}.
\end{proof}

\begin{Lemma}[Supra-graph is a geometric spanner]\label{lem:spanner}
For $x,x'\in V$ corresponding to $u^{(1)},u'^{(1)}\in V^{(1)}$, there exist $C_\downarrow,C_\uparrow>0$ such that
\[
C_\downarrow\, d_{\mathcal{G}}(x,x') \;\le\; d_{\mathbb{G}_t}\!\big(u^{(1)},u'^{(1)}\big) 
\;\le\; C_\uparrow\, d_{\mathcal{G}}(x,x') \;\; +\; O(\delta).
\]
\end{Lemma}

\begin{proof}
Upper bound: traverse $u^{(1)}\to u^{(2)}$ (inter-layer), then within layer~2 along a geodesic-approximating path, and back to $u'^{(1)}$; constants follow from Lemma~\ref{lem:bi-lip}.  
Lower bound: any supra-path composes the three bi-Lipschitz maps, yielding a uniform co-Lipschitz constant.
\end{proof}

\begin{Theorem}[GH convergence of the supra-graph]\label{thm:GH-supra}
Under Assumptions~\textbf{A1}--\textbf{A5}, together with the geometric recovery regime R1--R3 and the schedules of Algorithm~\ref{alg:moduli_sgd}, there exist $C_1,C_2>0$ such that
\[
d_{\mathrm{GH}}\!\Big( (V^{(1)}\!, d_{\mathbb{G}_t}),\,\mathcal{G} \Big) \;\le\; C_1\,\delta \;\; +\; C_2\, t^{-1/2}.
\]
Moreover, the clique complex of the supra-graph at a threshold below the injectivity radius satisfies $\beta_k=\beta_k(\mathcal G)$ for $k=0,1$ with high probability for large $t$.
\end{Theorem}

\begin{proof}
Use Lemma~\ref{lem:spanner} and the same edge-length stability plus SGD error bounds as in Theorem~\ref{th:gh}; the homology claim follows from the nerve argument applied to short-chord subgraphs within the supra-graph.
\end{proof}

\section{Proofs for Sections~\ref{sec:supra} and \ref{sec:equivalence}: global stationary formulations and represented equivalence}\label{app:equiv}
\paragraph{Proof traceability.} This appendix proves Propositions~\ref{prop:main-global-regularity} and \ref{prop:main-global-unique}, Theorems~\ref{thm:main-global-exact}, \ref{thm:main-supra-penalized}, \ref{thm:main-supra-adjoint}, \ref{thm:main-ffgn-sgn}, \ref{thm:main-directed-reduction}, and \ref{thm:main-super}, and Corollaries~\ref{cor:main-depthwidth} and \ref{cor:main-compactness}.

\subsection{Depth-width duality}
This subsection restates the feed-forward computation with stationary Schr\"odinger-type layers, introduces an {injection} formulation (so no Dirichlet boundary data are needed), and develops a precise and fully proved orientability reduction (directed $\Rightarrow$ undirected by diagonal similarity). All results are stated and proved in a form directly usable by later parts of the paper.

\paragraph{Standing context.}
We work with the notation and assumptions fixed earlier: each hidden layer $\ell$ has a learned graph $G_\ell=(V_\ell,E_\ell,w_\ell)$ and a Schr\"odinger-type right-hand side
\begin{equation}
\label{eq:layer-flow}
\dot \psi \;=\; F_\ell(\psi;w_\ell,q_\ell)
\;:=\;
-\mathrm{i}\big(\Delta(w_\ell)+\mathrm{diag}(|\psi^{\,0}_\ell|^2)\big)\psi
\;\; -\;\gamma\,P^\perp_\psi\Big(\Delta(w_\ell)\psi+\mathrm{diag}(|\psi|^2-|\psi^{\,0}_\ell|^2)\psi\Big)
\;\; +\; S_\ell(q_\ell),
\end{equation}
with $\psi\in\C^{V_\ell}\!\setminus\!\{0\}$, $\gamma>0$, and a smooth {injection} $S_\ell:\R^{m_\ell}\to \C^{V_\ell}$ encoding the incoming signal $q_\ell$ from the previous layer. The injection term replaces boundary conditions; equilibria are defined on all of $V_\ell$.

\paragraph{Layerwise feed-forward.}
Given an input $x$, set $q_1=q_1(x)$ and find the unique exponentially stable equilibrium $\psi_{\mathrm s}^{(1)}(w_1,q_1)$ solving $F_1(\psi;w_1,q_1)=0$. Let $y_1=A_1\psi_{\mathrm s}^{(1)}+b_1$ (linear readout), then $q_2=q_2(y_1)$, solve for $\psi_{\mathrm s}^{(2)}(w_2,q_2)$, and so on up to layer $L$. The network output is $f(x)=k(\psi_{\mathrm s}^{(L)}(w_L,q_L))$.

\paragraph{Well-posedness with injections.}
The following is a direct Corollary of the smoothness and stability results already established.

\begin{Lemma}[Smooth well-posedness under injections]\label{lem:inj-wellposed}
Fix a layer $\ell$ and parameter boxes $w_\ell\in W_\ell$ and $q_\ell\in Q_\ell$ (compact). Suppose for every $(w_\ell,q_\ell)\in W_\ell\times Q_\ell$ the stationary equation $F_\ell(\psi;w_\ell,q_\ell)=0$ admits an {isolated exponentially stable} equilibrium $\psi_{\mathrm s}^{(\ell)}(w_\ell,q_\ell)\neq 0$. Then the map $(w_\ell,q_\ell)\mapsto \psi_{\mathrm s}^{(\ell)}(w_\ell,q_\ell)$ is $C^\infty$ on $W_\ell\times Q_\ell$, and the Jacobian $D_\psi F_\ell(\psi_{\mathrm s}^{(\ell)};w_\ell,q_\ell)$ is uniformly Hurwitz there.
\end{Lemma}

\begin{proof}
Fix $(w_\ell,q_\ell)$. The map $(\psi,w_\ell,q_\ell)\mapsto F_\ell(\psi;w_\ell,q_\ell)$ is $C^\infty$ on $(\C^{V_\ell}\!\setminus\!\{0\})\times W_\ell\times Q_\ell$ because $\Delta(w_\ell)$ is linear in $w_\ell$, $P^\perp_\psi$ is analytic in $\psi\neq 0$, and $|\psi|^2\psi$ is polynomial; $S_\ell$ is smooth by assumption. Exponential stability of the equilibrium $\psi_{\mathrm s}^{(\ell)}$ implies invertibility of $D_\psi F_\ell(\psi_{\mathrm s}^{(\ell)};w_\ell,q_\ell)$ (Hurwitz). The implicit function Theorem yields a $C^\infty$ equilibrium branch locally; compactness of $W_\ell\times Q_\ell$ and uniqueness allow gluing to a global $C^\infty$ branch. Uniform Hurwitzness follows by continuity of the spectrum of $D_\psi F_\ell$ and compactness.
\end{proof}

\paragraph{Directed vs.~undirected intra-layer operators.}
Within a layer $\ell$, one may wish to encode directional preferences by a {directed} Laplacian $L_\ell^{\rightarrow}$ in place of the symmetric $\Delta(w_\ell)$. We now give a precise orientability criterion and a full reduction proof.

\begin{Definition}[Orientability: Doob (diagonal) transform]\label{def:orientable}
A linear operator $L^{\rightarrow}$ on $\R^{V}$ (or $\C^{V}$) is {orientable} if there exist a strictly positive diagonal $D_h=\mathrm{diag}(h)$ with $h:V\to\R_{>0}$ and a symmetric Laplacian $L=L^\top$ such that
\[
L^{\rightarrow}\;=\;D_h^{-1}\,L\,D_h.
\]
 (Doob's $h$-transform; cf.~\cite{Doob1959}).
 for weighted digraphs, we use the standard cycle condition.
\end{Definition}

\begin{Lemma}[Cycle-curl criterion]\label{lem:cycle-curl}
Let $w^{\rightarrow}_{uv}>0$ denote arc weights for $u\to v$; define $L^{\rightarrow}$ by
\[
(L^{\rightarrow}\psi)(u)\;:=\;\sum_{u\to v}w^{\rightarrow}_{uv}\big(\psi(u)-\psi(v)\big).
\]
Then $L^{\rightarrow}$ is orientable iff for every directed cycle $C=(v_0\to v_1\to\cdots\to v_{k-1}\to v_0)$,
\[
\prod_{i=0}^{k-1}\frac{w^{\rightarrow}_{v_i v_{i+1}}}{w^{\rightarrow}_{v_{i+1} v_i}}\;=\;1
\quad\text{(equivalently } \sum_i \log w^{\rightarrow}_{v_i v_{i+1}}-\log w^{\rightarrow}_{v_{i+1} v_i}=0\text{)}.
\]
\end{Lemma}
\noindent{Comment.} This is the Kolmogorov cycle condition for reversibility in Markov chains; see \cite{Kelly1979,Norris1997}. A cycle-basis formulation makes the criterion explicit \cite{Biggs1993,GodsilRoyle2001}.

\begin{proof}
($\Rightarrow$) If $L^{\rightarrow}=D_h^{-1}LD_h$ with $L$ symmetric and $L_{uv}=-a_{uv}$ where $a_{uv}=a_{vu}\ge 0$, then for $u\neq v$
\(
w^{\rightarrow}_{uv}=\!-L^{\rightarrow}_{uv}=\!-h(u)^{-1}L_{uv}h(v)=h(u)^{-1}a_{uv}h(v).
\)
Hence
\(
\frac{w^{\rightarrow}_{uv}}{w^{\rightarrow}_{vu}}=\frac{h(v)^2}{h(u)^2}.
\)
Multiplying along a directed cycle telescopes to $1$.

($\Leftarrow$) Suppose every directed cycle satisfies the ratio-$1$ condition. Fix a spanning tree $T$ of the underlying undirected graph. Choose $h$ inductively on $T$: set $h(v_0)=1$ at a root $v_0$, and for an edge $\{u,v\}\in T$ define $h(v)$ by
\[
\frac{h(v)}{h(u)}\;=\;\sqrt{\frac{w^{\rightarrow}_{uv}}{w^{\rightarrow}_{vu}}}
\quad\text{if the arc }u\to v\text{ exists (if not, swap roles)}.
\]
This is well-defined along $T$. For any non-tree edge $\{u,v\}$, the ratio $\frac{h(v)}{h(u)}$ computed along the unique cycle equals the product of square-roots of arc ratios on that cycle; by the hypothesis each full cycle product equals $1$, hence consistency holds. Define $a_{uv}:=\sqrt{w^{\rightarrow}_{uv}w^{\rightarrow}_{vu}}=a_{vu}\ge 0$ and $L$ by $L_{uv}=-a_{uv}$ for $u\neq v$ and $L_{uu}=\sum_{u\sim v} a_{uv}$. Then $L$ is symmetric Laplacian and $L^{\rightarrow}=D_h^{-1}LD_h$ by construction.
\end{proof}

\begin{Theorem}[Diagonal reduction of a directed stationary layer]
\label{th:diag-reduction}
Let $L^{\rightarrow}$ be orientable: $L^{\rightarrow}=D_h^{-1} L D_h$ with $L=L^\top$ a symmetric Laplacian and $h>0$. Consider the stationary equation on $V$
\begin{equation}
\label{eq:dir-stationary}
L^{\rightarrow}\psi\; +\;\mathcal N(\psi)\; +\;b\;=\;0,
\qquad \mathcal N(\psi):=\mathrm{diag}(|\psi|^2-|\psi^0|^2)\psi,
\quad b\in \C^{V}.
\end{equation}
Define the change of variables $\phi:=D_h\psi$ and the transformed nonlinearity and source
\[
\widetilde{\mathcal N}(\phi):=D_h\,\mathcal N(D_h^{-1}\phi),
\qquad
\widetilde b:=D_h b.
\]
Then $\psi$ solves \eqref{eq:dir-stationary} iff $\phi$ solves the undirected stationary equation
\begin{equation}
\label{eq:undir-stationary}
L\phi\; +\;\widetilde{\mathcal N}(\phi)\; +\;\widetilde b\;=\;0.
\end{equation}
Moreover, if the dissipative ODE $\dot\psi= -\mathrm{i}H\psi-\gamma P^\perp_\psi(\cdots)$ with $H$ Hermitian is used to reach equilibria in the directed coordinates, then the conjugated ODE in $\phi$-coordinates preserves norms and has Jacobian similar to the original one at equilibria; thus uniqueness and exponential stability of equilibria are equivalent between \eqref{eq:dir-stationary} and \eqref{eq:undir-stationary}.
\end{Theorem}
\noindent{Remark.} For intuition via electrical networks and random walks, see \cite{DoyleSnell1984}.

\begin{proof}
Substitute $\psi=D_h^{-1}\phi$ in \eqref{eq:dir-stationary}:
\[
D_h^{-1}L\phi\; +\;\mathcal N(D_h^{-1}\phi)\; +\;b\;=\;0.
\]
Multiplying by $D_h$ gives \eqref{eq:undir-stationary}. Conversely, dividing \eqref{eq:undir-stationary} by $D_h$ recovers \eqref{eq:dir-stationary}. For the dynamical part, the Hamiltonian term $-\mathrm{i}H\psi$ with $H$ Hermitian conjugates to $-\mathrm{i}\,\widetilde H \phi$ with $\widetilde H:=D_h H D_h^{-1}$, which remains similar to a Hermitian operator (hence has purely imaginary spectrum); the dissipative projector $P^\perp_\psi=I-\frac{\psi\psi^\dagger}{\|\psi\|^2}$ transforms to $I-\frac{\phi\phi^\dagger D_h^{-\dagger}D_h^{-1}}{\|\psi\|^2}$, which still annihilates the component along the state (positive diagonal similarity preserves the nullspace direction). At an equilibrium the Jacobians are related by similarity via $D_h$, so spectral abscissae coincide; thus uniqueness and exponential stability carry over.
\end{proof}

\paragraph{Parametric sufficient conditions for orientability.}
For general independent arc weights, the cycle conditions of Lemma~\ref{lem:cycle-curl} are non-generic (they define an algebraic subvariety). In our learning pipeline we use parametrizations that {guarantee} orientability. Two equivalent sufficient constructions are recorded for later use.

\begin{Proposition}[Exponentiated potential parametrization]
\label{prop:param-orient}
Fix symmetric nonnegative base weights $a_{uv}=a_{vu}\ge 0$ and a vertex potential $\varphi:V\to\R$. Define directed weights
\[
w^{\rightarrow}_{uv}\;:=\;a_{uv}\,\exp(\varphi(v)-\varphi(u)).
\]
Then the corresponding $L^{\rightarrow}$ is orientable with $h=e^{\varphi}$ and $L$ given by the symmetric Laplacian with off-diagonals $-a_{uv}$.
\end{Proposition}

\begin{proof}
Immediate: $D_h^{-1} L D_h$ has off-diagonals $-h(u)^{-1}a_{uv}h(v)=-a_{uv}e^{\varphi(v)-\varphi(u)}=-w^{\rightarrow}_{uv}$; diagonals match by row-sum identities.
\end{proof}

\begin{Proposition}[Cycle-curl penalization enforces orientability at stationary points]
\label{prop:penalty-orient}
Let $\mathcal J(\theta)$ be a differentiable training objective over parameters $\theta$ inducing arc weights $w^{\rightarrow}_{uv}(\theta)$. Suppose the augmented loss
\[
\mathcal J_\lambda(\theta)\;:=\;\mathcal J(\theta)\; +\;\frac{\lambda}{2}\sum_{C\in\mathcal C}\Big(\sum_{(u\to v)\in C}\log \tfrac{w^{\rightarrow}_{uv}(\theta)}{w^{\rightarrow}_{vu}(\theta)}\Big)^{\!2}
\]
uses a cycle basis $\mathcal C$ of the underlying graph and a fixed $\lambda>0$. If $\hat\theta$ is a (local) minimizer of $\mathcal J_\lambda$ with $w^{\rightarrow}_{uv}(\hat\theta)>0$ on all arcs, then all cycle-curls vanish at $\hat\theta$, hence $L^{\rightarrow}(\hat\theta)$ is orientable.
\end{Proposition}

\begin{proof}
The penalty is a sum of squares of smooth functions of $\theta$. At a local minimizer $\hat\theta$ with strictly positive arc weights, the gradient of $\mathcal J_\lambda$ vanishes. The only way the penalty gradient can vanish for all cycle directions (a full-rank linear map in the logs of weights along a cycle basis) is that each squared term is $0$. Hence the cycle-curl of every basis cycle is zero, and therefore of every cycle. Lemma~\ref{lem:cycle-curl} applies.
\end{proof}

\paragraph{Consequences for our architecture.}
\begin{itemize}
\item {Within a layer}, if directed effects are desired, either use the parametrization of Proposition~\ref{prop:param-orient} (ensuring orientability by construction) or add the penalty of Proposition~\ref{prop:penalty-orient} (ensuring orientability at learned stationary points).
\item {Across layers}, we will couple equilibria via symmetric quadratic constraints in the global stationary formulation; no inter-layer orientation is needed at the operator level.
\item {No Dirichlet boundaries} are required anywhere: all exogenous information enters through $S_\ell(q_\ell)$ and symmetric couplings, consistent with the norm-preserving dissipative dynamics used to reach equilibria.
\end{itemize}

What follows develops a global stationary formulation for the layered architecture with Schr\"odinger-type intra-layer blocks and linear inter-layer couplings, and establishes the equivalence between: (i) the usual layerwise feed-forward sequence of stationary solves, (ii) a global stationary problem on the supra-graph, and (iii) reverse-mode backpropagation and the adjoint of the global stationary system. Throughout we adopt the orientability reduction: any directed intra-layer operator is replaced by its undirected Doob-conjugate; inter-layer couplings are modeled by symmetric linear constraints. Hence all operators appearing below are real-Hermitian (symmetric) unless stated otherwise.

\paragraph{Notation and standing hypotheses.}
Fix an integer $L\ge 1$. For each layer $\ell\in\{1,\dots,L\}$:
\begin{itemize}
\item $G_\ell=(V_\ell,E_\ell,w_\ell)$ is a learned (undirected) weighted graph with Laplacian $L_\ell:=\Delta(w_\ell)\in\R^{n_\ell\times n_\ell}$, $n_\ell:=|V_\ell|$.
\item The layer state is $\psi^\ell\in\C^{n_\ell}\setminus\{0\}$.
\item The injection map is $S_\ell:\R^{m_\ell}\to\C^{n_\ell}$, smooth.
\item The stationary equation is
\begin{equation}
\label{eq:single-layer-stationary}
F_\ell(\psi^\ell;w_\ell,q_\ell)\;:=\;L_\ell\psi^\ell\; +\;\mathcal N_\ell(\psi^\ell)\; +\;b_\ell(q_\ell)\;=\;0,
\end{equation}
where $\mathcal N_\ell(\psi):=\mathrm{diag}(|\psi|^2-|\psi^{\,0}_\ell|^2)\psi$ and $b_\ell(q_\ell)$ abbreviates the (undirected) form of $S_\ell(q_\ell)$ plus the on-site linear piece.
\item Inter-layer coupling is linear and {directed at the level of signals}, not as a diffusion operator: a readout $y_\ell=A_\ell\psi^\ell+c_\ell$ is mapped to the next input by $q_{\ell+1}=B_{\ell+1}y_\ell+d_{\ell+1}$, with fixed matrices $A_\ell\in\C^{p_\ell\times n_\ell}$, $B_{\ell+1}\in\C^{m_{\ell+1}\times p_\ell}$ and vectors $c_\ell\in\C^{p_\ell}$, $d_{\ell+1}\in\C^{m_{\ell+1}}$ lying in compact parameter boxes.
\end{itemize}
We assume the following:
\begin{description}
\item[(H1) (Layer stability).] For each $\ell$, there is an open set $U_\ell$ of parameters $(w_\ell,q_\ell)$ such that \eqref{eq:single-layer-stationary} admits a unique isolated exponentially stable equilibrium $\psi_{\mathrm s}^{(\ell)}(w_\ell,q_\ell)\neq 0$, and the Jacobian $J_\ell:=D_{\psi}F_\ell(\psi_{\mathrm s}^{(\ell)};w_\ell,q_\ell)$ is {Hurwitz} uniformly on compact subsets of $U_\ell$. Moreover $(w_\ell,q_\ell)\mapsto \psi_{\mathrm s}^{(\ell)}$ is $C^\infty$ on $U_\ell$.
\item[(H2) (Acyclic inter-layer signal flow).] The directed acyclic graph on $\{1,\dots,L\}$ is the chain $1\to 2\to\cdots\to L$, possibly with skip connections forward in index but no backward edges. Thus $q_{\ell+1}$ depends only on $(\psi^1,\dots,\psi^\ell)$ via linear maps $A_k,B_k$ and constants.
\item[(H3) (Compact parameter box).] All parameters lie in fixed compact boxes on which (H1) holds and the inter-layer maps are bounded.
\end{description}

\paragraph{Global supra-graph and block variables.}
Define the disjoint union $V:=\bigsqcup_{\ell=1}^L V_\ell$ and the block variable $\Psi:=\big(\psi^1,\dots,\psi^L\big)\in\C^{n}$ with $n:=\sum_\ell n_\ell$. Let $L_{\mathrm{blk}}:=\mathrm{diag}(L_1,\dots,L_L)\in\R^{n\times n}$. To encode inter-layer linear couplings, define the affine constraints
\begin{equation}
\label{eq:interlayer-constraint}
q_{\ell+1}\;=\;B_{\ell+1}\big(A_\ell \psi^\ell+c_\ell\big)+d_{\ell+1},\qquad \ell=1,\dots,L-1,
\end{equation}
and write $b_\ell(q_\ell)$ in \eqref{eq:single-layer-stationary} with $q_1=q_1(x)$ (external input).

\paragraph{Global stationary system (exact constraints).}
Introduce Lagrange multipliers $\Lambda:=(\lambda^1,\dots,\lambda^L)$ with $\lambda^\ell\in\C^{n_\ell}$. Consider the block system
\begin{equation}
\label{eq:KKT-stationary}
\begin{cases}
F_\ell(\psi^\ell;w_\ell,q_\ell)\;=\;0,& \ell=1,\dots,L,\\[2pt]
q_{\ell+1}-B_{\ell+1}(A_\ell\psi^\ell+c_\ell)-d_{\ell+1}\;=\;0,& \ell=1,\dots,L-1.
\end{cases}
\end{equation}
We call \eqref{eq:KKT-stationary} the {global stationary system with exact couplings}. It can also be obtained as the KKT system for the constrained optimization problem below.

\begin{Definition}[Global constrained energy]
\label{def:global-energy}
Define the real-valued functional $\mathcal E:\C^n\to \R$ by
\[
\mathcal E(\Psi)
\;:=\;
\sum_{\ell=1}^L\Big(\tfrac12 \langle \psi^\ell,L_\ell \psi^\ell\rangle \; +\; \Phi_\ell(\psi^\ell)\; +\;\Re\langle b_\ell(q_\ell),\psi^\ell\rangle\Big),
\]
where $\Phi_\ell(\psi):=\sum_{j=1}^{n_\ell}\phi_\ell(|\psi_j|^2)$ with $\phi_\ell'(r)=\tfrac12(r-|\psi^{\,0}_{\ell,j}|^2)$ (so that $\nabla_{\psi}\Phi_\ell=\mathcal N_\ell(\psi)$). Consider:
\begin{equation}
\label{eq:constrained-min}
\min_{\Psi\in\C^n}\;\mathcal E(\Psi)
\quad\text{s.t.}\quad
\eqref{eq:interlayer-constraint}.
\end{equation}
\end{Definition}

\begin{Lemma}[KKT vs.\ stationary equations]
\label{lem:KKT-stationary}
Suppose each $L_\ell$ is symmetric positive semidefinite and $\phi_\ell$ is $C^2$ strictly convex on compact sublevel sets covering the admissible region. Then any KKT point $(\Psi^\star, \Xi^\star)$ of \eqref{eq:constrained-min} (with Lagrange multipliers $\Xi^\star$ for \eqref{eq:interlayer-constraint}) satisfies the first-order stationarity conditions
\[
\nabla_{\psi^\ell}\mathcal E(\Psi^\star) \; +\; (A_\ell^\ast B_{\ell+1}^\ast)\,\xi^{\ell+1\,\star}\; -\;\xi^{\ell\,\star}_{\mathrm{in}}\;=\;0\quad(\ell=1,\dots,L),
\]
with suitable partition of $\Xi^\star=(\xi^{2},\dots,\xi^{L})$ across constraints ($\xi^1_{\mathrm{in}}:=0$), and the constraints \eqref{eq:interlayer-constraint}. If additionally $b_\ell$ depends affinely on $q_\ell$ and the auxiliary multipliers are eliminated, then the primal stationarity reduces to $F_\ell(\psi^{\ell\,\star};w_\ell,q_\ell^\star)=0$ for all $\ell$, hence \eqref{eq:KKT-stationary}.
\end{Lemma}

\begin{proof}
The Lagrangian is
\[
\mathcal L(\Psi,\Xi)
=\mathcal E(\Psi)
+\sum_{\ell=1}^{L-1} \Re\left\langle \xi^{\ell+1},\,q_{\ell+1}-B_{\ell+1}(A_\ell\psi^\ell+c_\ell)-d_{\ell+1}\right\rangle.
\]
Stationarity in $\psi^\ell$ yields
\[
\nabla_{\psi^\ell}\mathcal E(\Psi) \; -\; A_\ell^\ast B_{\ell+1}^\ast \xi^{\ell+1}\; +\;\underbrace{\partial_{\psi^\ell}q_\ell^\ast}_{\text{only for }\ell\ge 2}\,\xi^{\ell}\;=\;0.
\]
Because $q_1$ is external, we set $\xi^1_{\mathrm{in}}:=0$. If $b_\ell$ is affine in $q_\ell$, the terms carrying $\partial_{\psi^\ell}q_\ell$ and those carrying $b_\ell'(q_\ell)$ cancel (by the chain rule and the linearity of $q_\ell$ in upstream variables), leaving $\nabla_{\psi^\ell}\mathcal E(\Psi)=0$ except for the forward coupling $A_\ell^\ast B_{\ell+1}^\ast \xi^{\ell+1}$. Eliminating multipliers by the constraints recovers $F_\ell(\psi^\ell;w_\ell,q_\ell)=0$ as the primal stationarity. The remaining KKT conditions are the constraints themselves, which are exactly \eqref{eq:interlayer-constraint}.
\end{proof}

\begin{Theorem}[Equivalence: layerwise feed-forward $\Longleftrightarrow$ global exact stationarity]
\label{th:equiv-global-exact}
Under \textup{(H1)}--\textup{(H3)} and the convexity regularity of Lemma~\ref{lem:KKT-stationary}, the following are equivalent for a fixed external input $q_1$:
\begin{enumerate}[leftmargin=1.4em,label=(E\arabic*)]
\item The layerwise feed-forward procedure finds the unique equilibria $\psi_{\mathrm s}^{(\ell)}(w_\ell,q_\ell)$ sequentially with $q_{\ell+1}=B_{\ell+1}(A_\ell\psi_{\mathrm s}^{(\ell)}+c_\ell)+d_{\ell+1}$.
\item The global constrained program \eqref{eq:constrained-min} has a unique KKT point $(\Psi^\star,\Xi^\star)$, and its primal component $\Psi^\star$ equals $\big(\psi_{\mathrm s}^{(1)},\dots,\psi_{\mathrm s}^{(L)}\big)$.
\item The block system \eqref{eq:KKT-stationary} has a unique solution, which equals the tuple of layerwise equilibria.
\end{enumerate}
\end{Theorem}

\begin{proof}
(E1)$\Rightarrow$(E3): By construction, the tuple $\Psi_{\mathrm s}:=\big(\psi_{\mathrm s}^{(1)},\dots,\psi_{\mathrm s}^{(L)}\big)$ satisfies $F_\ell(\psi_{\mathrm s}^{(\ell)};w_\ell,q_\ell)=0$ and the constraints \eqref{eq:interlayer-constraint}. Uniqueness follows because each $F_\ell(\cdot;w_\ell,q_\ell)$ has a unique isolated exponentially stable equilibrium and \eqref{eq:interlayer-constraint} is deterministic.

(E3)$\Rightarrow$(E2): Any solution of \eqref{eq:KKT-stationary} obeys primal stationarity $F_\ell=0$ and the coupling constraints. By Lemma~\ref{lem:KKT-stationary}, this corresponds to a KKT point for \eqref{eq:constrained-min}. Uniqueness of the primal component follows from (H1).

(E2)$\Rightarrow$(E1): At a KKT point, primal stationarity reduces to $F_\ell(\psi^\ell;w_\ell,q_\ell)=0$ for all $\ell$, hence each $\psi^\ell$ must be the unique equilibrium $\psi_{\mathrm s}^{(\ell)}(w_\ell,q_\ell)$ by (H1), and the constraints ensure the correct inter-layer propagation of $q_\ell$.
\end{proof}

\paragraph{Global stationary system (penalized couplings).}
Instead of enforcing the inter-layer couplings exactly, we may relax them by a symmetric quadratic penalty
\begin{equation}
\label{eq:penalty-energy}
\mathcal E_\tau(\Psi)
:=\mathcal E(\Psi)
+\frac{\tau}{2}\sum_{\ell=1}^{L-1}\Big\|q_{\ell+1}-B_{\ell+1}(A_\ell\psi^\ell+c_\ell)-d_{\ell+1}\Big\|_2^2,
\qquad \tau>0.
\end{equation}
This is a standard quadratic-penalty relaxation of the exact constrained problem: for finite $\tau$ it need not satisfy the coupling constraints exactly, but its stationary points converge to the exact constrained stationary point as $\tau\to\infty$.

\begin{Theorem}[Quadratic-penalty consistency]
\label{th:penalty-exactness}
Suppose \textup{(H1)}--\textup{(H3)} hold, $b_\ell$ is affine in $q_\ell$, and the exact constrained problem \eqref{eq:constrained-min} has a unique KKT point with primal solution $\Psi_{\mathrm s}$. For every $\tau>0$, let $\Psi_\tau$ denote the unique minimizer of the penalized energy \eqref{eq:penalty-energy}. Then
\[
\|r(\Psi_\tau)\|\to 0
\qquad\text{and}\qquad
\Psi_\tau\to \Psi_{\mathrm s}
\quad\text{as }\tau\to\infty,
\]
where $r(\Psi)$ stacks the coupling residuals in \eqref{eq:interlayer-constraint}. If, in addition, the reduced exact Hessian at $\Psi_{\mathrm s}$ is invertible, then locally
\[
\|\Psi_\tau-\Psi_{\mathrm s}\|+\|r(\Psi_\tau)\|=O(\tau^{-1}).
\]
\end{Theorem}

\begin{proof}
Write
\[
\mathcal E_\tau(\Psi)=\mathcal E(\Psi)+\frac{\tau}{2}\|r(\Psi)\|_2^2.
\]
Because the admissible parameter region is compact and the exact constrained problem has the feasible point $\Psi_{\mathrm s}$, the penalized minimizers remain in a compact set. Let $\tau_k\to\infty$ and extract a convergent subsequence $\Psi_{\tau_k}\to \bar\Psi$. Since $\Psi_\tau$ minimizes $\mathcal E_\tau$,
\[
\mathcal E_{\tau_k}(\Psi_{\tau_k})\le \mathcal E_{\tau_k}(\Psi_{\mathrm s})=\mathcal E(\Psi_{\mathrm s}),
\]
and therefore the nonnegative penalty term $\frac{\tau_k}{2}\|r(\Psi_{\tau_k})\|^2$ must remain bounded. This forces $\|r(\Psi_{\tau_k})\|\to 0$, hence $r(\bar\Psi)=0$. Passing to the limit in the first-order stationarity conditions for $\mathcal E_{\tau_k}$ yields the KKT system of the exact constrained problem at $\bar\Psi$. By uniqueness of the exact KKT point, $\bar\Psi=\Psi_{\mathrm s}$. Thus every convergent subsequence has the same limit, so the whole family satisfies $\Psi_\tau\to \Psi_{\mathrm s}$ and $\|r(\Psi_\tau)\|\to 0$. The local $O(\tau^{-1})$ estimate is the standard first-order penalty expansion obtained from the implicit function theorem around the exact KKT point.
\end{proof}

\paragraph{Factorization through global diffusion.}
Theorems~\ref{th:equiv-global-exact}--\ref{th:penalty-exactness} imply that feed-forward (chain of stationary solves) equals solving {one} stationary system \eqref{eq:KKT-stationary} (or minimizing \eqref{eq:penalty-energy} with large $\tau$). We now show that, after orientability reduction, the global system can be seen as a {single} diffusion on the supra-graph plus a linear post-processing.

\begin{Definition}[Supra-graph Laplacian and coupling lift]
\label{def:supra-lift}
Let $L_{\mathrm{blk}}=\mathrm{diag}(L_1,\dots,L_L)$ and define the linear coupling operator $C:\C^n\to \C^m$ that stacks constraints $q_{\ell+1}-B_{\ell+1}(A_\ell\psi^\ell+c_\ell)-d_{\ell+1}$, $m:=\sum_{\ell=1}^{L-1} m_{\ell+1}$. Define the symmetric positive semidefinite operator
\[
\mathcal L_\tau\;:=\;L_{\mathrm{blk}}+ C^\ast (\tau I) C,
\]
and the nonlinear block map $\mathcal N(\Psi):=(\mathcal N_1(\psi^1),\dots,\mathcal N_L(\psi^L))$, together with the block source $b:=(b_1(q_1),0,\dots,0)$ (the only external input is in layer 1).
\end{Definition}

\begin{Proposition}[Global diffusion with penalty]
\label{prop:global-diff}
For every $\tau>0$, the global penalized stationary equation
\begin{equation}
\label{eq:global-diff}
\mathcal L_\tau \Psi \; +\; \mathcal N(\Psi)\; +\;b\;=\;0
\end{equation}
is the Euler equation of the penalized energy \eqref{eq:penalty-energy}. Its unique solution $\Psi_\tau$ coincides with the unique minimizer of $\mathcal E_\tau$. Moreover,
\[
\Psi_\tau \to \Psi_{\mathrm s}
\qquad\text{as }\tau\to\infty,
\]
where $\Psi_{\mathrm s}$ is the exact feed-forward equilibrium tuple. Thus \eqref{eq:global-diff} provides a global quadratic-penalty relaxation of the exact coupled stationary problem.
\end{Proposition}

\begin{proof}
Equation \eqref{eq:global-diff} is precisely the first-order stationarity condition $\nabla \mathcal E_\tau(\Psi)=0$. Under the strong monotonicity assumptions already imposed in this appendix, $\mathcal E_\tau$ has a unique minimizer, hence a unique stationary point, denoted $\Psi_\tau$. The convergence $\Psi_\tau\to\Psi_{\mathrm s}$ follows directly from Theorem~\ref{th:penalty-exactness}.
\end{proof}

\paragraph{Reverse-mode (backprop) as global adjoint.}
This is the standard adjoint-state viewpoint \cite{GilesPierce2000}, which coincides with classical backpropagation \cite{RumelhartHintonWilliams1986}; see also algorithmic differentiation for reverse-mode calculus \cite{GriewankWalther2008}.

Let the scalar loss be $\mathcal J:=\ell(k(\psi^L),y)$ with $\ell:[-1,1]\times[-1,1]\to[0,1]$ $L_\ell$--Lipschitz in the first argument and $k$ $C^{1,1}$. We compute gradients w.r.t.\ any parameter $\theta$ (e.g. an edge weight, or an inter-layer matrix entry) in two ways: (i) {layerwise backprop} through the chain of implicit maps $(w_\ell,q_\ell)\mapsto \psi_{\mathrm s}^{(\ell)}$, and (ii) {global adjoint} for \eqref{eq:global-diff}. We show they coincide.

\begin{Lemma}[Layerwise implicit differentiation]
\label{lem:layer-implicit}
Under \textup{(H1)}--\textup{(H3)}, the differential of $\psi_{\mathrm s}^{(\ell)}$ w.r.t.\ a perturbation $\delta\theta$ in any parameter satisfies
\[
J_\ell\,\delta \psi^\ell \; +\; \partial_{\theta} F_\ell\,\delta\theta \; +\; \partial_{q_\ell}F_\ell\,\delta q_\ell \; =\; 0,
\qquad
\delta q_{\ell+1}\;=\;B_{\ell+1}\big(A_\ell\,\delta\psi^\ell+\delta A_\ell\,\psi^\ell+\delta c_\ell\big)+\delta d_{\ell+1}.
\]
Hence $\delta\psi^\ell$ can be computed by backward substitution starting from $\ell=L$ with the terminal sensitivity $\nabla_{\psi^L}\mathcal J$.
 The backward recursion uses the chain rule as formalized in algorithmic differentiation \cite{GriewankWalther2008}.\end{Lemma}

\begin{proof}
Differentiate $F_\ell(\psi_{\mathrm s}^{(\ell)};w_\ell,q_\ell)=0$; invertibility of $J_\ell$ (Hurwitz) gives the first relation. The second is the linearization of \eqref{eq:interlayer-constraint}. The chain rule for $\delta \mathcal J=\langle \nabla_{\psi^L}\mathcal J,\delta\psi^L\rangle + \text{parametric terms}$ implies a backward (reverse-mode) recursion when solving for $\delta\psi^\ell$ in terms of $\delta\psi^{\ell+1}$ through $\delta q_{\ell+1}$.
\end{proof}

\begin{Theorem}[Global adjoint equals backprop]
\label{th:adjoint-backprop}
Let $\mathcal A_{\mathrm{ex}}$ denote the linearization of the exact global KKT system \eqref{eq:KKT-stationary} at the exact stationary tuple $\Psi_{\mathrm s}$, and let the exact adjoint co-state $\Lambda_{\mathrm{ex}}$ solve
\begin{equation}
\label{eq:global-adjoint}
\mathcal A_{\mathrm{ex}}^\ast \Lambda_{\mathrm{ex}} \;=\; \nabla_{\Psi}\mathcal J(\Psi_{\mathrm s}),
\end{equation}
where $\nabla_\Psi \mathcal J$ is nonzero only in the terminal block. Then for any parameter $\theta$,
\[
\frac{d\mathcal J}{d\theta}
\;=\; -\,\Re\langle \Lambda_{\mathrm{ex}},\;\partial_\theta F_{\mathrm{ex}}\rangle,
\]
and this value equals the gradient produced by the layerwise backpropagation of Lemma~\ref{lem:layer-implicit}. If $\Lambda_\tau$ denotes the adjoint associated with the penalized problem at $\Psi_\tau$, then
\[
\Lambda_\tau \to \Lambda_{\mathrm{ex}}
\qquad\text{as }\tau\to\infty.
\]
\end{Theorem}

\begin{proof}
Differentiate the exact KKT system \eqref{eq:KKT-stationary} with respect to $\theta$:
\[
\mathcal A_{\mathrm{ex}}\,\delta\Psi + \partial_\theta F_{\mathrm{ex}}\,\delta\theta = 0.
\]
Pair this equation with the exact adjoint $\Lambda_{\mathrm{ex}}$ solving \eqref{eq:global-adjoint} and take real parts:
\[
\Re\langle \nabla_\Psi \mathcal J,\delta\Psi\rangle
=
\Re\langle \mathcal A_{\mathrm{ex}}^\ast\Lambda_{\mathrm{ex}},\delta\Psi\rangle
=
\Re\langle \Lambda_{\mathrm{ex}},\mathcal A_{\mathrm{ex}}\delta\Psi\rangle
=
-\,\Re\langle \Lambda_{\mathrm{ex}},\partial_\theta F_{\mathrm{ex}}\rangle\,\delta\theta.
\]
Hence
\[
\frac{d\mathcal J}{d\theta}
=
-\,\Re\langle \Lambda_{\mathrm{ex}},\partial_\theta F_{\mathrm{ex}}\rangle.
\]
Writing $\mathcal A_{\mathrm{ex}}$ in block form and solving the adjoint equation by backward substitution reproduces exactly the reverse-mode recursion of Lemma~\ref{lem:layer-implicit}. The convergence $\Lambda_\tau\to\Lambda_{\mathrm{ex}}$ follows from Theorem~\ref{th:penalty-exactness} together with continuity of the linearized adjoint solve under invertible perturbations.
\end{proof}

\paragraph{Directed/undirected factorization and computable post-processing.}
Reintroduce (optional) directed intra-layer operators $L_\ell^{\rightarrow}$ that are {orientable} in the sense of Definition~\ref{def:orientable}. Let $L_\ell^{\rightarrow}=D_{h_\ell}^{-1}L_\ell D_{h_\ell}$ and define the block positive diagonal $D_h:=\mathrm{diag}(D_{h_1},\dots,D_{h_L})$.

\begin{Theorem}[Directed feed-forward $\equiv$ undirected global diffusion $+$ diagonal post-processing]
\label{th:factorization}
Assume \textup{(H1)}--\textup{(H3)} and orientability for each layer: $L_\ell^{\rightarrow}=D_{h_\ell}^{-1}L_\ell D_{h_\ell}$. Consider the layerwise directed stationary chain (with injections already pulled back to the directed coordinates). Let $\Psi_{\mathrm s}^{\rightarrow}$ be its unique feed-forward equilibrium tuple. Define $\Phi_{\mathrm s}:=D_h\,\Psi_{\mathrm s}^{\rightarrow}$. Then $\Phi_{\mathrm s}$ is the unique solution of the undirected global diffusion \eqref{eq:global-diff} (with the appropriately transformed nonlinearities and sources as in Theorem~\ref{th:diag-reduction}), and
\[
\Psi_{\mathrm s}^{\rightarrow}\;=\;D_h^{-1}\,\Phi_{\mathrm s}.
\]
Moreover, the gradients of any scalar loss agree under the identification: $d\mathcal J/d\theta$ computed in directed feed-forward equals the global-adjoint value for the undirected problem with the diagonal pullback/pushforward of variations.
\end{Theorem}

\begin{proof}
Apply the diagonal change of variables layerwise (Theorem~\ref{th:diag-reduction}) to convert each directed stationary equation into an undirected one with transformed nonlinearity and source. Stack the layers and insert symmetric quadratic couplings as in Proposition~\ref{prop:global-diff}; by Theorem~\ref{th:penalty-exactness}, the undirected global solution coincides with the stacked undirected equilibria. Undoing the diagonal map yields the directed feed-forward equilibrium. For gradients, variations transform by $\delta\Phi=D_h\,\delta\Psi^{\rightarrow}$; the adjoint obeys the conjugated equation with $\mathcal A$ similar to the directed Jacobian, hence the inner products $-\Re\langle \Lambda,\partial_\theta F\rangle$ agree.
\end{proof}

\paragraph{Consequences and computational corollaries.}
\begin{itemize}
\item Feed-forward by sequential stationary solves may be replaced by a {single} solve of \eqref{eq:global-diff} with large penalty $\tau$, using any monotone-splitting or Newton--Krylov method; the resulting stationary state approximates the exact coupled solution and converges to it as $\tau\to\infty$ (Theorems~\ref{th:penalty-exactness}--\ref{prop:global-diff}).
\item Backpropagation equals solving the {global adjoint} \eqref{eq:global-adjoint}; block backward substitution reproduces the standard layerwise reverse-mode (Theorem~\ref{th:adjoint-backprop}).
\item If directed intra-layer effects are used, orientability allows a diagonal factorization into the undirected global diffusion plus a computable pointwise post-/pre-processing (Theorem~\ref{th:factorization}).
\end{itemize}

\paragraph{Well-posedness and uniqueness for the global problem.}
We close with sufficient conditions ensuring uniqueness of the global stationary solution (hence robustness of the equivalences).

\begin{Theorem}[Strong monotonicity $\Rightarrow$ unique global solution]
\label{th:strong-mono}
Assume each $L_\ell\succeq 0$ and there exists $\mu>0$ such that for all $\ell$ and all $\psi,\varphi\in\C^{n_\ell}$,
\[
\Re\langle \mathcal N_\ell(\psi)-\mathcal N_\ell(\varphi),\ \psi-\varphi\rangle \ \ge\ \mu\,\|\psi-\varphi\|_2^2.
\]
Let $\tau\ge 0$, and define $\mathcal L_\tau$ as in Definition~\ref{def:supra-lift}. Then the operator
\[
\mathcal F(\Psi)\;:=\;\mathcal L_\tau \Psi\; +\;\mathcal N(\Psi)\; +\;b
\]
is {strongly monotone} on $\C^n$ with constant $\mu$, hence the equation $\mathcal F(\Psi)=0$ admits a unique solution, which coincides with the feed-forward equilibrium tuple $\Psi_{\mathrm s}$ for $\tau\ge \tau_0$ (Theorem~\ref{th:penalty-exactness}).
\end{Theorem}

\begin{proof}
For any $\Psi,\Phi\in\C^n$,
\[
\Re\langle \mathcal F(\Psi)-\mathcal F(\Phi),\ \Psi-\Phi\rangle
= \Re\langle \mathcal L_\tau(\Psi-\Phi),\ \Psi-\Phi\rangle
  + \sum_{\ell=1}^L \Re\langle \mathcal N_\ell(\psi^\ell)-\mathcal N_\ell(\varphi^\ell),\ \psi^\ell-\varphi^\ell\rangle.
\]
The first term is $\ge 0$ because $\mathcal L_\tau\succeq 0$; the second is $\ge \mu\sum_\ell \|\psi^\ell-\varphi^\ell\|_2^2=\mu\|\Psi-\Phi\|_2^2$ by hypothesis. Thus $\mathcal F$ is $\mu$--strongly monotone, so $\mathcal F(\Psi)=0$ has a unique solution by Minty--Browder. For the exact constrained problem, the corresponding KKT solution is unique. For the penalized problem, the unique penalized stationary state converges to $\Psi_{\mathrm s}$ as $\tau\to\infty$ by Theorem~\ref{th:penalty-exactness}.
\end{proof}

Further, we develop the formal equivalence among three model classes:
\begin{enumerate}[leftmargin=1.4em,label=(\roman*)]
\item classical feed-forward neural networks (FFNN) with a broad class of (possibly implicit) activations;
\item layered {feed-forward graph networks} (FFGN) whose intra-layer mappings are defined by (unique) stationary solutions of Schr\"odinger-type blocks introduced earlier;
\item a {single} global stationary system on the supra-graph (SGN).
\end{enumerate}
We give explicit constructions in both directions and show that, on the admissible strongly monotone classes considered below, the corresponding represented hypothesis classes coincide. We also quantify parameterization compactness of the graph-based representations.

\paragraph{Standing notation and operator-theoretic background.}
For a (possibly set-valued) operator $M:\C^n\rightrightarrows \C^n$, the {resolvent} is $J_M:=(I+M)^{-1}$ whenever single-valued; $M$ is {(maximal) monotone} if $\Re\langle u-v,x-y\rangle\ge 0$ for all $u\in Mx$, $v\in My$ (and maximal w.r.t.\ graph inclusion).\footnote{All proofs below work over $\R^n$; we keep $\C^n$ to match the Schr\"odinger notation.} If $M=\partial \Phi$ is the subdifferential of a proper, closed, convex function $\Phi$, then $J_{\partial\Phi}=\mathrm{prox}_\Phi$ is the proximal map. If $M$ is $\mu$--strongly monotone, $J_M$ is single-valued and everywhere defined (Minty--Browder). We denote by $\mathbb{L}(m,n)$ the space of $m\times n$ complex matrices.

 Standard references on monotone operators and resolvents include \cite{BauschkeCombettes2011} and the classical proximal point theory of Moreau \cite{Moreau1965}.\paragraph{Model classes.}
Fix a depth $L\ge 1$.
\begin{Definition}[Classical FFNN with resolvent activations]
\label{def:ffnn-class}
An FFNN is a composition $f:\C^{d_0}\to\C^{d_L}$,
\[
z^0=x,\qquad u^\ell=W_\ell z^{\ell-1}+b_\ell,\qquad z^\ell=\sigma_\ell(u^\ell),
\qquad f(x)=C\,z^L+c,
\]
with $W_\ell\in\mathbb{L}(d_\ell,d_{\ell-1})$, $b_\ell\in\C^{d_\ell}$, $C\in\mathbb{L}(d_{\mathrm{out}},d_L)$, $c\in\C^{d_{\mathrm{out}}}$. We assume each activation $\sigma_\ell$ is the resolvent of a (maximal) monotone operator $M_\ell$:
\[
\sigma_\ell=J_{M_\ell}=(I+M_\ell)^{-1}.
\]
We write $\mathsf{FFNN}_{\mathrm{res}}$ for this hypothesis class. If additionally $M_\ell=\partial\Phi_\ell$ for separable convex $\Phi_\ell$ (coordinatewise sum), we write $\mathsf{FFNN}_{\mathrm{prox}}$.
\end{Definition}

\begin{Remark}
The class $\mathsf{FFNN}_{\mathrm{prox}}$ contains many popular activations: projection/ReLU ($\mathrm{prox}_{\iota_{\R_{\ge 0}}}$), leaky-ReLU and ELU (proximals of convex, piecewise-quadratic/exponential penalties), soft-threshold ($\mathrm{prox}_{\lambda\|\cdot\|_1}$), hardtanh (projection onto an interval), etc.\ The larger class $\mathsf{FFNN}_{\mathrm{res}}$ includes implicit/DEQ-style activations modeled as resolvents of strongly monotone operators.
\end{Remark}

\begin{Definition}[Layered feed-forward graph network (FFGN)]
\label{def:ffgn-class}
For each layer $\ell$ let $G_\ell=(V_\ell,E_\ell,w_\ell)$ with Laplacian $L_\ell\in\R^{n_\ell\times n_\ell}$, a $C^2$ convex potential $\Phi_\ell:\C^{n_\ell}\to\R\cup\{+\infty\}$ with $\mu_\ell$--strongly monotone subdifferential $\partial\Phi_\ell$, and an affine source $b_\ell(q_\ell)=B_\ell q_\ell+d_\ell$ where $q_\ell\in\C^{m_\ell}$ is the input to layer $\ell$. The {intra-layer mapping} is defined as the unique stationary solution
\begin{equation}
\label{eq:ffgn-layer}
\psi^\ell(x)\;=\;\arg\min_{\psi\in\C^{n_\ell}}\ \frac12\langle \psi,L_\ell \psi\rangle+\Phi_\ell(\psi)-\Re\langle B_\ell q_\ell(x)+d_\ell,\psi\rangle,
\end{equation}
and the inter-layer linear map is $q_{\ell+1}=A_{\ell}\psi^{\ell}+c_\ell$ with $A_\ell\in\mathbb{L}(m_{\ell+1},n_\ell)$, $c_\ell\in\C^{m_{\ell+1}}$. The overall predictor is $f(x)=C\,\psi^{L}(x)+c$ with $C\in\mathbb{L}(d_{\mathrm{out}},n_L)$, $c\in\C^{d_{\mathrm{out}}}$. We write $\mathsf{FFGN}$ for this class.
\end{Definition}

\begin{Definition}[Single global supra-graph (SGN)]
\label{def:sgn-class}
Stack variables $\Psi=(\psi^1,\dots,\psi^L)\in\C^{n}$, $n=\sum_\ell n_\ell$, and define the block energy
\[
\mathcal E(\Psi)
:=\sum_{\ell=1}^L \Big(\tfrac12\langle \psi^\ell,L_\ell\psi^\ell\rangle+\Phi_\ell(\psi^\ell)-\Re\langle B_\ell q_\ell + d_\ell,\psi^\ell\rangle\Big),
\]
subject to exact {linear} inter-layer constraints $q_{\ell+1}=A_\ell\psi^\ell+c_\ell$ (with $q_1$ given). The SGN output is $C\,\psi^L+c$. We write $\mathsf{SGN}$ for the set of maps $x\mapsto C\,\psi^L(x)+c$ where $\Psi(x)$ is the unique KKT solution of the constrained convex program $\min \mathcal E(\Psi)$ (existence and uniqueness hold by strong monotonicity as in Theorem~\ref{th:strong-mono}).
\end{Definition}

\subsubsection{Exact encoding of FFNN-prox into FFGN}

\begin{Theorem}[FFNN with proximal activations is an FFGN layer]
\label{th:prox-to-ffgn}
Fix $\ell$ and let $\sigma_\ell=\mathrm{prox}_{\Phi_\ell}$ for a proper, closed, convex $\Phi_\ell:\C^{d_\ell}\to\R\cup\{+\infty\}$. Define an FFGN layer by choosing
\[
n_\ell=d_\ell,\qquad 
L_\ell=I_{d_\ell},\qquad 
B_\ell=I_{d_\ell},\qquad
d_\ell=0,\qquad
q_\ell := u^\ell=W_\ell z^{\ell-1}+b_\ell,\qquad
A_\ell:=I,\ c_\ell:=0.
\]
Then the unique minimizer of \eqref{eq:ffgn-layer} satisfies
\(
\psi^\ell = \mathrm{prox}_{\Phi_\ell}(u^\ell)=\sigma_\ell(u^\ell).
\)
Consequently, any $f\in\mathsf{FFNN}_{\mathrm{prox}}$ is exactly representable by some $\tilde f\in\mathsf{FFGN}$ with the same depth and output: $\tilde f\equiv f$.
\end{Theorem}

\begin{proof}
With the stated choices, the layer energy is
\(
E(\psi;u^\ell)=\tfrac12\|\psi\|_2^2+\Phi_\ell(\psi)-\Re\langle u^\ell,\psi\rangle.
\)
First-order optimality is $0\in \psi-u^\ell+\partial\Phi_\ell(\psi)$, i.e., $\psi=J_{\partial\Phi_\ell}(u^\ell)=\mathrm{prox}_{\Phi_\ell}(u^\ell)=\sigma_\ell(u^\ell)$. Cascading these layers alongside the affine $u^\ell=W_\ell z^{\ell-1}+b_\ell$ reproduces the FFNN computation exactly.
\end{proof}

\begin{Corollary}[Parameter identity]
\label{cor:param-identity}
The construction in Theorem~\ref{th:prox-to-ffgn} preserves all affine parameters $(W_\ell,b_\ell)$ and introduces no additional trainable parameters besides those of $\Phi_\ell$ (which already underlie $\sigma_\ell$). Hence $\#\mathrm{params}(\tilde f)=\#\mathrm{params}(f)$.
\end{Corollary}

\subsubsection{Exact encoding of FFGN into FFNN-res}

The next result shows that any graph-stationary layer equals the resolvent of a {maximal monotone} operator followed by an affine map; hence it is an admissible activation in $\mathsf{FFNN}_{\mathrm{res}}$.

\begin{Lemma}[Graph-stationary layer is a resolvent]
\label{lem:layer-resolvent}
Let $\Phi_\ell$ be proper, closed, convex with $\partial\Phi_\ell$ $\mu_\ell$--strongly monotone, and $L_\ell\succeq 0$. Define the maximal monotone operator
\(
M_\ell:=L_\ell+\partial\Phi_\ell.
\)
Then $M_\ell$ is $\mu_\ell$--strongly monotone and $J_{M_\ell}$ is single-valued. If $u^\ell:=B_\ell q_\ell+d_\ell$, the unique minimizer of \eqref{eq:ffgn-layer} satisfies
\(
\psi^\ell = J_{M_\ell}(u^\ell).
\)
\end{Lemma}

\begin{proof}
Monotonicity: for any $(\psi,g)\in\partial\Phi_\ell$, $(\varphi,h)\in\partial\Phi_\ell$,
\[
\Re\langle (L_\ell\psi+g)-(L_\ell\varphi+h),\psi-\varphi\rangle
=\Re\langle L_\ell(\psi-\varphi),\psi-\varphi\rangle + \Re\langle g-h,\psi-\varphi\rangle
\ge 0+\mu_\ell\|\psi-\varphi\|^2,
\]
so $M_\ell$ is $\mu_\ell$--strongly monotone and maximal (sum of a bounded linear monotone operator and a maximal monotone subdifferential). The KKT condition for \eqref{eq:ffgn-layer} is $0\in L_\ell\psi + \partial\Phi_\ell(\psi) - u^\ell$, i.e., $u^\ell\in (I+M_\ell)(\psi)$, which is equivalent to $\psi=J_{M_\ell}(u^\ell)$.
\end{proof}

\begin{Theorem}[FFGN layer-by-layer encoding into $\mathsf{FFNN}_{\mathrm{res}}$]
\label{th:ffgn-to-ffnn}
Let a layer of an FFGN be given by \eqref{eq:ffgn-layer} with affine $u^\ell=B_\ell q_\ell+d_\ell$ and readout $q_{\ell+1}=A_\ell\psi^\ell+c_\ell$. Define the FFNN-res layer by
\[
u^\ell:=B_\ell q_\ell+d_\ell,\qquad z^\ell:=\sigma_\ell(u^\ell),\qquad \sigma_\ell:=J_{M_\ell},\quad M_\ell=L_\ell+\partial\Phi_\ell,\qquad q_{\ell+1}:=A_\ell z^\ell+c_\ell.
\]
Then $z^\ell\equiv \psi^\ell$ for all $\ell$. Consequently, any $f\in\mathsf{FFGN}$ is exactly representable by some $\hat f\in\mathsf{FFNN}_{\mathrm{res}}$ with $\hat f\equiv f$.
\end{Theorem}

\begin{proof}
By Lemma~\ref{lem:layer-resolvent}, $\psi^\ell=J_{M_\ell}(u^\ell)$, and the inter-layer affine maps coincide. Induction on $\ell$ yields equality for all layers.
\end{proof}

\begin{Remark}[Separable case and $\mathsf{FFNN}_{\mathrm{prox}}$]
If $\Phi_\ell$ is {separable} in the canonical basis (coordinatewise sum) and $L_\ell=\lambda_\ell I$ ($\lambda_\ell\ge 0$), then $M_\ell=\partial(\Phi_\ell+\tfrac{\lambda_\ell}{2}\|\cdot\|^2)$ and $\sigma_\ell=J_{M_\ell}=\mathrm{prox}_{\Phi_\ell+\tfrac{\lambda_\ell}{2}\|\cdot\|^2}$ is a classical proximal activation. Hence, in this important sub-class, the encoding lands in $\mathsf{FFNN}_{\mathrm{prox}}$.
\end{Remark}

We restate it as a representational correspondence between function classes.

\begin{Theorem}[Representational correspondence between $\mathsf{FFGN}$ and $\mathsf{SGN}$]
\label{th:ffgn-sgn-bijection}
Under the strong monotonicity and convexity assumptions of Definitions~\ref{def:ffgn-class}--\ref{def:sgn-class}, the classes $\mathsf{FFGN}$ and $\mathsf{SGN}$ have the same represented hypothesis class. Equivalently, every realized input-output map $x\mapsto C\,\psi^L(x)+c$ obtained from one class is realized by the other.
\end{Theorem}

\begin{proof}
\noindent{Convex-analytic background.} Equivalence of KKT conditions and stationary points for smooth convex problems with linear constraints is classical \cite{BoydVandenberghe2004,RockafellarWets2009}.

({Injectivity}) Given an FFGN, stack the layer energies to form $\mathcal E(\Psi)$ and impose exact linear constraints. By Theorem~\ref{th:equiv-global-exact}, the KKT solution equals the layered equilibria and the outputs coincide.

({Surjectivity}) Conversely, given an SGN instance, define the per-layer problems by freezing the constraints as Definitions of $q_{\ell+1}$; the KKT equations decompose into the per-layer stationarity and linear inter-layer updates. Uniqueness of solutions implies that the SGN map equals that of the constructed FFGN.

Therefore, the two constructions are inverses at the functional level.
\end{proof}

\subsubsection{Compact parameterization and sparsity advantages}

We compare intrinsic parameter counts needed to represent the same family of maps.

\begin{Definition}[Family dimensions]
For a fixed graph support $E_\ell$ on $n_\ell$ vertices, the cone of Laplacians $\{L_\ell=\Delta(w_\ell): w_\ell\in\R_{>0}^{E_\ell}\}$ is an $|E_\ell|$-dimensional linear manifold in the space of symmetric matrices with row-sum zero. We denote by
\[
\dim \mathcal{L}(E_\ell)=|E_\ell|,\qquad
\dim \mathcal{A}_\ell = \dim\{A_\ell,B_\ell\} = m_{\ell+1}n_\ell + n_\ell m_\ell
\]
the counts of free scalar parameters for Laplacian weights and inter-layer affine maps.
\end{Definition}

\begin{Lemma}[Minimality of edge-parameterization]
\label{lem:minimality}
Let $\mathcal{F}$ be the family of intra-layer linear quadratic forms $\psi\mapsto \tfrac12\langle \psi,L_\ell\psi\rangle$ with $L_\ell\in\mathcal{L}(E_\ell)$. Any linear parameterization $\theta\mapsto \tilde L(\theta)$ that surjects onto $\mathcal{L}(E_\ell)$ must have $\dim\theta\ge |E_\ell|$. The standard edge-weight parameterization achieves this lower bound with equality.
\end{Lemma}

\begin{proof}
$\mathcal{L}(E_\ell)$ is an $|E_\ell|$-dimensional linear subspace of the vector space of symmetric matrices satisfying the Laplacian constraints (by the linear independence of edge incidence rank-1 contributions in the Laplacian basis). Any linear surjection from a parameter space $\R^p$ onto a linear subspace of dimension $|E_\ell|$ must have $p\ge |E_\ell|$. The map $w_\ell\mapsto \Delta(w_\ell)$ is linear and injective, hence minimal with $p=|E_\ell|$.
\end{proof}

\begin{Theorem}[Parameter compactness of $\mathsf{FFGN}$ vs.\ dense FFNN]
\label{th:compactness}
Fix layer sizes $(n_0,\dots,n_L)$ and, for each $\ell$, a sparse support $E_\ell$ with $|E_\ell|\ll n_\ell^2$. Consider the family of maps realizable by $\mathsf{FFGN}$ with free edge-weights $w_\ell\in\R_{>0}^{E_\ell}$, inter-layer matrices $(A_\ell,B_\ell)$ and convex potentials $\Phi_\ell$ from a class with $O(n_\ell)$ parameters (e.g. separable penalties).

Any dense FFNN whose layers are restricted to affine $W_\ell$ and {separable} activations must use at least
\[
\sum_{\ell=1}^{L} \big(|E_\ell| + \dim \mathcal{A}_\ell\big)
\]
free scalar parameters to {cover} the same intra-layer quadratic family, whereas $\mathsf{FFGN}$ attains it with exactly $\sum_\ell |E_\ell|+\dim\mathcal{A}_\ell + O(\sum_\ell n_\ell)$. If $|E_\ell|=O(n_\ell)$ (geometric sparsity), then $\mathsf{FFGN}$ is linear-parameter in width, while dense FFNNs are quadratic-parameter unless additional structure is imposed.
\end{Theorem}

\begin{proof}
By Lemma~\ref{lem:minimality}, to represent all Laplacians with support $E_\ell$ one needs at least $|E_\ell|$ degrees of freedom. Separable activations cannot encode cross-node couplings; thus any dense FFNN intending to emulate the quadratic form must realize it by (learned) linear operators in the pre/post activations, which contributes at least $|E_\ell|$ {independent} degrees in the family (modulo invariances). The inter-layer affine maps require exactly $\dim\mathcal{A}_\ell$ parameters in both models. Potentials $\Phi_\ell$ from an $O(n_\ell)$-parameter class add linear terms only. Hence the lower bound for dense FFNN, and the matching upper bound for $\mathsf{FFGN}$. The sparsity claim is immediate.
\end{proof}

\subsubsection{Equivalence summary}

We summarize the representational correspondences established above.

\begin{Theorem}[Equivalence summary theorem]
\label{th:master-equivalence}
Under the hypotheses of Definitions~\ref{def:ffnn-class}--\ref{def:sgn-class} and strong monotonicity/uniqueness, the following hold:
\begin{enumerate}[leftmargin=1.4em,label=(M\arabic*)]
\item ({Layerwise equivalences}) $\mathsf{FFNN}_{\mathrm{prox}} \subset \mathsf{FFGN}$ (Theorem~\ref{th:prox-to-ffgn}) and $\mathsf{FFGN}\subset \mathsf{FFNN}_{\mathrm{res}}$ (Theorem~\ref{th:ffgn-to-ffnn}); hence
\[
\mathsf{FFNN}_{\mathrm{prox}} \ \subset\  \mathsf{FFGN}\ =\ \mathsf{FFNN}_{\mathrm{res}}.
\]
\item ({Representational correspondence}) $\mathsf{FFGN}$ and $\mathsf{SGN}$ have the same represented hypothesis class via stacking and KKT constraints (Theorem~\ref{th:ffgn-sgn-bijection}).
\item ({Computational equivalence}) Forward evaluation of the exact formulations is equal across the three classes; parameter gradients computed by layerwise backprop in $\mathsf{FFGN}$ coincide with the exact global adjoint of $\mathsf{SGN}$ (Theorem~\ref{th:adjoint-backprop}). The penalized global relaxation converges to the same primal and adjoint quantities as the penalty parameter tends to infinity.
\item ({Compactness}) For sparse intra-layer graphs (e.g. geometric or causal sparsity), $\mathsf{FFGN}$ offers linear-in-width parameterization while covering families that require (at least) the same dimensionality in dense FFNNs (Theorem~\ref{th:compactness}).
\end{enumerate}
\end{Theorem}

\begin{proof}
(M1) is Theorems~\ref{th:prox-to-ffgn} and \ref{th:ffgn-to-ffnn}. (M2) is Theorem~\ref{th:ffgn-sgn-bijection}. (M3) follows from Theorems~\ref{th:equiv-global-exact}, \ref{th:penalty-exactness}, \ref{th:adjoint-backprop}. (M4) is Theorem~\ref{th:compactness}.
\end{proof}

The operator-theoretic normalization via resolvents $J_{M}$ provides a common language for classical activations (prox maps), graph-stationary layers (resolvents of $L+\partial\Phi$), and the global supra-graph (block-diagonal sum plus linear constraints). Under strong monotonicity the maps are single-valued and differentiable a.e., enabling implicit differentiation and global adjoint formulations. In the intended sparse regime (geometric or causal sparsity), the Laplacian parameterization yields a compact structural description, and the representational correspondences allow the same model to be analyzed in whichever view is most convenient.

What follows introduces a sheaf-theoretic formulation for graph-based layers, shows that directed interactions can be encoded by a unitary connection (``vector potential'') on a cellular sheaf over the {undirected} graph, and proves represented equivalences among four hypothesis classes in the admissible strongly monotone regime:
\begin{enumerate}[leftmargin=1.4em,label=(\roman*)]
\item classical feed-forward networks with resolvent (or proximal) activations;
\item layered graph-stationary networks (each layer a unique stationary point of a strongly monotone energy);
\item a single global stationary system on the supra-graph with linear inter-layer constraints;
\item sheaf-based layers with vector potentials (unitary parallel transport) on an undirected base graph.
\end{enumerate}

\paragraph{Standing linear-algebraic notation.}
All vector spaces are finite-dimensional over $\C$. For a linear map $A$, $A^\ast$ is the Hermitian adjoint, and $\langle x,y\rangle=\sum_i \overline{x_i}y_i$ is the standard inner product. A (set-valued) operator $M$ is {$\mu$--strongly monotone} if $\Re\langle u-v,x-y\rangle\ge \mu\|x-y\|^2$ for all $u\in Mx$, $v\in My$. When $M$ is maximal monotone, its resolvent $J_M=(I+M)^{-1}$ is single-valued and 1-Lipschitz; if moreover $M$ is $\mu$--strongly monotone, $J_M$ is everywhere defined and $(1+\mu)^{-1}$--Lipschitz.

\subsubsection{Cellular Sheaves on a Graph, Connections, and Sheaf Laplacians}

\begin{Definition}[Undirected base graph and orientations]
Let $G=(V,E)$ be a finite, simple, connected undirected graph. Fix an arbitrary orientation of each edge $e=\{i,j\}\in E$ to a directed pair $e:t(e)\to h(e)$; all constructions below do not depend on this choice up to canonical unitary equivalence.
\end{Definition}

\begin{Definition}[Cellular sheaf and restriction maps]
A {cellular sheaf} $\mathcal{F}$ on $G$ consists of finite-dimensional stalks $\mathcal{F}(v)\cong\C^{k_v}$ for $v\in V$ and $\mathcal{F}(e)\cong\C^{k_e}$ for $e\in E$, together with linear restriction maps
\[
\rho_{e\to t(e)}:\mathcal{F}(e)\to\mathcal{F}(t(e)),\qquad
\rho_{e\to h(e)}:\mathcal{F}(e)\to\mathcal{F}(h(e)).
\]
The space of $0$--cochains is $C^0(G;\mathcal{F})=\bigoplus_{v\in V}\mathcal{F}(v)$ and of $1$--cochains is $C^1(G;\mathcal{F})=\bigoplus_{e\in E}\mathcal{F}(e)$.
\end{Definition}

\begin{Definition}[Sheaf coboundary and sheaf Laplacian]
Define $D_{\mathcal{F}}:C^0\to C^1$ by
\[
(D_{\mathcal{F}}x)_e\;:=\;\rho_{e\to h(e)}^\ast x_{h(e)}-\rho_{e\to t(e)}^\ast x_{t(e)},
\qquad x=(x_v)_{v\in V}\in C^0.
\]
For a positive-definite block-diagonal weight $W=\mathrm{diag}(W_e)_{e\in E}$ on $C^1$ (each $W_e\succ 0$ on $\mathcal{F}(e)$), the {sheaf Laplacian} is
\[
L_{\mathcal{F},W}\;:=\;D_{\mathcal{F}}^\ast\,W\,D_{\mathcal{F}}\ \succeq\ 0\quad \text{on }C^0(G;\mathcal{F}).
\]
\end{Definition}

\begin{Lemma}[Block structure and positive semidefiniteness]
For any sheaf $\mathcal{F}$ and $W\succ 0$ as above, $L_{\mathcal{F},W}$ is Hermitian positive semidefinite. Moreover, in coordinates the $v$--diagonal block equals
\[
(L_{\mathcal{F},W})_{vv}\;=\;\sum_{e\sim v}\rho_{e\to v}\, W_e\, \rho_{e\to v}^\ast\ \succeq\ 0,
\]
and for $u\neq v$ the $(u,v)$--block equals $-\sum_{e:u\text{---}v}\rho_{e\to u}\, W_e\, \rho_{e\to v}^\ast$.
\end{Lemma}

\begin{proof}
$L_{\mathcal{F},W}=D^\ast W D$ is manifestly Hermitian psd. The block formulas follow by expanding $D$ and $D^\ast$ with respect to the direct-sum decompositions.
\end{proof}

\begin{Definition}[Unitary connection (vector potential)]
A {unitary connection} on $\mathcal{F}$ is the choice, for each oriented edge $e:t\to h$, of a unitary $U_e:\mathcal{F}(t)\to \mathcal{F}(h)$. We encode it by modifying the coboundary to
\[
(D_{\mathcal{F},U}x)_e\;:=\; x_{h(e)} - U_e x_{t(e)}\quad\text{when }\mathcal{F}(e)=\mathcal{F}(h)=\mathcal{F}(t)
\]
and, in the general sheaf, by {twisting} the restriction maps:
\[
\tilde\rho_{e\to h(e)}:=\rho_{e\to h(e)},\qquad
\tilde\rho_{e\to t(e)}:=\rho_{e\to t(e)}\, U_e^{-1}.
\]
The corresponding Laplacian is $L_{\mathcal{F},W,U}:=D_{\mathcal{F},U}^\ast W D_{\mathcal{F},U}$.
\end{Definition}

\begin{Definition}[Gauge transformation]
A {gauge} is a tuple of unitaries $G=(G_v)_{v\in V}$ with $G_v:\mathcal{F}(v)\to\mathcal{F}(v)$. It acts on a connection by
\[
U_e\ \mapsto\ U_e^{(G)} := G_{h(e)}\,U_e\,G_{t(e)}^{-1},\quad e\in E,
\]
and on $0$--cochains by $x\mapsto Gx=(G_v x_v)_v$.
\end{Definition}

\begin{Lemma}[Gauge invariance of energies]
For any $x\in C^0$,
\[
\langle x,\ L_{\mathcal{F},W,U}\,x\rangle
\;=\;\langle Gx,\ L_{\mathcal{F},W,U^{(G)}}\,(Gx)\rangle.
\]
Consequently, the spectra of $L_{\mathcal{F},W,U}$ and $L_{\mathcal{F},W,U^{(G)}}$ coincide, and minimizers of convex energies $x\mapsto \tfrac12\langle x,L_{\mathcal{F},W,U}x\rangle+\Phi(x)-\Re\langle b,x\rangle$ are related by $x^\star \mapsto Gx^\star$ under the corresponding gauge-transformed problem with data $(U^{(G)},\Phi\circ G^{-1},Gb)$.
\end{Lemma}

\begin{proof}
$D_{\mathcal{F},U^{(G)}}\,G = \tilde G\, D_{\mathcal{F},U}$ where $\tilde G$ is the block-diagonal unitary on $C^1$ induced by $G$ on edge-stalks; hence
\[
\langle x,D^\ast W D x\rangle
= \langle Dx, W Dx\rangle
= \langle \tilde G D x, W \tilde G D x\rangle
= \langle D^{(G)} Gx,\, W\, D^{(G)} Gx\rangle,
\]
since $\tilde G^\ast W \tilde G=W$ (unitary) and $D^{(G)}=D_{\mathcal{F},U^{(G)}}$. The claims follow.
\end{proof}

\subsubsection{Directed Layers as Twisted Sheaf Diffusions on the Undirected Graph}

We now show that the {directed} Laplacians that arise in graph layers can be represented as sheaf Laplacians on the {undirected} base graph with a suitable unitary connection, up to a fixed isometry. This yields an exact factorization of resolvents and stationary solutions.

\begin{Definition}[Directed block operator]
Let $\widehat G=(V,\overrightarrow{E})$ be a directed graph obtained by orienting each undirected edge in both directions; let $\widehat W=\mathrm{diag}(w_{uv}I)$ weight each arc $(u\to v)$ with $w_{uv}>0$. The {directed incidence} is $(D_{\mathrm{dir}}x)_{u\to v}=x_v - x_u$, and the {directed Laplacian} is $L_{\mathrm{dir}}:=D_{\mathrm{dir}}^\ast\,\widehat W\, D_{\mathrm{dir}}$ on $\C^{|V|}$.
\end{Definition}

\begin{Lemma}[Unitary compression representation]
\label{lem:unitary-compression}
Let $G=(V,E)$ be the undirected base graph and build the {arc-sheaf} $\mathcal{F}_{\mathrm{arc}}$ with stalks $\mathcal{F}(v)=\C$ and edge-stalks $\mathcal{F}(e)=\C^2$, with restriction maps
\[
\rho_{e\to t(e)}(a,b)=a,\qquad \rho_{e\to h(e)}(a,b)=b.
\]
Let $W_e=\mathrm{diag}(w_{t(e)\to h(e)},\,w_{h(e)\to t(e)})\succ 0$ and define the unitary $J:\C^{|V|}\to C^0(G;\mathcal{F}_{\mathrm{arc}})$ by duplication $(Jx)_v=x_v$ (identity). Then there exists a unitary $P:C^1(G;\mathcal{F}_{\mathrm{arc}})\to \C^{2|E|}$ mapping edge-cochains to arc-values such that
\[
L_{\mathrm{dir}}\ =\ J^\ast\, D_{\mathcal{F}_{\mathrm{arc}}}^\ast\, P^\ast\, \widehat W\, P\, D_{\mathcal{F}_{\mathrm{arc}}}\, J.
\]
\end{Lemma}

\begin{proof}
By construction, $D_{\mathcal{F}_{\mathrm{arc}}}$ takes $x\in\C^{|V|}$ to the stack of differences $(x_{h(e)}-x_{t(e)},\ x_{t(e)}-x_{h(e)})_{e\in E}\in \bigoplus_e \C^2$. The unitary $P$ that permutes the second component to the coordinate labeled by the reverse arc identifies $\bigoplus_e \C^2$ with $\C^{2|E|}$ ordered by arcs; under $P$, $D_{\mathcal{F}_{\mathrm{arc}}}$ becomes the directed incidence $D_{\mathrm{dir}}$. Hence $D_{\mathrm{dir}}=P D_{\mathcal{F}_{\mathrm{arc}}} J$, which yields the identity for $L_{\mathrm{dir}}$.
\end{proof}

\begin{Theorem}[Directed resolvents factor through a sheaf resolvent]
\label{th:dir-factors-through-sheaf}
Let $\Phi:\C^{|V|}\to \R\cup\{+\infty\}$ be proper, closed, convex with $\mu$--strongly monotone subdifferential, and consider the directed-layer stationary map
\[
S_{\mathrm{dir}}(b)\;:=\;\arg\min_{x\in\C^{|V|}}\ \tfrac12\langle x,L_{\mathrm{dir}}\,x\rangle+\Phi(x)-\Re\langle b, x\rangle.
\]
Let $\mathcal{F}_{\mathrm{arc}},W,P,J$ be as in Lemma~\ref{lem:unitary-compression}. Define the sheaf energy on $C^0(G;\mathcal{F}_{\mathrm{arc}})$:
\[
\mathcal{E}_{\mathrm{sheaf}}(y;b)\ :=\ \tfrac12\langle y,\ L_{\mathcal{F}_{\mathrm{arc}},\widetilde W}\ y\rangle + \Phi(J^\ast y)-\Re\langle b, J^\ast y\rangle
\]
with $\widetilde W:=P^\ast \widehat W P$ (block-diagonal positive definite). Then
\[
S_{\mathrm{dir}}(b)\ =\ J^\ast\, \arg\min_{y}\ \mathcal{E}_{\mathrm{sheaf}}(y;b).
\]
Equivalently, in resolvent form
\[
S_{\mathrm{dir}}(b) \;=\; J^\ast\,\Big(I + L_{\mathcal{F}_{\mathrm{arc}},\widetilde W} + J\,\partial\Phi\,J^\ast\Big)^{-1}\, J\, b.
\]
\end{Theorem}

\begin{proof}
By Lemma~\ref{lem:unitary-compression}, $\tfrac12\langle x,L_{\mathrm{dir}}x\rangle = \tfrac12\langle Jx, D^\ast \widetilde W D Jx\rangle$ with $D=D_{\mathcal{F}_{\mathrm{arc}}}$. The affine term satisfies $\Re\langle b,x\rangle=\Re\langle b,J^\ast Jx\rangle=\Re\langle Jb, Jx\rangle$, but we keep $b$ in the base space and couple it via $J^\ast$. Therefore
\[
\min_{x}\ \tfrac12\langle x,L_{\mathrm{dir}}x\rangle+\Phi(x)-\Re\langle b,x\rangle
= \min_{y=Jx}\ \tfrac12\langle y, D^\ast \widetilde W D y\rangle+\Phi(J^\ast y)-\Re\langle b,J^\ast y\rangle,
\]
which is exactly $\min_y \mathcal{E}_{\mathrm{sheaf}}(y;b)$. Since $\partial\Phi$ is $\mu$--strongly monotone, $L_{\mathrm{dir}}+\partial\Phi$ is $\mu$--strongly monotone and both minimizers are unique, yielding the equality of argmins and resolvent forms.
\end{proof}

\begin{Remark}[Orientation erasure and post-processing]
Theorem~\ref{th:dir-factors-through-sheaf} shows that the {directed} stationary solution equals a fixed linear post-processing ($J^\ast$) of the {undirected} sheaf-diffusion stationary solution in an enlarged state space (the arc-sheaf domain). Thus, orientation information is completely captured by the choice of sheaf structure and weights; the base graph can remain undirected.
\end{Remark}

\subsubsection{Scalar Potentials on Disconnected Copies vs Vector Potentials (Gauge Equivalence)}

We next prove that, up to a gauge, {vector} potentials (unitary connections) on a single vertex stalk are equivalent to multiple {scalar} potentials on disconnected copies, and that inter-copy edges implement the mixing induced by the vector potential.

\begin{Definition}[Disjoint-copy lift]
Given $m\in\N$ and a base graph $G=(V,E)$, define the disjoint lift $G^{\sqcup m}$ with vertex set $V\times[m]$ and edge set $E\times [m]$ (each copy independent). A scalar potential $\Phi:\C^{|V|}\to\R\cup\{+\infty\}$ lifts to $\Phi^{\sqcup m}(x^{(1)},\dots,x^{(m)})=\sum_{r=1}^m \Phi(x^{(r)})$.
\end{Definition}

\begin{Definition}[Vector-potential sheaf]
Fix $m\in\N$ and let $\mathcal{F}(v)=\C^{m}$ for each $v$, $\mathcal{F}(e)=\C^{m}$, with restriction maps $\rho_{e\to t(e)}=\rho_{e\to h(e)}=I_m$. A unitary connection $U_e\in \mathsf{U}(m)$ along each edge gives the twisted coboundary $(D_U x)_e = x_{h(e)} - U_e x_{t(e)}$ and Laplacian $L_U=D_U^\ast W D_U$.
\end{Definition}

\begin{Lemma}[Fourier--mode decoupling for constant connection]
\label{lem:mode-decouple}
If all $U_e=U\in\mathsf{U}(m)$ are the same unitary and $U$ is diagonalizable as $U=Q^\ast \mathrm{diag}(e^{i\theta_1},\dots,e^{i\theta_m}) Q$, then under the unitary change of basis $x_v\mapsto Q x_v$ the Laplacian $L_U$ splits into $m$ independent scalar Laplacians
\[
\big(L_U x\big)_v\ =\ \sum_{r=1}^m \left(\Delta_{w^{(r)}} x^{(r)}\right)_v \otimes e_r,
\]
with edge-weights $w^{(r)}(e)=\langle e_r,\, W_e\, e_r\rangle$ and phases absorbed into the incidence through $e^{i\theta_r}$ which cancel in $D^\ast W D$.
\end{Lemma}

\begin{proof}
Compute $(D_U x)_e = Q^\ast (\tilde x_{h}- \Lambda \tilde x_{t})$ with $\tilde x_v=Q x_v$ and $\Lambda=\mathrm{diag}(e^{i\theta_r})$. Then $D_U^\ast W D_U = \sum_r D_r^\ast W_r D_r$ with $(D_r \tilde x)_e = \tilde x_{h}^{(r)} - e^{i\theta_r} \tilde x_{t}^{(r)}$ and $W_r = \langle e_r, W_e e_r\rangle$ on each $e$. The phase cancels after $D_r^\ast W_r D_r$ since $D_r^\ast$ carries conjugation $e^{-i\theta_r}$, yielding the standard scalar Laplacian with weights $W_r$.
\end{proof}

\begin{Theorem}[Equivalence: vector potential vs disjoint scalar copies]
\label{th:vector-vs-disjoint}
Let $U_e\in\mathsf{U}(m)$ be a unitary connection on the vector-potential sheaf and $\Phi:\C^{|V|}\to\R\cup\{+\infty\}$ convex, separable across coordinates. Then there exists a gauge $G=(G_v)_v$ and an isometry $\Xi:\big(C^0(G;\C)\big)^{\oplus m}\to C^0(G;\C^m)$ such that, for every source $b\in C^0(G;\C^m)$,
\[
\arg\min_{x}\ \tfrac12\langle x, L_U x\rangle + \Phi^\oplus(x) - \Re\langle b,x\rangle
\;=\;
\Xi\ \arg\min_{(x^{(1)},\dots,x^{(m)})}\ \sum_{r=1}^m \Big(\tfrac12\langle x^{(r)}, \Delta_{w^{(r)}} x^{(r)}\rangle + \Phi(x^{(r)}) - \Re\langle \tilde b^{(r)}, x^{(r)}\rangle\Big),
\]
where $(w^{(r)})_r$ and $(\tilde b^{(r)})_r$ are determined by the gauge-diagonalization of $U$ as in Lemma~\ref{lem:mode-decouple}, and $\Phi^\oplus$ is the separable lift of $\Phi$ to $\C^{m}$ at each vertex. Consequently, the vector-potential problem is unitarily equivalent to $m$ independent scalar-potential problems on the disjoint lift $G^{\sqcup m}$.
\end{Theorem}

\begin{proof}
Apply a vertex-wise gauge to simultaneously block-diagonalize $U_e$ in a fixed basis (possible since $U_e$ live in the same compact group; when they differ, block-diagonalize fiberwise and use the direct-sum argument below). In each block where the connection is constant along edges, Lemma~\ref{lem:mode-decouple} yields decoupling into scalar Laplacians. When $U_e$ vary with $e$, decompose $\C^m=\bigoplus_\alpha \mathcal{E}_\alpha$ into minimal $U$--invariant subspaces (simultaneous unitary representation). The energy splits across $\alpha$ because $L_U$ is block-diagonal in that decomposition. Each block reduces to the previous constant-connection case after choosing any local eigenbasis along the block; phases cancel in $D^\ast W D$. The isometry $\Xi$ is the inverse of the stacking that assembles the $m$ scalar fields into a single vector field; it is unitary because the decomposition is orthonormal. Separable $\Phi^\oplus$ and the linear term split accordingly. Uniqueness of minimizers (from strong monotonicity after adding a small ridge if needed) gives the equality of argmins.
\end{proof}

\begin{Corollary}[Edges as couplers for vector coordinates]
\label{cor:edges-as-couplers}
Inter-coordinate mixing induced by a vector potential is equivalent to adding {inter-copy} edges between the $m$ disjoint copies with weights inherited from $W$ after the unitary change of basis. Thus, increasing stalk dimension (vector potential) is equivalent to increasing the number of disconnected copies and then introducing coupler edges among copies.
\end{Corollary}

\begin{proof}
In the proof of Theorem~\ref{th:vector-vs-disjoint}, the change of basis converts the vector Laplacian into a block-diagonal sum of scalar Laplacians on independent copies; coupler edges are precisely the off-diagonal terms that would appear in a non-diagonal basis. Choosing the diagonal basis eliminates them; conversely, introducing couplers reconstructs the original $U$.
\end{proof}

\subsubsection{Sheaf-Based Layers and Equivalence with Graph-Stationary Layers} For topological and sheaf-theoretic signal processing perspectives see \cite{Barbarossa2020}.

\begin{Definition}[Sheaf layer with convex potential]
Fix a sheaf $(\mathcal{F},U)$ on $G$, edge weight $W\succ 0$, and a proper, closed, convex function $\Phi:C^0(G;\mathcal{F})\to \R\cup\{+\infty\}$ with $\mu$--strongly monotone subdifferential. The {sheaf layer} maps $b\in C^0(G;\mathcal{F})$ to
\[
S_{\mathrm{sheaf}}(b)\ :=\ \arg\min_{x\in C^0}\ \tfrac12\langle x,\ L_{\mathcal{F},W,U}\,x\rangle + \Phi(x) - \Re\langle b,x\rangle
\ =\ \big(I+L_{\mathcal{F},W,U}+\partial\Phi\big)^{-1}(b).
\]
\end{Definition}

\begin{Lemma}[Graph layer as a special sheaf layer]
\label{lem:graph-is-sheaf}
Let a graph-stationary layer be given by the (undirected) Laplacian $L=\Delta(w)$ on $G$, with scalar stalks and potential $\phi:\C\to\R\cup\{+\infty\}$ separable across vertices. Then it is a sheaf layer with $\mathcal{F}(v)=\C$, $\mathcal{F}(e)=\C$, restrictions $\rho_{e\to v}=1$, $U\equiv I$, and $\Phi(x)=\sum_{v}\phi(x_v)$.
\end{Lemma}

\begin{proof}
With the stated choices, $D_{\mathcal{F}}$ is the standard incidence, $L_{\mathcal{F},W}=\Delta(w)$, and the energy equals that of the graph layer.
\end{proof}

\begin{Theorem}[Equivalence between sheaf layers and graph-stationary layers]
\label{th:sheaf-graph-equivalence}
For any sheaf layer $S_{\mathrm{sheaf}}$ with unitary connection $U$, there exist a linear isometry $J$ and a graph-stationary layer $S_{\mathrm{graph}}$ on a (possibly enlarged) undirected graph $\widetilde G$ with scalar stalks, such that
\[
S_{\mathrm{sheaf}}(b)\ =\ J\, S_{\mathrm{graph}}(J^\ast b).
\]
Conversely, any graph-stationary layer is a sheaf layer (Lemma~\ref{lem:graph-is-sheaf}).
\end{Theorem}

\begin{proof}
Use Theorem~\ref{th:dir-factors-through-sheaf} in reverse: $L_{\mathcal{F},W,U}$ can be written as $J L_{\mathrm{dir}} J^\ast$ for an appropriate arc embedding $J$ on an undirected base, and Theorem~\ref{th:vector-vs-disjoint} further reduces $L_{\mathrm{dir}}$ to a block-diagonal sum of scalar Laplacians on a disjoint union $\widetilde G=G^{\sqcup m}$ after a unitary change of basis. The convex separable $\Phi$ follows the same isometry. Therefore the sheaf resolvent equals the post-processing of a graph resolvent on $\widetilde G$.
\end{proof}

\subsubsection{Sheaf Feed-Forward Networks and Represented Equivalence}

\begin{Definition}[Sheaf feed-forward network (SFFN)]
Fix depth $L$. Each layer $\ell$ is a sheaf layer $x\mapsto (I+L_{\mathcal{F}_\ell,W_\ell,U_\ell}+\partial\Phi_\ell)^{-1}(B_\ell q_\ell + d_\ell)$ with linear inter-layer map $q_{\ell+1}=A_\ell x + c_\ell$ and final readout $C\psi^L+c$. The hypothesis class is denoted $\mathsf{SFFN}$.
\end{Definition}

\begin{Lemma}[Layerwise resolvent identity]
\label{lem:resolvent-identity}
Each SFFN layer is the resolvent of a maximal $\mu$--strongly monotone operator $M_\ell:=L_{\mathcal{F}_\ell,W_\ell,U_\ell}+\partial\Phi_\ell$, hence a valid activation in the resolvent-FFNN class. Conversely, any resolvent activation $J_{M_\ell}$ with $M_\ell$ linear-plus-subdifferential admits a sheaf representation.
\end{Lemma}

\begin{proof}
$L_{\mathcal{F},W,U}$ is Hermitian psd; adding $\partial\Phi_\ell$ with $\mu$--strong monotonicity yields a maximal $\mu$--strongly monotone operator. The resolvent is single-valued and 1-Lipschitz. Conversely, any linear self-adjoint psd operator is a sheaf Laplacian $D^\ast W D$ for some sheaf (choose $D$ as a Cholesky factor; realize it as a coboundary by introducing auxiliary edge-stalks), and any convex subdifferential is a separable potential in the stalk coordinates.
\end{proof}

\begin{Theorem}[Represented equivalence among FFNN\textsubscript{res}, FFGN, SGN, and SFFN]
\label{th:super-bijection}
Assume for each layer $\ell$ the subdifferential $\partial\Phi_\ell$ is $\mu_\ell$--strongly monotone for some $\mu_\ell>0$, and inter-layer maps are affine. Then the corresponding represented hypothesis classes coincide:
\[
\mathcal H(\mathsf{FFNN}_{\mathrm{res}})\ =\ \mathcal H(\mathsf{FFGN})\ =\ \mathcal H(\mathsf{SGN})\ =\ \mathcal H(\mathsf{SFFN}).
\]
Explicitly:
\begin{enumerate}[leftmargin=1.4em,label=(E\arabic*)]
\item ({FFNN\textsubscript{res} $\to$ SFFN}) Each resolvent activation $J_{L+\partial\Phi}$ is an SFFN sheaf layer by taking a sheaf with coboundary factorization $L=D^\ast W D$ and potential $\Phi$.
\item ({SFFN $\to$ FFGN}) By Theorem~\ref{th:sheaf-graph-equivalence}, each sheaf layer equals an isometric post-processing of a graph-stationary layer on a (possibly enlarged) undirected graph; inter-layer affine maps commute with the isometry.
\item ({FFGN $\leftrightarrow$ SGN}) Stacking per-layer energies with exact linear constraints yields a one-shot convex KKT system whose unique solution reproduces the layered fixed points; conversely, any such KKT program decomposes into layers by reading off the block structure.
\item ({FFGN $\to$ FFNN\textsubscript{res}}) Each graph layer is the resolvent of $L+\partial\Phi$ (maximal strongly monotone), hence a legitimate activation.
\end{enumerate}
\end{Theorem}

\begin{proof}
(E1) is Lemma~\ref{lem:resolvent-identity}. (E2) is Theorem~\ref{th:sheaf-graph-equivalence}. (E3) follows by writing the global energy $\sum_\ell\big(\tfrac12\langle x^\ell,L_\ell x^\ell\rangle+\Phi_\ell(x^\ell)-\Re\langle B_\ell q_\ell + d_\ell,x^\ell\rangle\big)$ with constraints $q_{\ell+1}=A_\ell x^\ell+c_\ell$, whose KKT conditions match layerwise stationarity; uniqueness holds by strong monotonicity. (E4) is the layer-resolvent representation $J_{L+\partial\Phi}$.
\end{proof}

\begin{Theorem}[Parameter compactness with sheaves]
\label{th:compactness-sheaf}
Suppose each layer has a sparse base graph with $|E_\ell|=O(n_\ell)$ and stalk dimension $m_\ell=O(1)$. Then sheaf layers require $O(|E_\ell| m_\ell^2)$ parameters to specify $W_\ell$ and $U_\ell$ (edge-local blocks) plus $O(n_\ell m_\ell)$ for separable potentials, i.e., linear in width. Any dense FFNN with separable activations that realizes the same family of quadratic forms must use $\Omega(n_\ell^2)$ parameters at that layer unless it encodes the same sparsity explicitly.
\end{Theorem}

\begin{proof}
$W_\ell$ contributes $\sum_{e\in E_\ell}\tfrac{m_\ell(m_\ell+1)}{2}=O(|E_\ell| m_\ell^2)$ real degrees; each $U_e\in \mathsf{U}(m_\ell)$ contributes $m_\ell^2$ real parameters but these can be partially gauged away (vertex-wise unitaries), leaving $O(|E_\ell|m_\ell^2)$ effective parameters. Potentials add $O(n_\ell m_\ell)$ under separability. A dense FFNN with separable activation can only realize cross-node quadratic interactions through its linear layers, which require at least as many independent parameters as the dimension of the Laplacian cone on the dense graph, i.e., $\Omega(n_\ell^2)$.
\end{proof}

\paragraph{Consequences and Interpretation.}
The results above show that, within the orientable and admissible-lift regime analyzed in this appendix:
\begin{itemize}
\item directed stationary layers can be represented through undirected sheaf layers with unitary connection, so orientation need not be treated as a separate primitive at the represented level in this regime;
\item vector-valued stalks admit equivalent lifted scalar realizations after the gauge/spectral reductions used above, with optional inter-copy couplings recovering the induced transport structure;
\item sheaf layers, graph-stationary layers, resolvent-activation FFNNs, and global supra-graph solvers have the same represented hypothesis class under the finite-dimensional strongly monotone admissible-lift assumptions used in this appendix.
\end{itemize}

\section{Proofs for Section~\ref{sec:statistical}: full statistical theory}\label{app:gen}
\paragraph{Proof traceability.} This appendix proves Proposition~\ref{prop:main-complexity}, Theorem~\ref{thm:main-stat}, and Corollary~\ref{cor:main-sparse-regimes}.

\subsection{Generalization bounds}
\paragraph{Goal and scope.}
We develop a self-contained theory showing that {learning the interaction structure} (the edge set of a latent graph, and, in the two-layer variant, the supra-graph) can yield {sharper} generalization bounds than non-structural baselines (dense fully-connected maps and dense self-attention--type interactions) when the learned support is sparse, under the same loss, optimization schedules, and parameter constraints fixed earlier in the paper.
The improvement is established through three complementary families of bounds:
\begin{itemize}[leftmargin=1.2em]
\item[(i)] {PAC--Bayes bounds} with a {structure-coding prior} on edge sets, which shrink as the algorithm selects sparse edge sets;
\item[(ii)] {Uniform-stability/Lipschitz bounds} whose constants depend on the {maximum degree} of the learned graph and on the {Gromov--Hausdorff} distortion of the learned metric (continuous case), as opposed to $N$ or $N^2$ scaling for dense models;
\item[(iii)] {Rademacher-complexity bounds} whose leading term depends on the number of {active interactions} $|E_T|$ (or active supra-edges), rather than all $O(N^2)$ pairs.
\end{itemize}
We then integrate these structure-aware bounds~\cite{VanDerVaartWellner1996} with {geometric} (manifold) and {causal} (CPDAG~\cite{Spirtes2000,Meek1995}) settings, so that the complexity terms track intrinsic geometric/topological quantities in the continuous regime and causal sparsity/orientability in the discrete regime.

\medskip
\noindent
The remainder of this section is organized as follows: we first state PAC--Bayes~\cite{McAllester1998,Catoni2007} bounds driven by structure codes; then derive Lipschitz/stability controls via degree and GH distortion; then give Rademacher-type bounds in geometric and causal regimes; and finally synthesize these into consolidated generalization inequalities with compact summary tables. Throughout, connective remarks clarify how bounds interact and how learned sparsity tightens them compared to dense baselines.

\subsubsection{PAC--Bayes bounds with structure coding}

\paragraph{Coding the structure.}
Let $V$ be the fixed vertex set, $N=|V|$, and $P=\binom{N}{2}$ the number of undirected pairs. 
Encode each edge set $E\subseteq \binom{V}{2}$ by a prefix-free code of length
\begin{equation}\label{eq:code-length}
L(E)\ \le\ |E|\log\!\frac{e\,P}{|E|}\ +\ 2,
\end{equation}
the standard combinatorial code length for subsets of a size-$P$ ground set. 
Define a prior on hypotheses $(E,w,A_1,b_1,a_3,b_3)$ by
\[
\Pi(E)\ \propto\ 2^{-L(E)},\qquad
\Pi(\text{parameters}\mid E)\ \text{uniform on the parameter boxes}.
\]
Let $Q$ be the (degenerate) posterior supported on the learner’s output $(E_T,w_T,A_1,b_1,a_3,b_3)$.

\begin{Theorem}[PAC--Bayes with structure coding]\label{th:pacbayes-structure}
Fix $\delta\in(0,1)$. With probability at least $1-\delta$ over $S\sim\D^M$ and the algorithm randomness,
\begin{equation}\label{eq:pacbayes-generic}
R(f_T)\ \le\ \widehat R_S(f_T)\;\; +\;
\sqrt{\frac{\KL(Q\Vert \Pi)+\ln\frac{2\sqrt{M}}{\delta}}{2M}}
\;\; +\;\frac{1}{M}.
\end{equation}
On the identification event $E_T=E_{\mathrm{true}}$ (Theorem~\ref{thm:ident}), 
\begin{equation}\label{eq:KL-structure}
\KL(Q\Vert \Pi)\ \le\ C_0\,|E_{\mathrm{true}}|\,\log\!\frac{e\,P}{|E_{\mathrm{true}}|}\ +\ C_1,
\end{equation}
where $C_0,C_1$ depend only on the fixed parameter boxes and not on $N$. Consequently,
\[
R(f_T)\ \le\ \widehat R_S(f_T)\;\; +\;
\sqrt{\frac{C_0\,|E_{\mathrm{true}}|\,\log\!\frac{e\,P}{|E_{\mathrm{true}}|}\ +\ \ln\frac{2\sqrt{M}}{\delta}}{2M}}
\;\; +\;O\!\Big(\frac{1}{M}\Big).
\]
For any dense baseline with a fixed full edge set ($|E|=P$) under the {same} loss and parameter boxes,
\[
R(f^{\mathrm{dense}})\ \le\ \widehat R_S(f^{\mathrm{dense}})\;\; +\;
\sqrt{\frac{C_0\,P\ +\ \ln\frac{2\sqrt{M}}{\delta}}{2M}}\;\; +\;O\!\Big(\frac{1}{M}\Big).
\]
Hence, whenever $|E_{\mathrm{true}}|\ll P$, the structural term is smaller than in the dense comparison bound.
\end{Theorem}

\begin{proof}
Inequality \eqref{eq:pacbayes-generic} is a standard PAC--Bayes bound for bounded losses (Catoni/Seeger form), see classical references; we use the simplest sub-Gaussian version and keep $1/M$ explicitly. Since $Q$ is supported on the learned tuple, 
\(
\KL(Q\Vert\Pi)= -\ln \Pi(E_T)+\KL(\text{param posterior}\Vert \text{uniform box}).
\)
The second term is a constant $C_1$ depending only on the parameter-box volumes. By the prior definition, $-\ln \Pi(E_T)\le L(E_T)\ln 2 + O(1)$. With $E_T=E_{\mathrm{true}}$ and \eqref{eq:code-length}, we obtain \eqref{eq:KL-structure} (absorbing $\ln 2$ into $C_0$). Substituting this into \eqref{eq:pacbayes-generic} yields the claimed bound; the dense case follows by setting $|E|=P$.
\end{proof}

\paragraph{Causal PAC--Bayes with CPDAG prior.}
Define $\mathsf{CDL}(G):=|E(G)|\log\!\frac{e\,\binom{d}{2}}{|E(G)|}+\log|\![G]\!|$ (skeleton code length plus orientation multiplicity~\cite{Rissanen1978,Grunwald2007}) and the prior
\[
\Pi(G)\ \propto\ 2^{-\mathsf{CDL}(G)},\qquad 
\Pi(\vartheta\mid G)\ \text{uniform on the fixed parameter boxes}.
\]

\begin{Theorem}[Causal PAC--Bayes bound with CPDAG prior]
\label{th:pacbayes-causal}
Assume the $L_\ell$--Lipschitz bounded loss. Let $Q$ be any posterior supported on $(G_T,\vartheta_T)$ learned by the algorithm.  
Then for any $\delta\in(0,1)$, with probability at least $1-\delta$ over $S\sim\D^M$ and the algorithm randomness,
\begin{equation}\label{eq:pacbayes-causal-ineq}
R(f_T)\ \le\ \widehat R_S(f_T)\;\; +\;
\sqrt{\frac{\KL(Q\Vert\Pi)+\ln\frac{2\sqrt{M}}{\delta}}{2M}}
\;\; +\;\frac{1}{M}.
\end{equation}
Moreover, on the identification event of Theorem~\ref{th:ours-causal},
\begin{equation}\label{eq:KL-causal-CDL}
\KL(Q\Vert\Pi)\ \le\ C_0\,\mathsf{CDL}(G^\star)\;\; +\;C_1
\ =\ C_0\!\left(|E^\star|\log\!\frac{e\,\binom{d}{2}}{|E^\star|}\,+\,\log|\![G^\star]\!|\right)+C_1,
\end{equation}
with $C_0,C_1$ independent of $d$.  
\end{Theorem}

\begin{proof}
For bounded losses in $[0,1]$, the same PAC--Bayes inequality as in Theorem~\ref{th:pacbayes-structure} applies for any choice of prior/posterior. Since $Q$ is supported on a single hypothesis $(G_T,\vartheta_T)$,
\[
\KL(Q\Vert\Pi)=-\ln\Pi(G_T)-\ln\Pi(\vartheta_T\mid G_T)\le \mathsf{CDL}(G_T)\ln 2 + C_1.
\]
On the identification event of Theorem~\ref{th:ours-causal}, $G_T$ equals the CPDAG of $G^\star$, hence $\mathsf{CDL}(G_T)=\mathsf{CDL}(G^\star)$ and we obtain \eqref{eq:KL-causal-CDL} after absorbing $\ln 2$ into $C_0$. Substituting into \eqref{eq:pacbayes-causal-ineq} gives the result.
\end{proof}

\paragraph{SRM~\cite{Vapnik1998} over visited strata (remark).}
If identification is not yet complete, one may union-bound over a finite set of visited strata $\mathcal{M}_{\mathrm{eff}}$, adding only $\ln|\mathcal{M}_{\mathrm{eff}}|$ in the numerator of \eqref{eq:pacbayes-generic}; the dominant dependence remains through $|E_T|$.

\subsubsection{Lipschitz and stability via degree and GH distortion}

\begin{Lemma}[Operator control by degree and GH]
\label{lem:lip-degree}
Let $G_T=(V,E_T,w_T)$ be the learned graph on a terminal stratum. 
Then the Schr\"odinger layer $L_2$ satisfies
\[
\mathrm{Lip}(L_2)\ \le\ C_{\mathrm{stab}}\ \|\Delta(w_T)\|
\ \le\ C_{\mathrm{stab}}\big(\deg_{\max}(G_T)+\|w_T\|_\infty\,\deg_{\max}(G_T)\big),
\]
where $\Delta(w_T)$ is the weighted Laplacian and $C_{\mathrm{stab}}$ depends only on the uniform exponential-stability gap~\cite{Chung1997}. 
If $(V,d_{G_T})$ is $(1\pm\varepsilon)$-bi-Lipschitz~\cite{Federer1969} to $(\mathcal{G},d_{\mathcal{G}})$, then for the full predictor $f=L_3\!\circ\!L_2\!\circ\!L_1$,
\[
\mathrm{Lip}_{\mathcal G}(f)\ \le\ C(\varepsilon)\ L_1L_2L_3,\qquad C(\varepsilon)\to 1\;\text{as}\;\varepsilon\to 0.
\]
\end{Lemma}

\begin{proof}
For a finite graph, $\|\Delta(w)\|\le \deg_{\max}\|w\|_\infty+\deg_{\max}$. 
Exponential stability of the nonlinear flow yields a bounded Fr\'echet derivative for the stationary map (input $\mapsto$ stationary state), hence a Lipschitz bound proportional to $\|\Delta(w)\|$. 
Bi-Lipschitz equivalence of metrics transfers Lipschitz constants between $(V,d_{G_T})$ and $(\mathcal{G},d_{\mathcal{G}})$ up to a factor $C(\varepsilon)$, yielding the second claim.
\end{proof}

\begin{Theorem}[Uniform replace-one stability]\label{th:stab}
For ERM (or projected SGD converging to a terminal minimizer) with $L_\ell$--Lipschitz loss on bounded parameter boxes, the uniform stability constant satisfies
\[
\beta_{\mathrm{struct}}\ \le\ \frac{C\,\mathrm{Lip}(f)}{M}
\ \lesssim\ \frac{C'\,\deg_{\max}(G_T)}{M}.
\]
For dense fully-connected (or dense self-attention with $O(N^2)$ nonzeros) models under the same parameter boxes, 
\[
\beta_{\mathrm{dense}}\ \gtrsim\ \frac{c\,N}{M},\qquad
\beta_{\mathrm{attn}}\ \gtrsim\ \frac{c'\,N^2}{M}.
\]
\end{Theorem}

\begin{proof}
Bousquet--Elisseeff stability gives~\cite{BousquetElisseeff2002,HardtRechtSinger2016} $\beta\lesssim L_\ell\,\mathrm{Lip}(f)/M$; constants from the parameter boxes are absorbed. Combine with Lemma~\ref{lem:lip-degree}. For dense maps, the relevant operator norms scale at least linearly with $N$ (fully-connected) or with the number of nonzeros ($N^2$ for dense attention) under the same per-weight bounds, yielding the lower bounds.
\end{proof}

\paragraph{Causal locality and Lipschitz reduction.}
\begin{Lemma}[Local Lipschitz by in--degree]\label{lem:lip-causal}
Consider the causal (DAG--aligned) version of our architecture. Let $G=(V,E)$ be a DAG on $d=|V|$ nodes with maximum in--degree
\(
\Delta_{\max}:=\max_{j}\,|\mathrm{Pa}_{G}(j)|.
\)
Let $f=L_3\circ L_2\circ L_1$ be the predictor, where $L_1,L_3$ are Euclidean Lipschitz with constants $L_1,L_3$, and the Schr\"odinger--type layer $L_2$ maps node features $z\in\mathbb{R}^{d\times q}$ to $L_2(z)\in\mathbb{R}^{d\times q}$ by a causal, parentwise interaction rule (each node $j$ depends only on $z_j$ and $\{z_i:i\in\mathrm{Pa}_{G}(j)\}$). Assume further that on the parameter box the Fr\'echet derivative of $L_2$ w.r.t.\ its input exists and satisfies a uniform per--edge bound
\[
\big\|\partial_{z_i}(L_2(z))_j\big\|_{\mathrm{op}}\ \le\ C_{\mathrm{edge}}
\qquad\text{for all } i=j \text{ or } i\in\mathrm{Pa}_{G}(j),
\]
and is zero otherwise (causal sparsity). Then the input--output Lipschitz constant of $L_2$ obeys
\[
\mathrm{Lip}(L_2)\ \le\ (\Delta_{\max}+1)\,C_{\mathrm{edge}}.
\]
Consequently, the full predictor satisfies
\[
\mathrm{Lip}(f)\ \le\ L_3\,(\Delta_{\max}+1)\,C_{\mathrm{edge}}\,L_1
\ \le\ C\,(\Delta_{\max}+1).
\]
\end{Lemma}

\begin{proof}
Let $J_2(z)$ be the Jacobian of $L_2$ w.r.t.\ its input. By causal locality, the block row of $J_2(z)$ for node $j$ has nonzero blocks only in columns $i=j$ or $i\in\mathrm{Pa}_G(j)$; thus each block row has at most $\Delta_{\max}+1$ nonzero blocks, each with operator norm $\le C_{\mathrm{edge}}$. Hence 
\(
\|J_2(z)\|_{1}\le(\Delta_{\max}+1)C_{\mathrm{edge}}
\)
and 
\(
\|J_2(z)\|_{\infty}\le(\Delta_{\max}+1)C_{\mathrm{edge}}.
\)
Using $\|J_2(z)\|_{2}\le \sqrt{\|J_2(z)\|_{1}\|J_2(z)\|_{\infty}}$ yields $\|J_2(z)\|_{2}\le(\Delta_{\max}+1)C_{\mathrm{edge}}$ uniformly; therefore $\mathrm{Lip}(L_2)\le(\Delta_{\max}+1)C_{\mathrm{edge}}$. For $f=L_3\circ L_2\circ L_1$, Lipschitz constants multiply, giving the last inequality after absorbing constants.
\end{proof}

\subsubsection{Rademacher complexity: geometric and causal facets}

\paragraph{Geometric Rademacher via capacity of the manifold}

\begin{Definition}[Geometric capacity functional]\label{def:geom-capacity}
Let $(\M,g)$ be a compact connected $d_\M$-dimensional Riemannian manifold without boundary, with diameter $\operatorname{diam}(\M)$, volume $\operatorname{Vol}(\M)$, and reach $\tau>0$.
For $\varepsilon>0$, denote by $N_\M(\varepsilon)$ the minimal cardinality of an $\varepsilon$-net of $\M$ in the geodesic metric $d_g$.
Then the {geometric capacity functional} at resolution $\varepsilon_0\in(0,\tau]$ is
\begin{equation}\label{eq:geom-capacity}
\mathcal{C}_{\mathrm{geo}}(\M;\varepsilon_0)
:= \frac{1}{\operatorname{diam}(\M)}\int_{\varepsilon_0}^{\operatorname{diam}(\M)}
\sqrt{\log N_\M(\varepsilon)}\,d\varepsilon.
\end{equation}
\end{Definition}

\begin{Lemma}[Covering number bounds under bounded curvature and reach]\label{lem:covering}
Suppose that $(\M,g)$ has sectional curvatures bounded in absolute value by $\kappa_{\max}$, reach $\tau>0$, and diameter $D:=\operatorname{diam}(\M)$. Then there exist constants $C_1,C_2>0$ depending only on $(d_\M,\kappa_{\max},\tau,\operatorname{Vol}(\M))$ such that
\begin{equation}\label{eq:covering-bounds}
C_1\Big(\frac{D}{\varepsilon}\Big)^{\!d_\M}
\;\le\;
N_\M(\varepsilon)
\;\le\;
C_2\Big(\frac{D}{\varepsilon}\Big)^{\!d_\M},
\qquad 0<\varepsilon\le \tau.
\end{equation}
\end{Lemma}

\begin{proof}
The upper bound follows by a volume-packing argument: one can cover $\M$ by at most $\operatorname{Vol}(\M)/\operatorname{Vol}(B_g(\varepsilon/2))$ geodesic balls of radius $\varepsilon/2$ when curvature and injectivity radius are bounded. The lower bound follows from disjointness of $\varepsilon$-balls centered on an $\varepsilon$-separated set. Constants depend on volume comparison (Bishop--Gromov).
\end{proof}

\begin{Corollary}[Scaling of geometric capacity]\label{cor:geom-capacity-scaling}
Under Lemma~\ref{lem:covering}, 
\[
\mathcal{C}_{\mathrm{geo}}(\M;\varepsilon_0)
\sim
\sqrt{d_\M}\,\log\!\frac{\operatorname{diam}(\M)}{\varepsilon_0},
\]
for $\varepsilon_0\le \tau$.
\end{Corollary}

\begin{proof}
Substitute \eqref{eq:covering-bounds} into \eqref{eq:geom-capacity} and integrate:
\[
\mathcal{C}_{\mathrm{geo}}(\M;\varepsilon_0)
\le
\frac{\sqrt{d_\M}}{D}
\int_{\varepsilon_0}^{D}
\!\sqrt{\log\!\frac{C_2 D^{d_\M}}{\varepsilon^{d_\M}}}\,d\varepsilon
\;\lesssim\;
\sqrt{d_\M}\,\log\!\frac{D}{\varepsilon_0}.
\]
The lower bound is analogous using $C_1$.
\end{proof}

\begin{Remark}
The functional $\mathcal{C}_{\mathrm{geo}}(\M;\varepsilon_0)$ controls the entropy integral in the Rademacher complexity bound for Lipschitz functions $f:\M\to[-1,1]$:
\[
\widehat{\mathfrak R}_S(\F_{L})\ \lesssim\ 
\frac{L\,\operatorname{diam}(\M)}{\sqrt{M}}
\Big(1+\mathcal{C}_{\mathrm{geo}}(\M;\varepsilon_0)\Big).
\]
\end{Remark}

\begin{Theorem}[Manifold Rademacher bound with geometric capacity] \label{th:manifold-rad}
Let $\F_{\M,L}$ be the class of $L$--Lipschitz predictors $f:\M\to[-1,1]$ on compact $(\M,g)$.
Then for any sample $S=\{x_i\}_{i=1}^M\subset \M$,
\begin{equation}\label{eq:manifold-rad-bound}
\widehat{\mathfrak R}_S(\F_{\M,L})
\ \le\
\frac{c\,L\,\operatorname{diam}(\M)}{\sqrt{M}}
\Big(1+\mathcal{C}_{\mathrm{geo}}(\M;\varepsilon_0)\Big),
\end{equation}
for any $\varepsilon_0\in(0,\tau]$, with $c>0$ universal.
\end{Theorem}

\begin{proof}
Apply Dudley~\cite{BartlettMendelson2002,Dudley1967,LedouxTalagrand1991,Vershynin2018}’s entropy integral:
\[
\widehat{\mathfrak R}_S(\F_{\M,L})
\ \le\
\frac{12}{\sqrt{M}}\int_0^{\operatorname{diam}(\M)}
\sqrt{\log N_\M(\varepsilon)}\,d\varepsilon.
\]
For $L$--Lipschitz functions, each $\varepsilon$-ball contributes oscillation $\le L\varepsilon$, so we truncate the integral at $\varepsilon_0$ (scales below the reach). Normalizing by $\operatorname{diam}(\M)$ yields the bound.
\end{proof}

\paragraph{Causal Rademacher and combined causal bounds}

\begin{Theorem}[Rademacher and stability bounds under causal sparsity]\label{th:rad-stab-causal}
Let $\mathcal{F}_{\mathrm{causal}}(\Delta,s)$ be the class realized by our causal architecture with DAG $G=(V,E)$, at most $s$ edges and maximum in-degree $\le \Delta$.  
All other parameters (matrix weights, biases, readout vectors) lie in fixed compact boxes independent of $d=|V|$.  
Then for every sample $S=\{(X_i,y_i)\}_{i=1}^M$:

\begin{enumerate}[label=(\roman*), leftmargin=1.2em]
\item \textbf{(Rademacher complexity)} There exists $C>0$ such that
\begin{equation}\label{eq:rad-causal}
\widehat{\mathfrak R}_S\big(\mathcal{F}_{\mathrm{causal}}(\Delta,s)\big)
\ \le\
\frac{C}{\sqrt{M}}\,
\sqrt{p+s}\,
\Lambda,
\qquad
\Lambda:=L_{\sigma_3}\,L_{\psi}\,\big(L_{\sigma_1}R_A+R_b\big)R_a+R_{b_3}.
\end{equation}

\item \textbf{(Uniform stability)} With $L_\ell$--Lipschitz loss,
\begin{equation}\label{eq:stab-causal}
\beta_{\mathrm{causal}}
\ \le\
\frac{C'}{M}\,(\Delta+1),
\end{equation}
where $C'$ depends only on the boxes and $L_\ell$.

\item \textbf{(Dense baseline scaling)} For dense noncausal baselines with $s=\Theta(d^2)$, $\Delta=\Theta(d)$,
\[
\widehat{\mathfrak R}_S\big(\mathcal{F}_{\mathrm{dense}}\big)
\ =\ \Omega\!\left(\frac{\sqrt{p+d^2}}{\sqrt{M}}\right),
\qquad
\beta_{\mathrm{dense}}\ =\ \Omega\!\left(\frac{d}{M}\right).
\]
\end{enumerate}
\end{Theorem}

\begin{proof}
\textbf{(i) Rademacher).} The edge-weight coordinates live in an $s$--dimensional box $[\theta,R_w]^s$ with covering number $(C/\varepsilon)^s$; combining with $p$ remaining Euclidean parameters gives $N(\varepsilon)\le (C/\varepsilon)^{p+s}$. Dudley/chaining~\cite{Dudley1967,LedouxTalagrand1991,Vershynin2018} with the global Lipschitz constant $\Lambda$ yields \eqref{eq:rad-causal}.\\
\textbf{(ii) Stability).} By Bousquet--Elisseeff, $\beta\lesssim L_\ell\,\mathrm{Lip}(f)/M$. Lemma~\ref{lem:lip-causal} gives $\mathrm{Lip}(f)\le C(\Delta+1)$.\\
\textbf{(iii) Dense).} With $s=\Theta(d^2)$, $\Delta=\Theta(d)$, the same derivations give the lower bounds above; constants absorbed.
\end{proof}

\paragraph{Rademacher gain from sparsity (structural class).}
\begin{Theorem}[Rademacher gain from sparsity]\label{th:rad-struct}
Let $\mathcal{F}_{\mathrm{struct}}(s)$ be the class induced by our architecture where at most $s$ edges (or supra-edges) are active and all other parameters lie in fixed norm balls. Then
\[
\widehat{\mathfrak R}_S(\mathcal{F}_{\mathrm{struct}}(s))
\ \le\ \frac{C}{\sqrt{M}}\ \sqrt{p+s}\ \cdot \ \Lambda.
\]
For any dense baseline with $s\simeq N^2$,
\[
\widehat{\mathfrak R}_S(\mathcal{F}_{\mathrm{dense}})
\ \ge\ \frac{c}{\sqrt{M}}\ \sqrt{p+ c' N^2}\ \cdot \ \Lambda'.
\]
\end{Theorem}

\begin{proof}
Cover the parameter boxes by Euclidean nets. Weights restricted to $s$ active coordinates in $[\theta,R_w]^P$ admit covering $\binom{P}{s}(CR_w/\varepsilon)^s \le (\tfrac{eP}{s})^s(CR_w/\varepsilon)^s$. Combine with the $p$ free parameters and apply Dudley’s integral; Lipschitz composition yields $\Lambda$. For dense models, $s\sim N^2$ gives the stated lower bound.
\end{proof}

\subsubsection{Geometric synthesis and supra-graph}

\begin{Lemma}[Intrinsic Lipschitz constant via GH and stability]\label{lem:L_eff_geom}
Under the standing assumptions and for $t\ge T_0$ (identification), the effective Lipschitz constant of $f_T=L_3\circ L_2\circ L_1$ along $(\M,d_g)$ admits
\[
\mathrm{Lip}_{(\M,d_g)}(f_T)
\;\le\; C_{\mathrm{GH}}\!\big(C_1\delta+C_2 t^{-1/2}\big)\cdot
\underbrace{L_1\,\Big(C_{\mathrm{stab}}\|\Delta(w_T)\|\Big)\,L_3}_{=:L_{\mathrm{alg}}(G_T)}.
\]
Here $C_{\mathrm{GH}}(\cdot)$ is continuous with $C_{\mathrm{GH}}(0)=1$, and
$\|\Delta(w_T)\|\le \deg_{\max}(G_T)\,(1+\|w_T\|_\infty)$.
\end{Lemma}

\begin{proof}
By Theorem~\ref{th:gh}, the bi-Lipschitz distortion between $(V,d_{G_T})$ and $(\M,d_g)$ is bounded by a continuous function of $d_{\mathrm{GH}}((V,d_{G_T}),(\M,d_g))\le C_1\delta+C_2 t^{-1/2}$. Transfer of Lipschitz constants across bi-Lipschitz maps gives the factor $C_{\mathrm{GH}}(\cdot)$. The layer-wise bound follows from Lemma~\ref{lem:lip-degree}.
\end{proof}

\begin{Theorem}[Structure-aware generalization on manifolds]\label{th:geom-main}
Fix $\delta\in(0,1)$ and $t\ge T_0$. With probability at least $1-\delta-\varepsilon$,
\begin{align*}
R(f_T)\ \le\ \widehat R_S(f_T)
&\;\; +\;\underbrace{\sqrt{\frac{C_0\,|E_T|\,\log\!\frac{e\,\binom{N}{2}}{|E_T|}+\ln\frac{2\sqrt{M}}{\delta}}{2M}}}_{\text{PAC--Bayes (structure prior)}}\\[2mm]
&\;\; +\;\underbrace{\frac{C\,\deg_{\max}(G_T)}{M}}_{\text{uniform stability}}
\;\; +\;\underbrace{\frac{c\,\diam(\M)}{\sqrt{M}}\Big(1+\mathcal{C}_{\mathrm{geo}}(\M;\varepsilon_0)\Big)\,L_{\mathrm{alg}}(G_T)\,C_{\mathrm{GH}}}_{\text{Rademacher (intrinsic)}}
\;\; +\;O\!\Big(\frac{1}{M}\Big).
\end{align*}
\end{Theorem}

\begin{proof}
Add the PAC--Bayes term~\cite{Catoni2007} from Theorem~\ref{th:pacbayes-structure}, the stability term from Theorem~\ref{th:stab}, and the intrinsic Rademacher term from Theorem~\ref{th:manifold-rad} with $L=L_{\mathrm{alg}}(G_T)\,C_{\mathrm{GH}}$ controlled by Lemma~\ref{lem:L_eff_geom}.
\end{proof}

\paragraph{Two-layer supra-graph.}
Recall the supra-graph $\mathbb{G}_T$ built from both layers and inter-layer couplings.  
By Lemma~\ref{lem:spanner} and Theorem~\ref{thm:GH-supra}, $d_{\mathbb{G}_T}$ is a bi-Lipschitz spanner of $d_g$ with constants independent of $t$ on terminal strata, while the number of active (supra-)edges remains $s=O(N)$.

\begin{Corollary}[Supra-graph generalization]\label{cor:supra-gen}
Under the assumptions of Theorem~\ref{thm:GH-supra}, with probability $\ge 1-\delta-\varepsilon$,
\[
R(f_T)\ \le\ \widehat R_S(f_T)\ +\
\sqrt{\frac{C_0\,s\,\log\!\frac{e\,\binom{2N}{2}}{s}+\ln\frac{2\sqrt{M}}{\delta}}{2M}}
\ +\ \frac{\tilde C\,\deg_{\max}(\mathbb{G}_T)}{M}
\ +\ \frac{\tilde c\,\diam(\M)}{\sqrt{M}}\Big(1+\mathcal{C}_{\mathrm{geo}}(\M;\varepsilon_0)\Big)\,\tilde L_{\mathrm{alg}}\,\tilde C_{\mathrm{GH}},
\]
with $s=O(N)$ and $\deg_{\max}(\mathbb{G}_T)=O(1)$ under bounded-geometry sampling.  
\end{Corollary}

\subsubsection{Causal synthesis}

\begin{Theorem}[Structure-aware generalization for causal models]\label{th:causal-main}
Assume the identification event of Theorem~\ref{thm:cpdag} with probability $\ge 1-\varepsilon$.  
Then, for any $\delta\in(0,1)$, with probability at least $1-\delta-\varepsilon$,
\begin{align*}
R(f_T)\ \le\ \widehat R_S(f_T)
&\;\; +\;\underbrace{\sqrt{\frac{C_0\,\mathsf{CDL}(CPDAG^\star)+\ln\frac{2\sqrt{M}}{\delta}}{2M}}}_{\text{PAC--Bayes (CPDAG prior)}}\\[1mm]
&\;\; +\;\underbrace{\frac{C'(\Delta_{\max}+1)}{M}}_{\text{uniform stability}}\;\; +\;
\underbrace{\frac{C}{\sqrt{M}}\sqrt{p+|E^\star|}\,\Lambda}_{\text{Rademacher}}.
\end{align*}
\end{Theorem}

\begin{proof}
Combine Theorem~\ref{th:pacbayes-causal} (PAC--Bayes with CPDAG prior), Lemma~\ref{lem:lip-causal} and Theorem~\ref{th:rad-stab-causal}. Replace learned by true quantities on the identification event.
\end{proof}

\paragraph{Consolidated comparison under learned sparsity.}
\begin{Corollary}[Sharpening under learned sparsity]\label{cor:strict-improvement}
Suppose (i) bounded-geometry sampling on $\M$ so that the geodesic neighborhood graph has $\deg_{\max}(G^\star)\le C_{\deg}$ and $|E^\star|=O(N)$; or (ii) causal sparsity with $\Delta_{\max}=O(1)$ and $|E^\star|=O(d)$.  
Then, for fixed loss, schedules, and parameter boxes, the structural terms in all three families of bounds (PAC--Bayes, stability, Rademacher) improve relative to their dense counterparts, with the gain governed by the learned sparsity profile.
\end{Corollary}

\begin{proof}
Immediate from Theorems~\ref{th:geom-main}, \ref{cor:supra-gen}, and \ref{th:causal-main}, together with the stated sparsity regimes.
\end{proof}

\subsubsection{Compact summary tables}

\paragraph{Manifold (continuous) regime.}
\begin{table}[!ht]
\centering
\begin{threeparttable}
\scriptsize
\caption{Manifold regime: structure-aware vs.\ dense (constants suppress layer Lipschitz and box radii).}
\label{tab:geom-compare-compact}
\begin{tabularx}{\linewidth}{lYYY}
\toprule
Model &
\makecell[l]{PAC--Bayes\\(code)} &
\makecell[l]{Stability\\(replace-one)} &
\makecell[l]{Rademacher\\(intrinsic)} \\
\midrule
Dense (FC/attn) &
\(\sqrt{\frac{C_0\,P+\ln(2\sqrt{M}/\delta)}{2M}}\) &
\(\frac{c\,N}{M}\) (FC), \(\frac{c'\,N^2}{M}\) (attn) &
\(\frac{\mathrm{diam}(\M)}{\sqrt{M}}\big(\sqrt{\mathcal{N}(\lambda)}+\sqrt{p+N^2}\big)\) \\
\addlinespace[2pt]
\textbf{Learned graph (ours)} &
\(\sqrt{\frac{C_0\,|E_T|\log\!\frac{eP}{|E_T|}+\ln(2\sqrt{M}/\delta)}{2M}}\) &
\(\frac{C\,\deg_{\max}(G_T)}{M}\) &
\(\frac{\mathrm{diam}(\M)}{\sqrt{M}}\big(\sqrt{\mathcal{N}(\lambda)}+\sqrt{p+|E_T|}\big)\,C_{\mathrm{GH}}\) \\
\bottomrule
\end{tabularx}
\end{threeparttable}
\end{table}

\paragraph{Causal (discrete) regime.}
\begin{table}[!ht]
\centering
\begin{threeparttable}
\scriptsize
\caption{Causal regime: structure-aware vs.\ dense.}
\label{tab:causal-compare-compact}
\begin{tabularx}{\linewidth}{lYYY}
\toprule
Model &
\makecell[l]{PAC--Bayes} &
\makecell[l]{Stability} &
\makecell[l]{Rademacher} \\
\midrule
Dense noncausal &
\(\sim O(d)\) &
\(\frac{c\,d}{M}\) &
\(\frac{C}{\sqrt{M}}\sqrt{p+\Theta(d^2)}\) \\
\addlinespace[2pt]
\textbf{Ours} &
\(\sim \sqrt{\big(|E^\star|\log\frac{e\binom{d}{2}}{|E^\star|}+\log|\![G^\star]\!|\big)}\) &
\(\frac{C'(\Delta_{\max}+1)}{M}\) &
\(\frac{C}{\sqrt{M}}\sqrt{p+|E^\star|}\) \\
\bottomrule
\end{tabularx}
\end{threeparttable}
\end{table}

\subsubsection{Comparative generalization for structured models}\label{subsec:comparative}

\paragraph{Geometry --- Continuous Setting.}
Let $(\M,g)$ be compact with intrinsic dimension $d_\M$. Let $\{-\Delta_g\}$ have eigenpairs $(\lambda_j,\phi_j)$ and define $N(\lambda)=\#\{j:\lambda_j\le \lambda\}$.

\begin{Lemma}[Weyl and effective dimension]\label{lem:weyl}
There exist constants $c_-,c_+>0$ such that for all sufficiently large $\lambda$,
\begin{equation}\label{eq:weyl-law}
c_-\,\lambda^{d_\M/2}\le N(\lambda)\le c_+\,\lambda^{d_\M/2}.
\end{equation}
For any Mercer kernel $K$ with eigenvalues $\{\mu_j\}$ aligned with the Laplace spectrum, the effective dimension
\[
\mathcal{N}(\lambda):=\mathrm{Tr}\big((T_K+\lambda I)^{-1}T_K\big)
 =\sum_{j}\frac{\mu_j}{\mu_j+\lambda}
\]
satisfies $\mathcal{N}(\lambda)\sim \lambda^{-d_\M/2}$.
\end{Lemma}

\begin{proof}
Weyl’s law is classical for Laplace--Beltrami operators. If $\mu_j=\phi(\lambda_j)$ for a monotone decay associated with $K$ (e.g.\ heat or Sobolev kernels), then
\[
\mathcal{N}(\lambda)\approx \int_0^\infty \frac{\phi(u)}{\phi(u)+\lambda}\,dN(u)
 \sim \int_0^\infty \frac{u^{d_\M/2-1}}{1+u/\lambda}\,du \sim \lambda^{-d_\M/2}.
\]
\end{proof}

\paragraph{Baseline A: Manifold Regularization.}
\[
\hat f = \arg\min_{f\in\mathcal{H}_K}
\frac1M\sum_{i=1}^M\ell(f(x_i),y_i)
 +\lambda_A\|f\|_{\mathcal{H}_K}^2 + \lambda_I\|f\|_{\mathrm{man}}^2.
\]

\begin{Theorem}[Manifold Regularization generalization rate]\label{th:mr}
Assume squared loss, $f^\star=(T_K)^r g$ with $\|g\|\le B$ and $r\in(0,1]$, and capacity $\mathcal{N}(\lambda)\lesssim \lambda^{-d_\M/2}$.  
For $\lambda_A\sim M^{-1/(2r+1+d_\M/2)}$,
\[
\mathbb{E}\big[(\hat f(x)-y)^2\big]-\mathbb{E}\big[(f^\star(x)-y)^2\big]
 = O\!\left(M^{-\frac{2r+1}{2r+1+d_\M/2}}\right).
\]
\end{Theorem}

\begin{proof}
Using the integral-operator approach, decompose the error into bias $\|T_K^{r}(T_K+\lambda I)^{-1}g - T_K^{r}g\|$ and variance $\mathcal{N}(\lambda)/M$.  
Balancing $\lambda^{2r}$ and $\mathcal{N}(\lambda)/M$ under $\mathcal{N}(\lambda)\sim \lambda^{-d_\M/2}$ yields $\lambda\sim M^{-1/(2r+1+d_\M/2)}$ and the rate.
\end{proof}

\paragraph{Baseline B: Kernel Ridge on Manifolds.}
\[
\hat f=\arg\min_{f\in\mathcal{H}_K}\frac1M\sum_i(f(x_i)-y_i)^2+\lambda\|f\|_{\mathcal{H}_K}^2.
\]

\begin{Theorem}[Kernel Ridge on Manifolds]\label{th:krr}
Under the same conditions as Theorem~\ref{th:mr},
\[
\mathbb{E}\big[(\hat f(x)-y)^2\big]-\mathbb{E}\big[(f^\star(x)-y)^2\big]
 = O\!\left(M^{-\frac{2r}{2r+1+d_\M/2}}\right).
\]
\end{Theorem}

\begin{proof}
Follows from \cite{CaponnettoDeVito2007}: the same bias--variance balance without the manifold term, yielding the exponent $\tfrac{2r}{2r+1+d_\M/2}$.
\end{proof}

\paragraph{Baseline C: Graph Neural Networks.}
MPNN with $L$ layers:
\[
h^{(0)}=X,\quad 
h^{(\ell+1)}=\sigma(P^{(\ell)}h^{(\ell)}W^{(\ell)}),\quad 
o=\mathrm{pool}(h^{(L)}),\quad 
\hat y=w^\top o,
\]
with $\|P^{(\ell)}\|_2\le1$.

\begin{Theorem}[MPNN margin generalization]\label{th:mpnn}
With margin $\gamma$, $\|W^{(\ell)}\|_2\le s_\ell$, $\|X\|\le R$, and maximum degree $\Delta$, with probability $\ge1-\delta$,
\[
\mathcal{E}_{\mathrm{gen}}\;\le\;
\frac{C R}{\gamma}\Big(\prod_{\ell=0}^{L-1}s_\ell\Big)
\sqrt{\frac{\log N + L\log \Delta + \log(1/\delta)}{M}}.
\]
\end{Theorem}

\begin{proof}
See \cite[Thm.~3]{GargJegelkaJaakkola2020}; based on Rademacher control of Lipschitz networks and spectral-norm propagation.
\end{proof}

\paragraph{Our latent-graph (spectral form).}
\begin{Lemma}[Homology-to-spectrum link]\label{lem:h2spec}
Under the good-cover condition and for scales below injectivity radius, the first Betti number satisfies
$\beta_1(\M)\le C\,\mathcal{N}(\lambda)$ for some constant $C>0$.
\end{Lemma}

\begin{proof}
The discrete Hodge Laplacian on the nerve complex $\check C_r(V)$ shares its nonzero spectrum with the restriction of $-\Delta_g$ up to $O(r^2)$ perturbations. Since $\beta_1$ is the dimension of the kernel of the discrete Laplacian, all nonzero eigenmodes below a spectral threshold correspond to harmonic $1$-forms. Counting modes up to that threshold yields the inequality using \eqref{eq:weyl-law}.
\end{proof}

\begin{Theorem}[Latent-graph generalization in spectral form]\label{th:ours-cont}
Under the assumptions above, with $(V,\delta)$ a $\delta$-net on $\M$ and $\mathcal{N}(\lambda)\sim \lambda^{-d_\M/2}$,
\[
\mathcal{E}_{\mathrm{gen}}
\le
C\!\left(\sqrt{\frac{\mathcal{N}(\lambda)+p+N|V|}{M}}
+\frac{\mathcal{N}(\lambda)\log M}{M}\right)
+ C_1\delta + C_2t^{-1/2}.
\]
\end{Theorem}

\begin{proof}
Substitute $\beta_1(\M)\le C\mathcal{N}(\lambda)$ from Lemma~\ref{lem:h2spec} into the latent-graph bound and use Lemma~\ref{lem:weyl} to express $\mathcal{N}(\lambda)$ through intrinsic geometry; combine with Theorem~\ref{th:gh} for GH terms.
\end{proof}

\begin{table}[!ht]
\centering
\caption{Continuous models: comparison of excess-risk scaling and capacity drivers (unified notation).}
\begin{threeparttable}
\begin{tabularx}{\linewidth}{lYYYY}
\toprule
Model & Capacity driver & Excess risk (tuned) & Additional guarantees \\
\midrule
Kernel ridge on $\M$ & $\mathcal{N}(\lambda)\!\sim\!\lambda^{-d_\M/2}$ & $M^{-\frac{2r}{2r+1+d_\M/2}}$ & --- \\
Manifold Regularization & same $\mathcal{N}(\lambda)$ & $M^{-\frac{2r+1}{2r+1+d_\M/2}}$ & semi-supervised \\
MPNN (GNN) & $\prod\|W\|_2$, $\Delta$, $L$ & $\tilde O\!\big(\tfrac{\prod\|W\|}{\gamma}\sqrt{\tfrac{\log N+L\log\Delta}{M}}\big)$ & --- \\
\textbf{Ours} & $\mathcal{N}(\lambda)$, $d_{\mathrm{GH}}$ & $\sqrt{\frac{\mathcal{N}(\lambda)+p+N|V|}{M}}+\frac{\mathcal{N}(\lambda)\log M}{M}$ & GH \& homology \\
\bottomrule
\end{tabularx}
\end{threeparttable}
\end{table}

\paragraph{Causality --- Discrete Setting.}
\begin{Theorem}[PC consistency]\label{th:pc}
Let $\Delta_{\max}$ be the maximum degree of the true DAG and $\rho_{\min}$ the minimal nonzero partial correlation.  
In Gaussian models, if $M\gtrsim \Delta_{\max}^2\log d/\rho_{\min}^2$, then PC recovers the correct CPDAG with probability $\to1$ as $M\to\infty$.
\end{Theorem}

\begin{proof}
See \cite{kalisch2007pc}: uniform convergence of Fisher's $z$-transformed partial correlations and union bound over conditioning sets of size $\le\Delta_{\max}$.
\end{proof}

\begin{Theorem}[I-MEC orientation consistency]\label{th:imec}
With perfect-intervention coverage $\bigcup_\ell I_\ell=[d]$, all compelled edges in the I-MEC are correctly oriented by the greedy I-MEC algorithm under consistent tests, w.p.~$\to1$ as $M\to\infty$.
\end{Theorem}

\begin{proof}
Follows from \cite{hauser2012imec}: interventions remove ambiguity in adjacencies touching intervened nodes; Meek closure completes orientations.
\end{proof}

\begin{table}[!ht]
\centering
\caption{Discrete causal models: comparison of complexity parameters.}
\begin{threeparttable}
\begin{tabularx}{\linewidth}{lYYY}
\toprule
Model & Complexity parameters & Sample/batch complexity & Target \\
\midrule
PC (constraint-based) & $\Delta_{\max}$, $\rho_{\min}$ & $M\!\gtrsim\! \Delta_{\max}^2\log d/\rho_{\min}^2$ & CPDAG \\
I-MEC (interventional) & coverage $\{I_\ell\}$ & finite with full coverage & CPDAG/I-MEC \\
\textbf{Ours} & $\Delta_{\max}$, $\Delta_{\mathrm{grad}}$ & $B_t\!\gtrsim\! \log(dt)/\Delta_{\mathrm{grad}}^2$ & CPDAG (w.h.p.)\\
\bottomrule
\end{tabularx}
\end{threeparttable}
\end{table}

\medskip
\noindent
\textbf{Connecting line.} In the geometric regime, our latent-graph model aligns with intrinsic-dimension rates while additionally controlling GH distortion and homology. In the causal regime, our gradient-based identification parallels PC/I-MEC consistency in its dependence on sparsity/coverage, and learned structure reduces generalization terms relative to dense baselines across PAC--Bayes, stability, and Rademacher families.

\end{document}